\documentclass[english,british,iicol]{sn-jnl}
\usepackage[T1]{fontenc}
\usepackage[latin9]{inputenc}
\usepackage{xcolor}
\usepackage{colortbl}
\usepackage{array}
\usepackage{verbatim}
\usepackage{cprotect}
\usepackage{url}
\usepackage{multirow}
\usepackage{amsmath}
\usepackage{amssymb}
\usepackage{graphicx}
\usepackage[numbers]{natbib}
\PassOptionsToPackage{normalem}{ulem}
\usepackage{ulem}
\usepackage{pdfpages}

\makeatletter

\providecommand{\tabularnewline}{\\}
\providecolor{lyxadded}{rgb}{0,0,1}
\providecolor{lyxdeleted}{rgb}{1,0,0}
\DeclareRobustCommand{\mklyxadded}[1]{\textcolor{lyxadded}\bgroup#1\egroup}
\DeclareRobustCommand{\mklyxdeleted}[1]{\textcolor{lyxdeleted}\bgroup\mklyxsout{#1}\egroup}
\DeclareRobustCommand{\mklyxsout}[1]{\ifx\\#1\else\sout{#1}\fi}

\newenvironment{lyxlist}[1]
	{\begin{list}{}
		{\settowidth{\labelwidth}{#1}
		 \setlength{\leftmargin}{\labelwidth}
		 \addtolength{\leftmargin}{\labelsep}
		 }}
	{\end{list}}

\usepackage{fancyhdr}
\usepackage{caption}
\ifdefined\showcaptionsetup
 \PassOptionsToPackage{caption=false}{subfig}
\fi
\usepackage{subfig}
\makeatother

\usepackage{babel}
\begin{document}
\author[1]{\fnm{Ana Christina} \sur{Almada Campos}}\email{campos.aca@outlook.com}
\author*[2]{\fnm{Bruno Vilhena} \sur{Adorno}}\email{bruno.adorno@manchester.ac.uk} 
\affil[1]{\orgdiv{Manchester Centre for Robotics and AI}, \orgname{The University of Manchester}, \orgaddress{\street{Oxford Rd}, \city{Manchester}, \postcode{M13 9PL}, \country{UK}}. ORCID: https://orcid.org/0000-0002-7800-5640}
\affil[2]{\orgdiv{Manchester Centre for Robotics and AI}, \orgname{The University of Manchester}, \orgaddress{\street{Oxford Rd}, \city{Manchester}, \postcode{M13 9PL}, \country{UK}}. ORCID: https://orcid.org/0000-0002-5080-8724}

\abstract{Communication between humans and artificial agents is essential
for their interaction. This is often inspired by human communication,
which uses gestures, facial expressions, gaze direction, and other
explicit and implicit means. This work presents interaction experiments
where humans and artificial agents interact through explicit and implicit
communication to evaluate the effect of mixed explicit-implicit communication
against purely explicit communication and the impact of the task difficulty
in this evaluation. Results obtained using Bayesian parameter estimation
show that the task execution time did not significantly change when
mixed explicit and implicit communications were used in neither of
our experiments, which varied in the type of artificial agent (virtual
agent and humanoid robot) used and task difficulty. The number of
errors was affected by the communication only when the human was executing
a more difficult task, and an impact on the perceived efficiency of
the interaction was only observed in the interaction with the robot,
for both easy and difficult tasks. In contrast, acceptance, sociability,
and transparency of the artificial agent increased when using mixed
communication modalities in both our experiments and task difficulty
levels. This suggests that task-related measures, such as time, number
of errors, and perceived efficiency of the interaction, as well as
the impact of the communication on them, are more sensitive to the
type of task and the difficulty level, whereas the combination of
explicit and implicit communications more consistently improves human
perceptions about artificial agents.}

\title{A study on the effects of mixed explicit and implicit communications
in human-artificial-agent interactions}

\maketitle
\pagestyle{fancy}
\fancyhf{}
\fancyhead[RO,LE]{\thepage}
\renewcommand{\headrulewidth}{0pt}

\selectlanguage{english}%
\global\long\def\dq#1{\underline{\boldsymbol{#1}}}%

\global\long\def\quat#1{\boldsymbol{#1}}%

\global\long\def\mymatrix#1{\boldsymbol{#1}}%

\global\long\def\myvec#1{\boldsymbol{#1}}%

\global\long\def\mapvec#1{\boldsymbol{#1}}%

\global\long\def\dualvector#1{\underline{\boldsymbol{#1}}}%

\global\long\def\dual{\varepsilon}%

\global\long\def\dotproduct#1{\langle#1\rangle}%

\global\long\def\norm#1{\left\Vert #1\right\Vert }%

\global\long\def\mydual#1{\underline{#1}}%

\global\long\def\hamilton#1#2{\overset{#1}{\operatorname{\mymatrix H}}\left(#2\right)}%

\global\long\def\hamiquat#1#2{\overset{#1}{\operatorname{\mymatrix H}}_{4}\left(#2\right)}%

\global\long\def\hami#1{\overset{#1}{\operatorname{\mymatrix H}}}%

\global\long\def\tplus{\dq{{\cal T}}}%

\global\long\def\getp#1{\operatorname{\mathcal{P}}\left(#1\right)}%

\global\long\def\getd#1{\operatorname{\mathcal{D}}\left(#1\right)}%

\global\long\def\swap#1{\text{swap}\{#1\}}%

\global\long\def\imi{\hat{\imath}}%

\global\long\def\imj{\hat{\jmath}}%

\global\long\def\imk{\hat{k}}%

\global\long\def\real#1{\operatorname{\mathrm{Re}}\left(#1\right)}%

\global\long\def\imag#1{\operatorname{\mathrm{Im}}\left(#1\right)}%

\global\long\def\imvec{\boldsymbol{\imath}}%

\global\long\def\vector{\operatorname{vec}}%

\global\long\def\mathpzc#1{\fontmathpzc{#1}}%

\global\long\def\cost#1#2{\underset{\text{#2}}{\operatorname{\text{cost}}}\left(\ensuremath{#1}\right)}%

\global\long\def\diag#1{\operatorname{diag}\left(#1\right)}%

\global\long\def\frame#1{\mathcal{F}_{#1}}%

\global\long\def\ad#1#2{\text{Ad}\left(#1\right)#2}%

\global\long\def\spin{\text{Spin}(3)}%

\global\long\def\spinr{\text{Spin}(3){\ltimes}\mathbb{R}^{3}}%

\global\long\def\norm#1{\left\Vert #1\right\Vert }%

\global\long\def\minim#1#2{ \begin{aligned} &  \underset{#1}{\min}  &   &  #2 \end{aligned}
 }%

\global\long\def\minimone#1#2#3{ \begin{aligned} &  \underset{#1}{\min}  &   &  #2 \\
  &  \text{sujeito a}  &   &  #3 
\end{aligned}
 }%

\global\long\def\minimtwo#1#2#3#4{ \begin{aligned} &  \underset{#1}{\min}  &   &  #2 \\
  &  \text{sujeito a}  &   &  #3 \\
  &   &   &  #4 
\end{aligned}
 }%

\global\long\def\minimthree#1#2#3#4#5{ \begin{aligned} &  \underset{#1}{\min}  &   &  #2 \\
  &  \text{sujeito a}  &   &  #3 \\
  &   &   &  #4 \\
  &   &   &  #5 
\end{aligned}
 }%
\selectlanguage{british}%

\section{Introduction}

Modern robots are increasingly expected to work alongside humans,
such as in assembly and transportation \citep{Unhelkar2018,Ljungblad2012},
collaboration with humans through physical interactions \citep{Ajoudani2018},
housework \citep{Asfour2006}, and assistance to people with disabilities
\citep{Chen2013}. In addition to performance indicators, those applications
also require comprehensive analysis of human-related aspects, such
as preferences, satisfaction, and burden during the human-robot interaction
(HRI) \citep{Shah2011,Gombolay2015,Gombolay2017}.

Social robots motivate social interactions, and people tend to attribute
human characteristics to robots that communicate, cooperate, and learn
\citep{Breazeal2003}. Consequently, people rely on human social interaction
models to understand and interact with these robots \citep{Breazeal2003}.
Socially interactive robots, whose main characteristic and purpose
are to socially interact \citep{Fong2003}, can be receptionists \citep{Gockley2005}
or play educational and companion roles \citep{Toh2016,Robins2018,Meghdari2018}.

Given the importance of shared information and intentions during collaboration
\citep{Sebanz2006,Bauer2008} and the need for natural communication
in socially interactive robots \citep{Fong2003}, communication is
an essential part of HRI. Robots should interpret human communications
and convey information clearly and naturally. Therefore, it is necessary
to define the communication type that best suits each context, considering
human perceptions and communication performance and efficiency, to
satisfactorily achieve the interaction goals in a natural, intuitive
way.

To develop robots and virtual agents to interact with humans, we need
an extensive study of aspects related to their interaction and communication.
In their review, Natarajan \emph{et al}.~\citep{Natarajan2023} included
communication on the list of the grand challenges in human-robot collaboration,
mentioning specifically that which modality to use is still an open
question. This work aims to address the following question:
\begin{quote}
\emph{How does the combination of explicit and implicit communication
modalities affect human-artificial-agent interactions when compared
to explicit communication?}
\end{quote}
With that goal, we compared using only explicit communication with
mixed explicit-implicit communications to observe the effects on task-related
outcomes and in human perceptions towards a virtual agent. Following
up on the results of this first experiment, we conducted a second
one to investigate how the task difficulty affects the impact of the
communication configuration in a human-robot interaction. The goal
is to provide insights, based on scientifically-sound empirical data,
about human-artificial-agent interactions and experimental procedures
that will serve as a stepping-stone for further works on the development
of techniques for intuitive communication in HRI.

\subsection{Contributions}

The contributions of this work are empirical and offer insights on
the design of artificial agents to interact with humans. We provide
scientifically sound empirical data showing that:
\begin{itemize}
\item combining explicit and implicit communications in interactions between
humans and artificial agents, as opposed to using only explicit communications,
can improve the acceptance, sociability, and transparency of the agent; 
\item the impact of the communication type in the number of task errors
is more sensitive to the task difficulty.
\item {unlike prior studies that have strongly stated that
combining explicit and implicit communication consistently improves}{\emph{
}}{HRI's task-related outcomes, the relationship between
explicit and implicit communication is more nuanced for time}{\emph{
}}{and perceived efficiency of the interaction and
might be affected by other aspects such as the agent's embodiment
and type of task.}
\end{itemize}
{From the perspective of experimental design, unlike
the mainstream literature, we propose:}
\begin{itemize}
\item {the separation between task-related and agent-related
outcomes when investigating and leveraging on the impact of different
communication configurations;}
\item {a shift from a frequentist approach (using null hypothesis
significance tests and }{\emph{p}}{-values)
to a Bayesian approach, in which the estimation of parameters related
to objective and subjective measures obtained in this research can
be more easily incorporated as prior distributions into future research
related to similar types of interactions presented in this work.}
\end{itemize}

\section{Human-robot communication}

According to Mavridis \citep{Mavridis2015}, two aspects motivate
the development of interactive robots that use natural human communications.
First, we can take advantage of the human interaction and teaching
capabilities so robots can learn and adapt while interacting, minimising
the need for experts to program and reprogram the robots. Second,
several applications benefit from this natural communication, such
as socially interactive robots assisting humans.

Humans use several means to communicate, from verbal languages and
gestures to more subtle communications, such as facial expressions,
speech intonation, and eye gaze, which can be cues for people's internal
state and be used to estimate intentions during interactions. These
different communication types can be classified as explicit and implicit.
Some authors define them based on intention: explicit communications
convey information deliberately (\textit{e.g.}, pointing and head
gestures) whereas in implicit communications the information is inherent
to the behaviour (\textit{e.g.}, facial expressions and eye gaze)
\citep{Breazeal2005a,Bauer2008}. Others treat deliberate and unambiguous
communication as explicit (\textit{e.g.}, haptic signals with predefined
meanings), and communication where information is incorporated to
a behaviour or action and for which interpretation is context-dependent
as implicit (\textit{e.g.}, change of direction and speed during movement)
\citep{Knepper2017,Che2020}. However, some modalities can be difficult
to classify under these definitions. For instance, people can use
and alter the intensity of facial expressions when they know they
are being observed, thus serving as a communicative act \citep{Bavelas1986},
which could make them explicit if classified based on intention.

In an attempt to make this classification easier, in the present work,
we define these communication types considering what the aforementioned
definitions have in common: explicit communications are directly interpreted
whereas implicit communications require more subjective inferences
and interpretation. Table~\ref{tab:definitions-ex-im} summarises
the definitions of explicit and implicit communications from the literature
and the present work.

\begin{table}
\caption{\label{tab:definitions-ex-im}Definitions of explicit and implicit
communications from the literature and the present work.}

\centering{}%
\begin{tabular}{>{\centering}p{0.24\columnwidth}>{\centering}m{0.24\columnwidth}>{\centering}m{0.35\columnwidth}}
 & {\footnotesize\textbf{Explicit}} & {\footnotesize\textbf{Implicit}}\tabularnewline[\doublerulesep]
\hline 
{\footnotesize\textbf{\citep{Breazeal2005a,Bauer2008}}} & {\footnotesize\vspace{0.1cm}
Information conveyed deliberately\vspace{0.1cm}
} & \centering{}{\footnotesize\vspace{0.1cm}
Information inherent to the behaviour\vspace{0.1cm}
}\tabularnewline
\hline 
{\footnotesize\textbf{\citep{Knepper2017,Che2020}}} & {\footnotesize\vspace{0.1cm}
Deliberate and unambiguous\vspace{0.1cm}
} & {\footnotesize\vspace{0.1cm}
Information incorporated to behavior/action, with context-dependent
interpretation\vspace{0.1cm}
}\tabularnewline
\hline 
{\footnotesize\textbf{Present work}} & {\footnotesize\vspace{0.1cm}
Directly interpreted\vspace{0.1cm}
} & {\footnotesize\vspace{0.1cm}
Subjective inferences and interpretation\vspace{0.1cm}
}\tabularnewline
\hline 
\end{tabular}
\end{table}

Communication modalities, such as gestures, speech, gaze, haptic and
physiological signals, and facial expressions, can be explored separately
in HRI, enabling the robot to convey information and also perceive
and interpret information conveyed by humans. However, platforms combining
multiple communication modalities to perceive and produce different
explicit and implicit communications might be used to create more
complex systems and a richer experience. Some works use humanoids
or virtual agents with abilities to speak, direct their gaze, and
make gestures and facial expressions, and systems to monitor and interpret
human speech, gestures, gaze direction, head orientation, eye movements,
and physiological signals to investigate physical, cognitive, emotional,
and behavioural aspects in HRI \citep{Lenz2010,Zhang2010,Lazzeri2014}.

\subsection{\label{subsec:comparative-studies}Interaction experiments and comparative
studies}

With all these communication possibilities, several questions arise:
\begin{enumerate}
\item Which communication type is more appropriate for each context?
\item Does including implicit communications improve HRI?
\item Do different communication types affect human perceptions and both
agents' performance during interaction?
\item How is the robot communication interpreted?
\end{enumerate}
Experiments proposing interactions between humans and robots or virtual
agents aim to answer some of these questions.

Bruce \textit{et al.} \citep{Bruce2002} investigated whether a more
expressive robot with a face producing facial expressions and head
movement to indicate gaze direction would affect people's willingness
to interact with it. In their experiments, the number of people willing
to interact with the robot increased when it used facial expressions.
Breazeal \textit{et al.} \citep{Breazeal2005a} explored explicit
and implicit communications in a task where a person teaches the robot
buttons' names and then make it press them. They compared two conditions:
when the robot uses only explicit communication, such as voice to
inform its internal state when requested, and another one in which
the robot uses both explicit and implicit communications, such as
voice, gaze, facial expressions, and eye blinking to convey vivacity.
In both conditions, the person communicated only explicitly through
voice and gestures. Their study indicates that participants had a
better understanding of the robot and created better mental models
about it when it used the two communication types. Also, in the mixed
explicit-implicit condition, the task execution time was smaller and
errors during the task were identified faster and better mitigated.
To understand how people use and interpret seemingly unintentional
cues leaked through the robot's gaze, Mutlu \textit{et al.} \citep{Mutlu2009}\textit{
}proposed a game where a person should find out an object the robot
chooses by asking \textit{yes} or \textit{no} questions. In the condition
including implicit communication, the robot glanced at the chosen
object before answering the question. They used two humanoid robots
and observed that people identified the correct object quicker and
with fewer questions when the android robot leaked cues through its
gaze. In a study of a long term interaction, Tanaka \textit{et al.}
\citep{Tanaka2012} obtained results suggesting that the company of
a communicative robot able to talk and nod can improve cognitive functions
and other aspects of the daily life of elderly women living alone,
when compared to the same robot without the communicative features.
Huang and Mutlu \citep{Huang2016} showed that, when the robot uses
the human gaze direction to anticipate explicit commands and act accordingly,
the task is better performed and the robot is perceived as more aware
of the interaction. Zhang \textit{et al.} \citep{Zhang2025} showed
that team performance, trust, and anthropomorphism perceived by the
human are improved when the robot is able to understand implicit information
conveyed by indirect speech acts. The authors also highlight that
this capability can affect differently depending on the context, and
therefore it should be used carefully. Six \textit{et al.} \citep{Six2025}
evaluated the use of animation features in virtual agents in brief
cognitive behavioural therapy based mental health apps. In this context,
their results suggest that a virtual agent with body movements and
facial expressions can improve user experience, in contrast to a static
one with blank facial expression. Monitoring physiological signals,
Apraiz \textit{et al.} \citep{Apraiz2025} observed reduced mental
effort and increased emotional valence when the user interface included
multiple modalities (visual and auditory) compared with a unimodal
approach (visual only) in disassembly tasks with a human-robot team.

Using their model for bidirectional gaze in human and virtual agent
interactions, Andrist et al. \citep{Andrist2017} observed improved
task performance when the virtual agent both produced gaze and responded
to human gaze. Participants also perceived the virtual agent as more
expressive and with greater cognitive abilities when it produced gaze,
and more competent when produced and responded to it. In Buschmeier
and Kopp's work \citep{Buschmeier2018}, an attentive speaker agent,
which estimated the human mental states during the interaction and
adapted its behaviour, received more feedback signals from the human
and was perceived as an attentive agent by them. Iwasaki \textit{et
al.} \citep{Iwasaki2019} observed that a robot recognising and responding
to people's behaviours encourages them to interact with it. Che \textit{et
al.} \citep{Che2020} studied the communication effects when the navigation
of one agent affects the navigation of another (\textit{i.e.}, social
navigation). Their experiment showed that when the robot communicated
its intention explicitly and implicitly and predicted human movements,
participants navigated more efficiently and it increased their trust
and understanding of the robot, compared to when the robot predicted
movements and communicated only explicitly, and when the robot executed
only collision avoidance without prediction. Hong et al. \citep{Hong2021}
proposed a system to recognise and classify human affect, based on
body language and vocal intonation. They also proposed a robot emotional
model that uses the human affect estimation, among other robot and
context specific aspects, to decide which emotion the robot should
convey through its eye colour, body language, and vocal intonation.
In their experiments with a robot suggesting healthy lifestyle meals
and exercises, the expressive robot with the emotional model induced
more neutral and positive valence and arousal than a neutral robot
that did not use the robot emotional model to generate emotions. 

It is also important to investigate how to use each available communication
modality. The way the robot communicates (\textit{e.g.}, how it speaks
and engages in touch interactions, and where it directs its gaze \citep{Takayama2009,Fiore2013,Hirano2018,Agrigoroaie2020,Belkaid2021})
must be carefully adjusted and can be influenced by the application
context and the general profile of people interacting with the robot.
Aspects such as the effects of robot's conveyed mood, transparency,
planning for communication, ethical concerns, and influence of people's
gender, age, culture, familiarity with robots, and other factors are
also concerns of HRI studies \citep{Gockley2006,Rau2009,Unhelkar2017,Guznov2019,VanMaris2020}.

\subsection{{Positioning our work in the literature}}

{The aforementioned literature suggests that combining
explicit and implicit communication modalities improves various aspects
of HRI, including performance indicators and human perceptions of
the artificial agent. The work hypotheses introduced in Section~\ref{subsec:Hypotheses}
are based on this literature, as it is reasonable to expect improvements
in both task- and agent-related outcomes when using this combination
of implicit and explicit modalities. However, the experimental results
presented in Sections~\ref{sec:Results} and \ref{sec:Follow-up-study}
suggest that, although the positive impact on agent-related outcomes
might be more general, task-related outcomes are more nuanced, and
the communication configuration might impact the interaction differently
depending on other aspects such as task difficulty and embodiment
of the artificial agent. Therefore, despite strong claims, the current
literature does not present a clear, robust picture of the role of
the combination of explicit and implicit communication modalities
when many different factors are at play. This paper helps shed some
light by trying to replicate similar trends described in the literature
(conceptual replication) and by analysing more nuanced factors such
as task difficulty and embodiment.}

{Also, unlike the vast majority of the literature,
which provides results in the form of }\textit{{p}}{-values,
we used a Bayesian approach to analyse our data (only one of the works
cited in this section uses a Bayesian method \citep{Buschmeier2018}).
This is a robust alternative methodology that allows results to be
easily incorporated in future research by defining more informed prior
distributions for further data analysis.}

\section{\label{sec:Experimental-design}Experimental design}

We investigate the effects of communication type on human perceptions
and task-related outcomes in a human-virtual-agent interaction. The
literature on HRI described in Section~\ref{subsec:comparative-studies}
suggests that combining explicit and implicit communications improve
the interaction. We define two communication configurations:
\begin{lyxlist}{00.00.0000}
\item [{\textit{EX:}}] Only explicit communications from human and virtual
agent.
\item [{\textit{EXIM:}}] Explicit and implicit communications from human
and virtual agent.
\end{lyxlist}

\subsection{Human-robot communication infrastructure}

We used a human-robot communication infrastructure with selected explicit
and implicit communication modalities \citep{Campos2020}. The system
is integrated in the Robot Operating System (ROS) and includes recognition
and interpretation of human pointing gestures and gaze direction,
and a virtual agent with voice, facial expressions, and gaze direction.

In addition to the systems described in \citep{Campos2020}, we included
other communication modalities such as screen applications for the
virtual agent to keep the human informed of the task progress. The
human can also insert explicit information using mouse and keyboard.
The virtual agent uses sound signals to indicate successes and errors,
and raises its eyebrows to implicitly draw the person's attention
during interaction. The human location during the interaction implicitly
indicates the current stage of the task and if instructions were followed.
Table~\ref{tab:communications} summarises the communication modalities
available in our infrastructure.
\begin{table}
\caption{\label{tab:communications}Available communication modalities for
human and virtual agent.}

\centering{}%
\begin{tabular}{>{\centering}p{0.15\columnwidth}>{\centering}m{0.28\columnwidth}>{\centering}m{0.38\columnwidth}}
 & {\footnotesize\textbf{Human}} & {\footnotesize\textbf{Virtual agent}}\tabularnewline[\doublerulesep]
\hline 
{\footnotesize\textbf{Explicit}} & {\footnotesize gestures}{\footnotesize\par}

{\footnotesize manual entries} & {\footnotesize\vspace{0.1cm}
voice}{\footnotesize\par}

{\footnotesize sound signals}{\footnotesize\par}

{\footnotesize information on screen\vspace{0.1cm}
}\tabularnewline
\hline 
{\footnotesize\textbf{Implicit}} & {\footnotesize gaze direction}{\footnotesize\par}

{\footnotesize location} & {\footnotesize\vspace{0.1cm}
facial expressions}{\footnotesize\par}

{\footnotesize raise of eyebrows}{\footnotesize\par}

{\footnotesize gaze direction\vspace{0.1cm}
}\tabularnewline
\hline 
\end{tabular}
\end{table}

Since we planned an structured interaction with well defined steps,
there was a ROS node responsible for integrating the individual systems
and managing the interaction. This manager node autonomously reads
important information and sends commands through ROS topics to each
of the other modules to carry on the interaction. We also used cameras
and markers to locate objects and other important elements in the
environment, such as the screens to display the virtual agent. The
human location was also inferred using markers. Given the specific
locations the human was instructed to be at each phase, we placed
a marker on the floor that would be occluded whenever the human reached
that specific location. After some camera frames without detection
of a specific marker on the floor, we would consider that the human
reached that location.

\subsection{\label{subsec:Hypotheses}Hypotheses}

We hypothesise that the combination of explicit and implicit communications
makes the agents' actions more transparent and predictable, which
is important to successfully achieve a collaborative goal. Hence,
we postulate the following:
\begin{lyxlist}{00.00.0000}
\item [{\textit{H1:}}] The task execution time will be smaller in the EXIM
configuration than in the EX configuration.
\item [{\textit{H2:}}] The number of task errors will be smaller in the
EXIM configuration than in the EX configuration.
\end{lyxlist}
If using explicit and implicit communications makes the virtual agent
more similar to human agents, we expect that humans understand it
better and perceive it more as a social agent, making the interaction
more natural and pleasant. Therefore, we introduce three additional
hypotheses:
\begin{lyxlist}{00.00.0000}
\item [{\textit{H3:}}] The acceptance of the virtual agent will be higher
in the EXIM configuration than in the EX configuration.
\item [{\textit{H4:}}] The virtual agent will be perceived as more sociable
in the EXIM configuration than in the EX configuration.
\item [{\textit{H5:}}] The virtual agent will be perceived as more transparent
in the EXIM configuration than in the EX configuration.
\end{lyxlist}
Lastly, we expect that the virtual agent's greater sociability and
the better task performance when combining communication types will
make the interaction be perceived as more efficient, resulting in
our final work hypothesis:
\begin{lyxlist}{00.00.0000}
\item [{\textit{H6:}}] The perceived interaction efficiency will be greater
in the EXIM configuration than in the EX configuration.
\end{lyxlist}

\subsection{Human and virtual agent interaction}

To evaluate our hypotheses, we chose to conduct the study in a well
controlled environment, so we could isolate the factor of interest
\citep{Hoffman2021}. Also, to reduce the sample size needed, we decided
for a within-subjects design, in which each participant completes
the task twice, one for each condition \citep{Hoffman2021}. We propose
an activity similar to a game with two phases that include actions
present in real collaborative scenarios, such as following instructions
and pointing to objects.

In the first phase, after introducing itself and giving instructions,
the virtual agent shows a four-colour sequence. The person should
then point to coloured boxes in the environment in the same order
as the sequence. Sound signals suggest correct and wrong indications,
and the screen application shows the task progress. The colour sequence
works as a password for the next phase, when the person is further
instructed to count the occurrence of some objects in images containing
several other items. There are four images in the workspace. Given
an object shown in each corner of the computer screen, the person
should count it on the respective image in the workspace and type
the number of occurrences on the screen application. The images' positions
were chosen to encourage people to move their heads to look at them.
In both phases, if a time limit is reached, the virtual agent finishes
the task, adding the password or filling the remaining count fields.

\begin{figure}
\begin{centering}
\begin{minipage}[t]{0.5\columnwidth}%
\begin{center}
\subfloat[Luna]{\begin{centering}
\includegraphics[width=0.7\columnwidth]{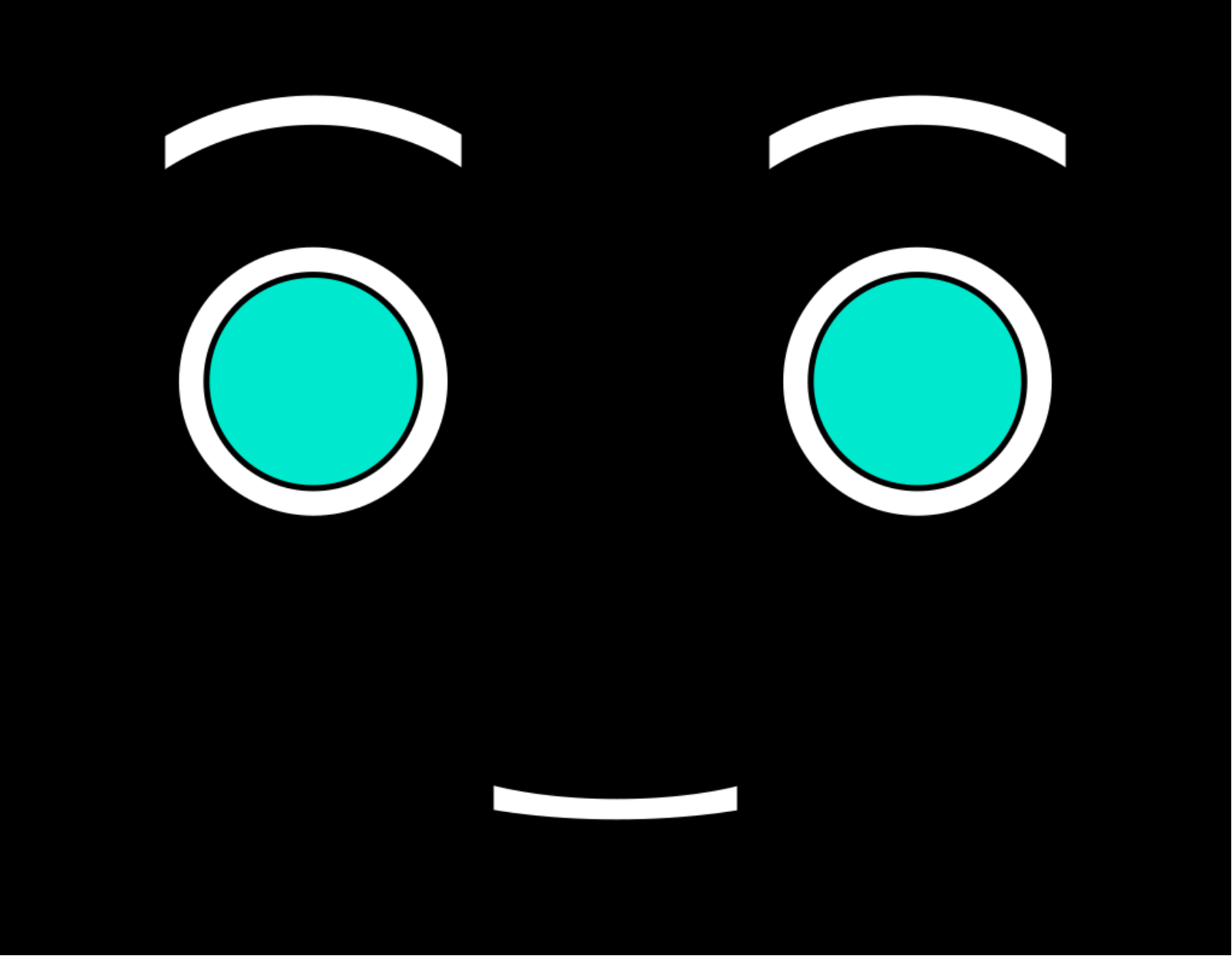}
\par\end{centering}
}
\par\end{center}%
\end{minipage}%
\begin{minipage}[t]{0.5\columnwidth}%
\begin{center}
\subfloat[Sofia]{\begin{centering}
\includegraphics[width=0.7\columnwidth]{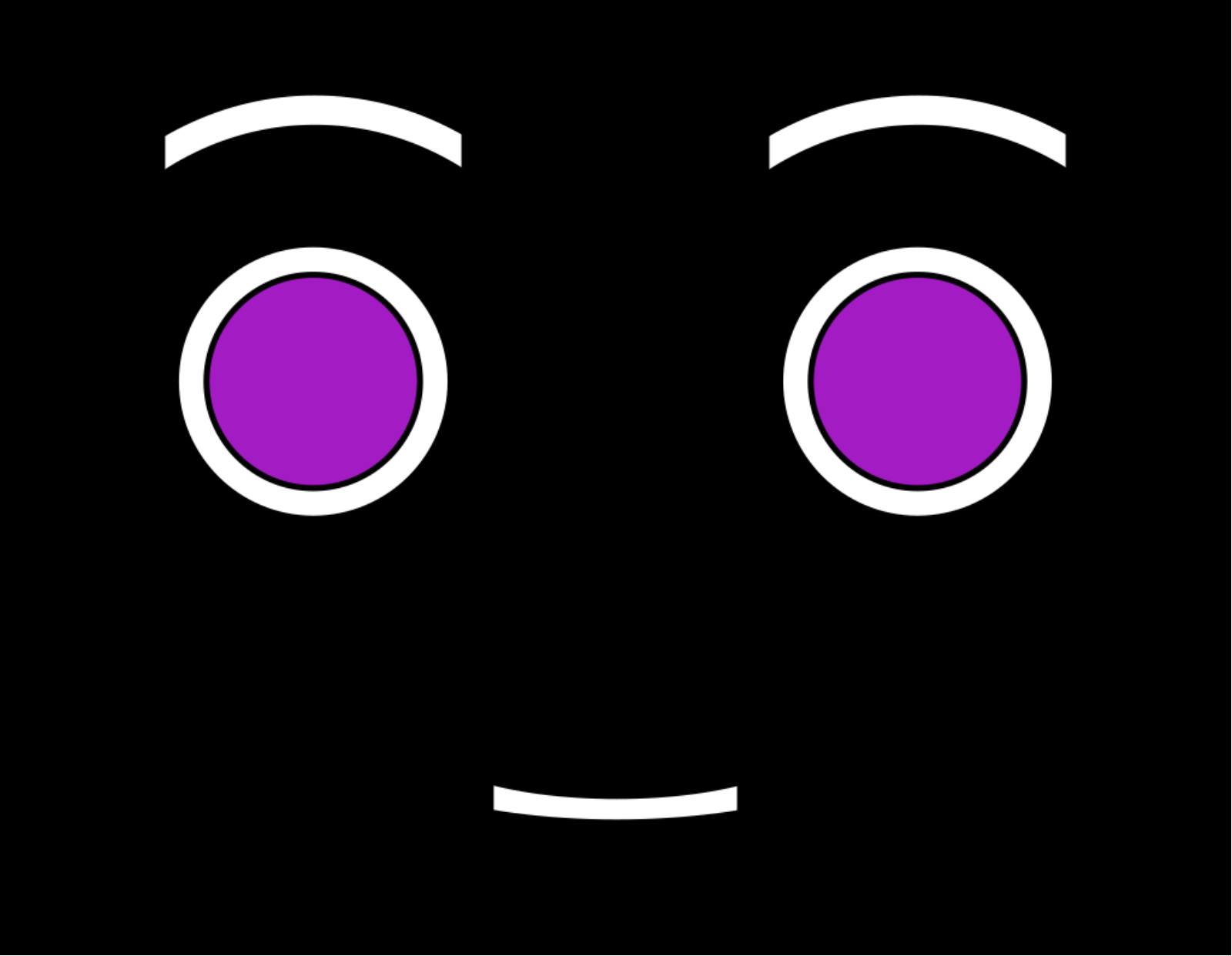}
\par\end{centering}
}
\par\end{center}%
\end{minipage}
\par\end{centering}
\centering{}\caption{\label{fig:virtual_agents}Two virtual agents created to interact
with people in the experiments.}
\end{figure}

Each participant completed both phases twice, once for each communication
configuration (EX and EXIM). Each configuration is represented by
a different virtual agent selected randomly by the system at the beginning
of the experiment, as well as the communication configuration order
for each participant. The two virtual agents, Luna and Sofia (see
Fig.~\ref{fig:virtual_agents}), differ by their names, eye colours,
and voice tones. As described in \citep{Campos2020}, we recorded
audio files with selected sentences to be the virtual agents' voices.
Since participants interact with both virtual agents, there are slight
differences in the sentences for each one to help differentiate them.
\begin{figure*}
\begin{centering}
\includegraphics[width=1\textwidth]{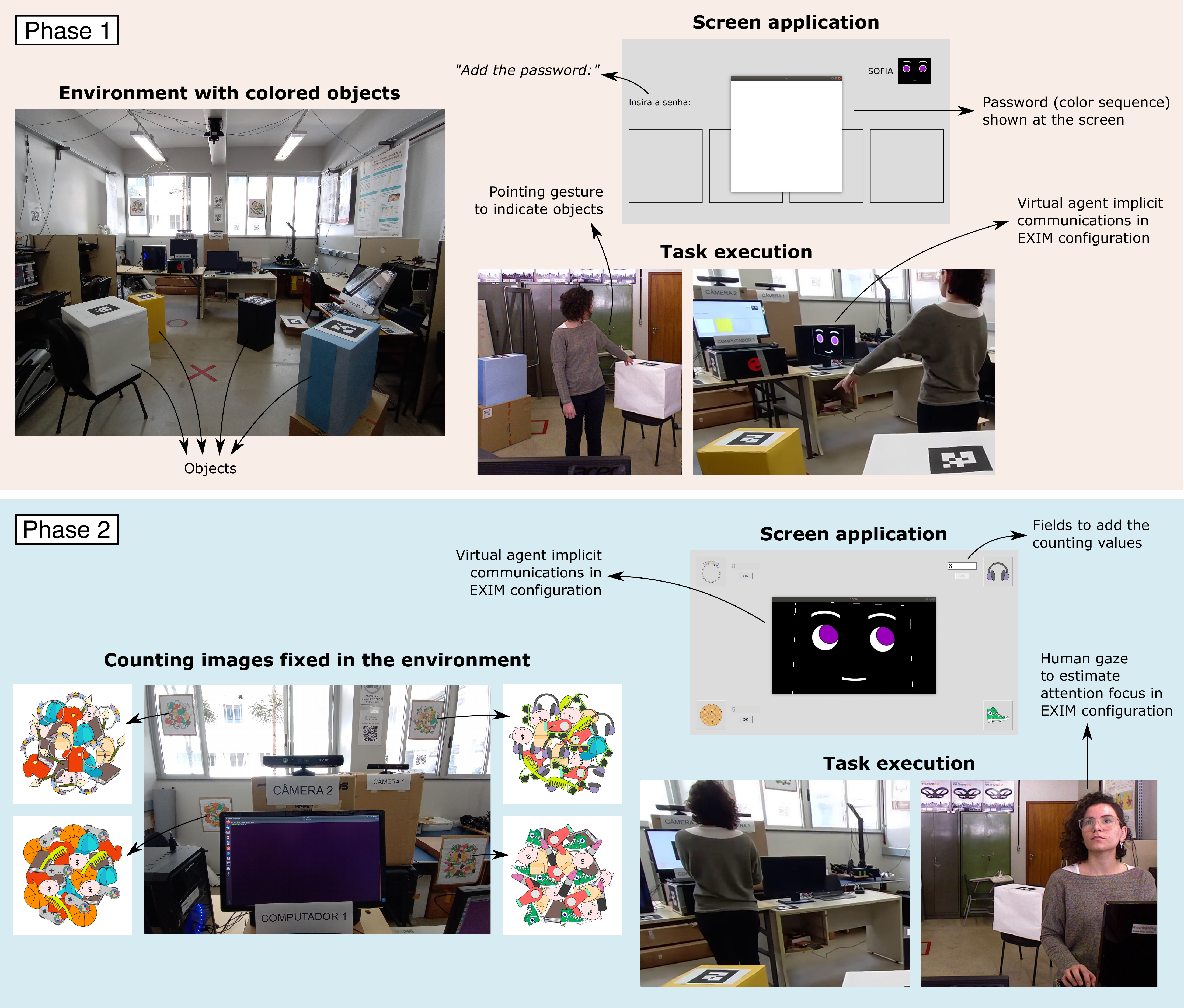}
\par\end{centering}
\caption{\label{fig:interacao}Illustrations of the proposed interaction.}
\end{figure*}

On the EX configuration, the virtual agent uses only voice with simulated
mouth movements, sound signals, and information on screen to communicate.
The agent always has a neutral expression, blinks periodically, and
looks straight ahead, while the human communicates through pointing
gestures in Phase 1 and manual entries by keyboard and mouse in Phase
2. On the EXIM configuration, along with the explicit communications
of the EX configuration, the virtual agent uses different facial expressions
according to the context, raises its eyebrows to call the human's
attention, and directs its gaze through the environment. The system
also uses implicit information from the location and the person's
gaze in specific moments of the interaction.

During EXIM's Phase 1, the virtual agent looks at the coloured box
that the person should indicate. If the person makes a mistake or
the password is repeated on screen after some wrong indications, the
virtual agent looks at the person, raises its eyebrows, and then looks
at the correct box again. After Phase 1 is finished, the virtual agent
instructs the person to go to a specific location to start Phase 2.

In the EX configuration, Phase 2 starts only after the person's explicit
command through the screen application. In the EXIM configuration,
the system anticipates the explicit command and starts Phase 2 whenever
it detects the person on the instructed location. During Phase 2 of
EXIM, if the system detects that the human is looking at one of the
images of shuffled items, the virtual agent also looks at it and verbally
suggests the correct counting value (\textit{e.g.}, ``{[}translated
from Portuguese{]} I guess it is five there.''). Moreover, when the
person opens one of the fields on the screen application to add the
counting value, the virtual agent direct its gaze to the open field.

In the EXIM configuration, the virtual agent always looks at the person's
face when speaking, except when it is giving a clue on Phase 2, since
in this moment its gaze indicates which image/object it refers to.
Fig.~\ref{fig:interacao} illustrates the interaction.

\subsection{\label{subsec:Measurements}Objective and subjective metrics}

The outcomes related to hypotheses H1 and H2 are time and number of
errors. The system registers the interaction duration, including both
phases, except for initialisation, verbal instructions, and audio
file execution times. More specifically, the time for the phase instructions
is excluded because they are only given in the first configuration.
We also exclude the times for the execution of the audio files for
the virtual agents' voices to prevent the differences in the sentences'
length from affecting the comparison between the two communication
configurations.

Errors are counted whenever the indicated colours in Phase 1 or counting
values in Phase 2 are wrong. We discard system errors, such as wrong
identification of a pointing gesture. When the task finishes due to
timeout, we count one extra error for each colour not added in Phase
1 and each counting value not inserted in Phase 2.
\begin{table}
\caption{\label{tab:likert_scales}Likert scales for the measurement of human
perception variables. Items marked with {\scriptsize (R)} are treated
with a reverse scale. The term VA is replaced by the name of the virtual
agent.}

\centering{}%
\begin{tabular}{>{\raggedright}m{0.95\columnwidth}}
\hline 
\noalign{\vskip0.02\columnwidth}
{\footnotesize\textbf{Acceptance of the virtual agent}}{\footnotesize\par}
\begin{enumerate}
\item[{\footnotesize 1.}] {\footnotesize I found VA intimidating. (R)}{\footnotesize\par}
\item[{\footnotesize 2.}] {\footnotesize I found VA friendly.}{\footnotesize\par}
\item[{\footnotesize 3.}] {\footnotesize I felt comfortable while interacting with VA.}{\footnotesize\par}
\item[{\footnotesize 4.}] {\footnotesize I liked to interact with VA.}{\footnotesize\par}
\item[{\footnotesize 5.}] {\footnotesize I found unpleasant to interact with VA. (R)}{\footnotesize\par}
\item[{\footnotesize 6.}] {\footnotesize I would like to interact more with VA.}{\footnotesize\par}
\end{enumerate}
\vfill{}
\tabularnewline
\hline 
\noalign{\vskip0.02\columnwidth}
{\footnotesize\textbf{Sociability of the virtual agent}}{\footnotesize\par}
\begin{enumerate}
\item[{\footnotesize 1.}] {\footnotesize I felt like VA understood what I was doing.}{\footnotesize\par}
\item[{\footnotesize 2.}] {\footnotesize When interacting with VA, I felt like I was with a
real person.}{\footnotesize\par}
\item[{\footnotesize 3.}] {\footnotesize Sometimes I felt like VA was really looking at me.}{\footnotesize\par}
\item[{\footnotesize 4.}] {\footnotesize Sometimes VA seemed to have real feelings.}{\footnotesize\par}
\item[{\footnotesize 5.}] {\footnotesize VA's behaviour is similar to a person's.}{\footnotesize\par}
\end{enumerate}
\vfill{}
\tabularnewline
\hline 
\noalign{\vskip0.02\columnwidth}
{\footnotesize\textbf{Transparency of the virtual agent}}{\footnotesize\par}
\begin{enumerate}
\item[{\footnotesize 1.}] {\footnotesize I understood VA.}{\footnotesize\par}
\item[{\footnotesize 2.}] {\footnotesize I was able to know what VA was ``thinking.''}{\footnotesize\par}
\item[{\footnotesize 3.}] {\footnotesize I knew when VA was paying attention on me.}{\footnotesize\par}
\item[{\footnotesize 4.}] {\footnotesize During the interaction, VA's intentions were clear
to me.}{\footnotesize\par}
\end{enumerate}
\vfill{}
\tabularnewline
\hline 
\noalign{\vskip0.02\columnwidth}
{\footnotesize\textbf{Perceived efficiency of the interaction}}{\footnotesize\par}
\begin{enumerate}
\item[{\footnotesize 1.}] {\footnotesize I could count with VA to help me during the task.}{\footnotesize\par}
\item[{\footnotesize 2.}] {\footnotesize VA helped me to execute the task.}{\footnotesize\par}
\item[{\footnotesize 3.}] {\footnotesize VA got in my way. (R)}{\footnotesize\par}
\item[{\footnotesize 4.}] {\footnotesize VA made no difference to my performance. (R)}{\footnotesize\par}
\end{enumerate}
\vfill{}
\tabularnewline
\hline 
\end{tabular}
\end{table}

The hypotheses H3 to H6 are related to subjective outcomes, namely
acceptance, transparency, and sociability of the virtual agents and
the perceived interaction efficiency. We measure each variable with
a Likert scale \citep{Likert1932} containing a set of \textit{items}
(sentences) that people answer with one of the following options:
(1) totally disagree, (2) disagree, (3) do not know, (4) agree, and
(5) totally agree. Therefore, we have five response \textit{levels}
(1 to 5), and we present them to the participants always in the same
order and without numbered labels.

Table~\ref{tab:likert_scales} shows the 19 items composing the Likert
scales, translated from Portuguese. They were defined according to
our experiment, but inspired and adapted from works such as the ones
by Heerink \textit{et al.} \citep{Heerink2010} and Iwasaki \textit{et
al.} \citep{Iwasaki2019}. In most cases, the lowest response level
(``totally disagree'') indicates a more negative perception. For instance,
to disagree with the first item of the transparency scale means that
the virtual agent was perceived as not very transparent. On the other
hand, a low level response for the first item in the acceptance scale
indicates good acceptance of the virtual agent because it uses a reverse
scale. Items with a reverse scale, for which low levels indicate positive
perceptions, are marked with an {\scriptsize (R)} in Table~\ref{tab:likert_scales}.
The 19 items from the table are presented randomly to each participant,
unlabelled and without the reverse scale indication.

\section{\label{subsec:experiment-analysis}Methodology for experimental analysis}

We use Bayesian parameter estimation in our data analysis as it provides
richer information when compared with frequentist analyses such as
null hypothesis significance tests, maximum likelihood estimation,
and confidence interval \citep{Kruschke2015,Kruschke2018,Wagenmakers2018,Kelter2020}.
Bayesian approaches are not so common in HRI studies as the analysis
using \textit{p}-values \citep{Baxter2016}, but they allow discussions
beyond the accepting or rejecting hypotheses dichotomy. In areas such
as statistics and psychology, Bayesian methods have been discussed
as alternatives to deal with the limitations and sometimes inadequate
interpretations of \textit{p}-values \citep{Kruschke2013,Kruschke2018,Wagenmakers2018,Kelter2020,Kelter2021}.
In the HRI context, Baxter \textit{et al.} \citep{Baxter2016} suggest
we focus on methods that can incrementally increase our knowledge
about phenomena of interest; that is, a Bayesian perspective. The
methods we chose to analyse our results are detailed below.

\subsection{Overview}

Bayesian methods rely on Bayes' rule to re-allocate credibility across
possibilities, using collected information. More specifically, an
initial belief on the value of a set of variables is updated after
collecting new data. Let $P(\mathcal{V})$ be the prior joint probability
of $n$ parameter values $\mathcal{V}=\{v_{1},v_{2},...,v_{n}\}$
without the data (initial belief), $P(\mathcal{D}\mid\mathcal{V})$
the likelihood to obtain the data $\mathcal{D}$ given $\mathcal{V}$,
and $P(\mathcal{D})$ the data likelihood according to the considered
model. Then, the Bayes' rule states that the posterior credibility
of $\mathcal{V}$ (updated belief) is given by
\begin{align*}
P(\mathcal{V} & \mid\mathcal{D})=\frac{P(\mathcal{D}\mid\mathcal{V})P(\mathcal{V})}{P(\mathcal{D})}.
\end{align*}

\begin{figure}
\begin{centering}
\includegraphics[width=1\columnwidth]{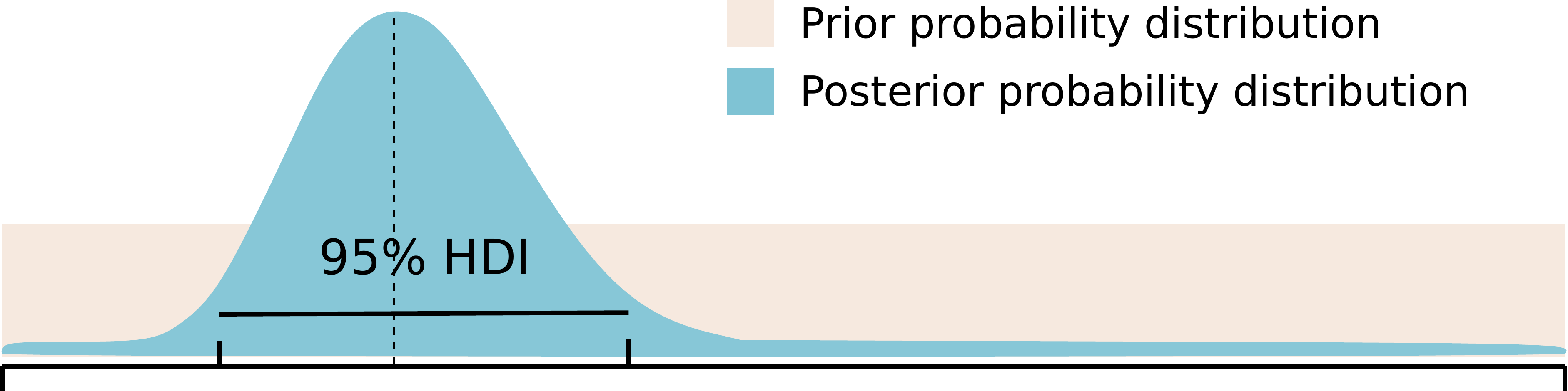}
\par\end{centering}
\caption{\label{fig:prior-and-posterior}Example of prior and posterior probability
distributions for parameter values in the Bayesian estimation. The
$95\%$ HDI includes $95\%$ of the posterior distribution and the
most credible parameter values.}
\end{figure}

The Bayesian parameter estimation allows estimating a parameter value
or the magnitude of an effect of interest. When there is few prior
information about the parameter values, we usually use prior probability
distributions that are vague and uninformative so that they have minimal
influence on the estimation \citep{Kruschke2018,Kelter2020}. Fig.~\ref{fig:prior-and-posterior}
shows a uniform prior distribution, which assigns the same credibility
for all parameter values inside an interval. The Bayes' rule provides
the posterior distribution with updated credibilities for each parameter
value, which are summarised using measures of central tendency, such
as mean, mode, and median, and intervals such as the HDI (Highest
Density Interval).

The HDI includes a percentage of the distribution, and the parameter
values inside the interval are more credible than the ones outside
of it. Therefore, the $95\%$ HDI contains $95\%$ of the distribution,
and parameter values inside of it are more credible than parameter
values outside of it \citep{Kruschke2015} because there is a 95\%
probability that the true parameter value is inside this interval.
Also, the $95\%$ HDI width indicates the estimation uncertainty:
the smaller the interval, the more precise is the estimation and the
more certain we are about the parameter value.

\begin{figure}
\begin{centering}
\includegraphics[width=1\columnwidth]{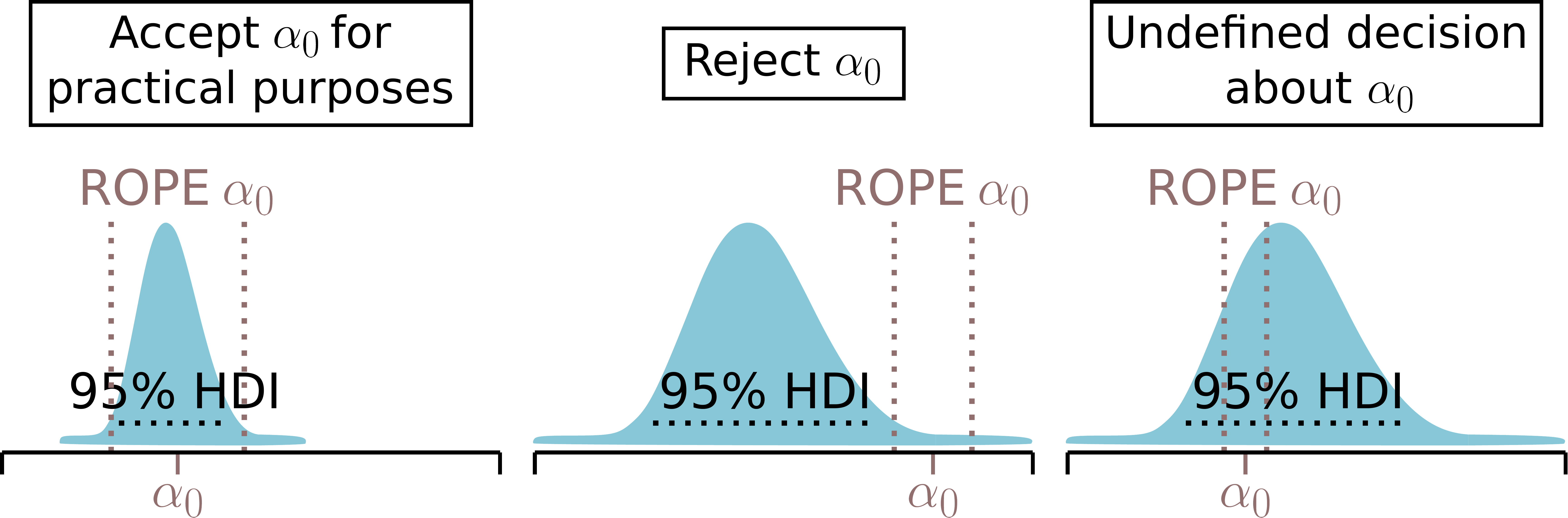}
\par\end{centering}
\caption{\label{fig:decision-rope}Examples of possible relations between the
$95\%$ HDI of the posterior distribution and a ROPE around a value
of interest, $\alpha_{0}$, for the parameter.}
\end{figure}

The posterior distribution alone already provides information about
the parameter value. Nonetheless, we can also evaluate the credibility
of specific values such as a null value indicating the absence of
an effect. The Region Of Practical Equivalence (ROPE) is defined around
the parameter value of interest to indicate a set of values that are
practically equivalent to the one of interest \citep{Kruschke2013}.
When deciding about accepting or rejecting a specific value, Kruschke
\citep{Kruschke2015} proposes a decision rule, illustrated in Fig.~\ref{fig:decision-rope},
using a ROPE around the value of interest and the $95\%$ HDI of the
posterior distribution. If the entire $95\%$ HDI is inside the ROPE,
the value of interest is accepted for practical purposes. Conversely,
if the entire $95\%$ HDI is outside the ROPE, the value of interest
is considered incredible and, therefore, rejected. If none of those
occur, the available data set is considered insufficient to make a
decision about the specific parameter value. To define the ROPE around
the value of interest, the context of the application must be taken
into consideration for it to reflect practical equivalence. With an
adequate ROPE, a value of interest is accepted only when there is
a sufficiently precise parameter estimation, which implies a sufficiently
narrow $95\%$ HDI that could fit into the ROPE.

\subsection{\label{subsec:Objective-measures}Objective measures}

In our experiment, participants interact with two virtual agents,
one for each communication configuration (EX and EXIM), resulting
in measurement pairs. One way commonly used to cancel individual variations
is to take the difference between the two observations and run the
analysis with a single group \citep{Montgomery2011}. For time and
number of errors, we take the differences
\begin{gather}
\Delta t\triangleq t_{\mathrm{EX}}-t_{\mathrm{EXIM}}\text{ and }\Delta e\triangleq e_{\mathrm{EX}}-e_{\mathrm{EXIM}},\label{eq:time and error differences}
\end{gather}
respectively, between the observations in each configuration, and
$\Delta t$ and $\Delta e$ are the final measurements associated
with each participant. There is no difference between the configurations
when $\Delta t=\Delta e=0$, and positive differences ($\Delta t>0$
and $\Delta e>0$) favour our hypotheses.

\begin{figure}
\begin{centering}
\includegraphics[width=0.8\columnwidth]{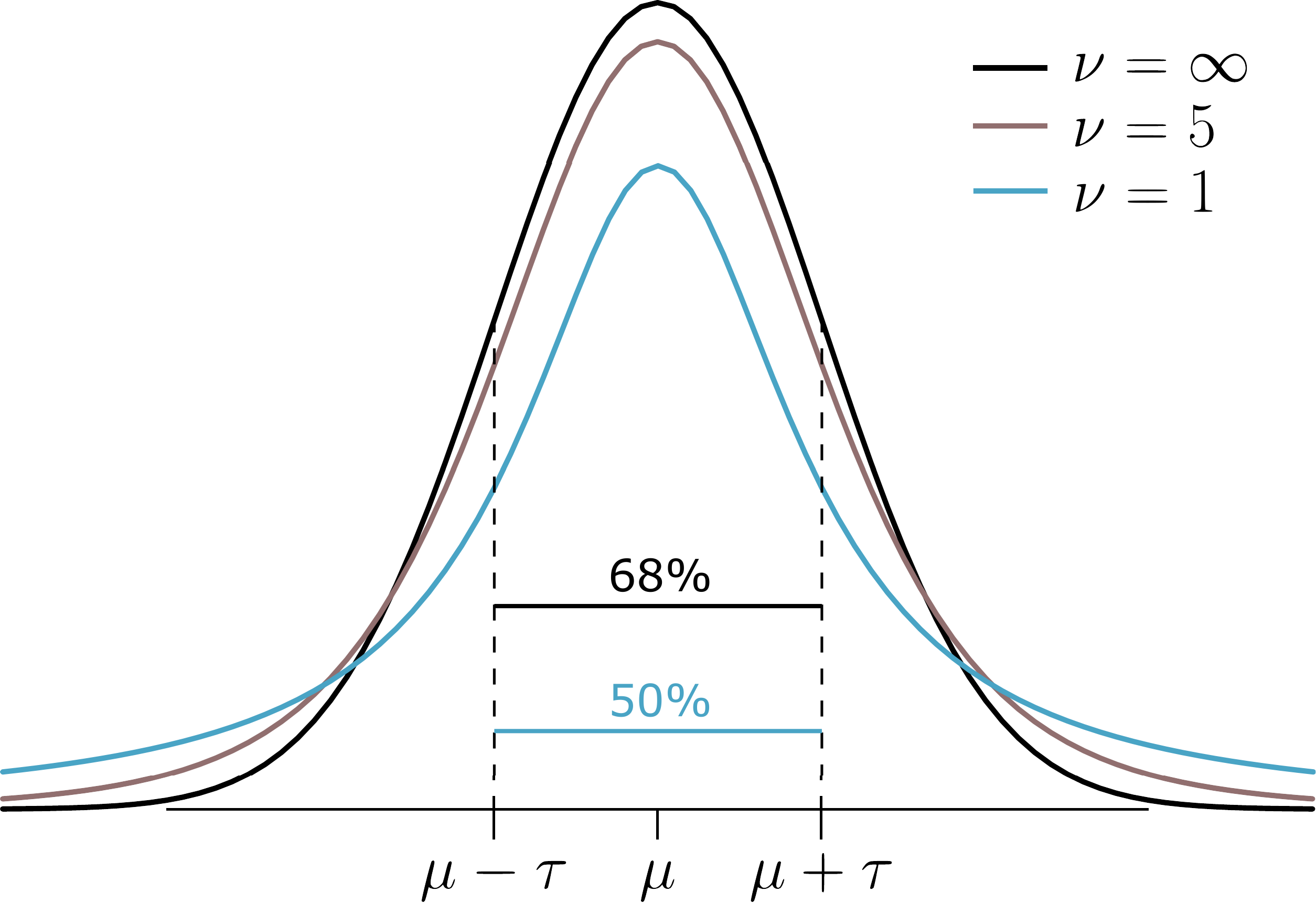}
\par\end{centering}
\caption{\label{fig:t-distribution}Examples of \textit{t} distributions with
mean $\mu$, scale $\tau$, and different normality parameters $\nu$.
The greater $\nu$ is, the closer the \textit{t} distribution is to
a normal distribution. The scale $\tau$ is related to the spread
of the data and covers $50\%$ of the \textit{t} distribution with
$\nu=1$ and $68\%$ of the distribution with $\nu=\infty$.}
\end{figure}

We treat time and number of errors as metric variables in interval
or ratio scales and represent them using $t$ distributions because
of the heavier (with higher probabilities) tails that accommodate
outliers better than the normal distribution \citep{Kruschke2015}.
The distribution is described using mean $\mu$, scale $\tau$, and
a normality parameter $\nu\in(0,\infty)$, all illustrated in Fig.~\ref{fig:t-distribution}.
The scale $\tau$ is related to the spread of the data and $\nu$
determines the heaviness of the distribution tails. The greater $\nu$
is, the closer the \textit{t} distribution is to a normal distribution.\footnote{For more about the scale and normality parameters of the \textit{t}
distribution, please refer to Section~16.2 of \citep{Kruschke2015}.} The goal of the Bayesian inference is to estimate the parameters
$\mu_{\Delta t}$, $\tau_{\Delta t}$, and $\nu_{\Delta t}$ for the
time difference and $\mu_{\Delta e}$, $\tau_{\Delta e}$, and $\nu_{\Delta e}$
for the difference in the number of errors. 

\begin{figure}
\begin{centering}
\includegraphics[width=1\columnwidth]{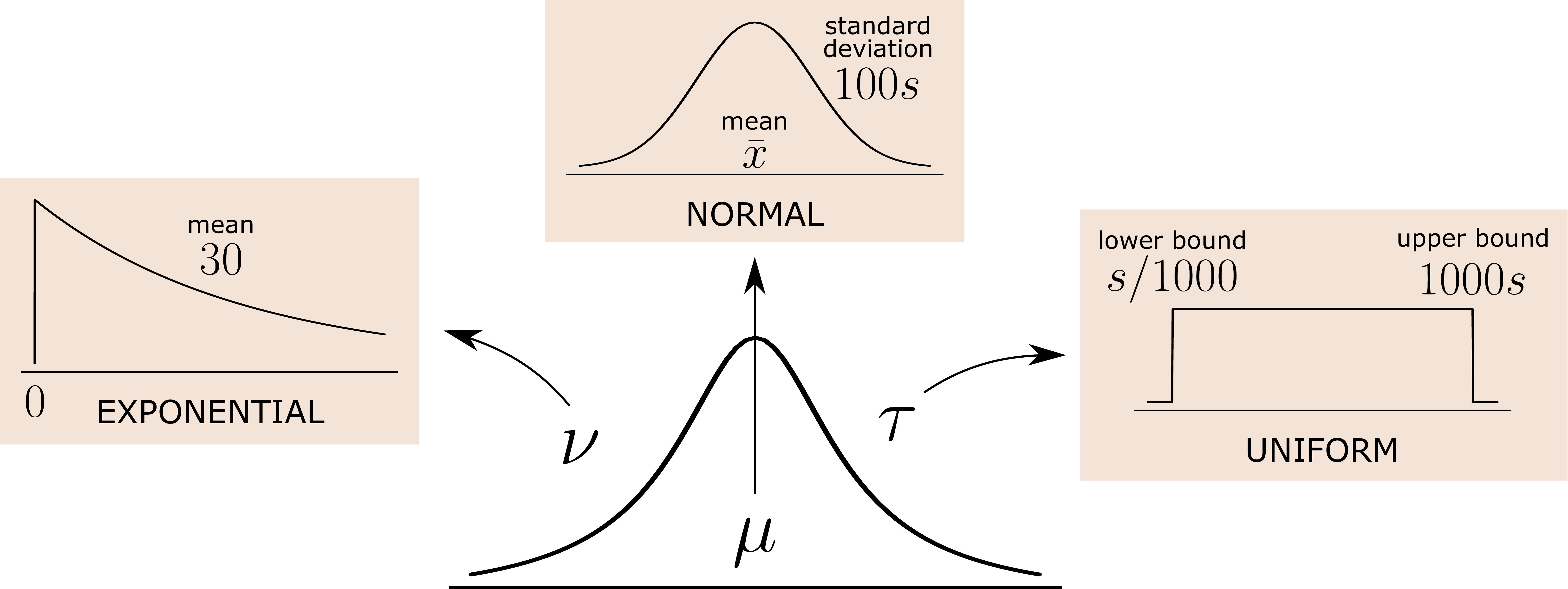}
\par\end{centering}
\caption{\label{fig:obj-priors}Diagram of the Bayesian estimation for the
metric variables $\Delta t$ and $\Delta e$. A \textit{t} distribution
describes the data and we estimate the parameters $\mu$, $\tau$,
and $\nu$ of each variable, using the indicated priors, where $\bar{x}$
and $s$ are the sample mean and standard deviation, respectively.}
\end{figure}

Since we do not have previous information about the parameters, all
priors are broad and vague to minimally influence the estimation (\textit{e.g.},
avoid biasing). Therefore, both for $\Delta t$ and $\Delta e$, the
prior distributions for the mean and scale parameters are normal and
uniform distributions \citep{Kruschke2015}, respectively. When $\nu>30$,
the \textit{t} distribution closely approximates a normal. Hence,
large differences between the normal and $t$ distributions occur
when $\nu$ is small (see Fig.~\ref{fig:t-distribution}), which
is considered credible in the posterior distribution only if smaller
values for $\nu$ are more credible in the prior distribution, or
if the sample contains a lot of outliers, rare by definition \citep{Kruschke2015}.
As a consequence, we use an exponential with mean of $30$ as a prior
distribution for the normality parameters to consider small values
in the estimation while still allowing high values. The parameters
and their priors are illustrated in Fig.~\ref{fig:obj-priors}.

\subsection{\label{subsec:Subjective-measures}Subjective measures}

We obtain the subjective measures through questionnaires with Likert
scales (see Section~\ref{subsec:Measurements}). Following Likert's
original work \citep{Likert1932}, one frequently used way to deal
with this type of data is using the average or the sum of the points
of the items in a scale, and then treating this value as an observation
from each participant. There is a discussion in the literature on
whether this data set generated from the average or sum of points
should be treated as interval or ordinal measures and which tests
apply to them.\footnote{Some references that discuss the subject, especially in the context
of frequentist analyses (for example, comparing the \emph{t} and Wilcoxon
signed-rank tests) are \citep{Nanna1998,Meek2007,Carifio2008,Harpe2015}.} Liddell and Kruschke \citep{Liddell2018} show that treating ordinal
data from a single item as metric leads to systematic errors of false
positives, failures in detecting effects, and even the inversion of
an effect. They also show that using the average points from a set
of items does not solve the problems. Since there is no consensus
in the literature, we treat data from subjective measures as ordinal.

Kruschke suggests a cumulative normal model (see Chapter 23 of \citep{Kruschke2015})
for the analysis of ordinal data from a single item, and Liddell and
Kruschke \citep{Liddell2018} extend the model to multiple items.
They treat items together but without aggregating points into a single
measure. The idea is that the measured variable is in a continuous
metric scale but cannot be accessed directly; that is, it is a latent
variable. Therefore, the set of items in the Likert scale is a way
of accessing the latent variable through a discrete and ordinal response
scale. As in the metric model, using a \textit{t} latent distribution
instead of a normal makes the model more robust to outliers.

\begin{figure}
\begin{centering}
\includegraphics[width=1\columnwidth]{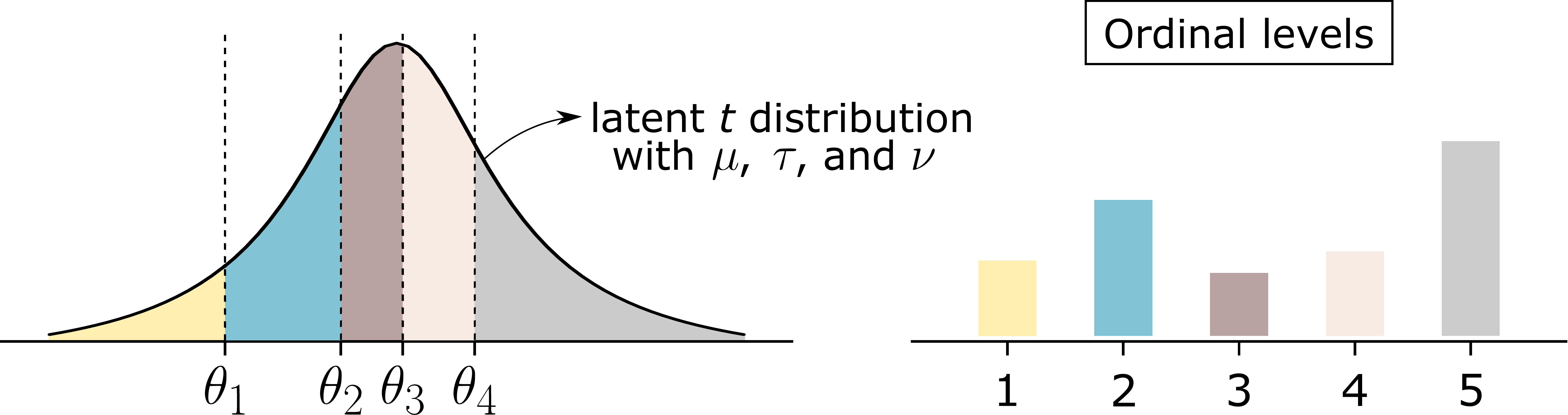}
\par\end{centering}
\caption{\label{fig:ordinal}The probability of the ordinal response $y=k$
is given by the cumulative probability between the thresholds $\theta_{k-1}$
and $\theta_{k}$ on the latent \textit{t} distribution of mean $\mu$,
scale $\tau$, and normality parameter $\nu$, with $k\in\{1,\ldots,K\}$,
$\theta_{0}=-\infty$ and $\theta_{K}=\infty$.}
\end{figure}

For a single item with $K\in\mathbb{N}$ response levels, thresholds
$\theta_{1},\ldots,\theta_{K-1}$ divide the latent distribution into
$K$ intervals, as shown in Fig.~\ref{fig:ordinal} (for $K=5$).
On the ordinal model, the probability assigned to each response level
is the cumulative probability of each interval, calculated as the
area under the latent distribution between the respective thresholds,
or between the outer thresholds ($\theta_{1}$ and $\theta_{K-1}$)
and open boundaries at $-\infty$ and $\infty$ (\textit{i.e.}, $\theta_{0}\triangleq-\infty$
and $\theta_{K}\triangleq\infty$). For a latent \textit{t} distribution
with mean $\mu$, scale $\tau$, and normality parameter $\nu$, the
probability of the ordinal response $y=k$, with $k\in\{1,\ldots,K\}$,
is
\begin{multline}
P(y=k\mid\mu,\tau,\nu,\theta_{1},\dots,\theta_{K-1})=\\
\Psi_{\mu,\tau,\nu}\left(\theta_{k}\right)-\Psi_{\mu,\tau,\nu}\left(\theta_{k-1}\right),\label{eq:p(y=00003Dk)}
\end{multline}
where $\Psi_{\mu,\tau,\nu}(u)=\int^{u}_{-\infty}f_{\mu,\tau,\nu}(x)dx$
is the cumulative \textit{t} function with
\begin{multline*}
f_{\mu,\tau,\nu}(x)=\\
\frac{\varGamma\left(\frac{\nu+1}{2}\right)}{\varGamma\left(\frac{\nu}{2}\right)}\left(\frac{1}{\tau^{2}\nu\pi}\right)^{\frac{1}{2}}\left[1+\frac{(x-\mu)^{2}}{\tau^{2}\nu}\right]^{-\frac{(\nu+1)}{2}}
\end{multline*}
and $\varGamma$ represents the gamma function $\varGamma(w)=\int^{\infty}_{0}z^{w-1}e^{-z}dz$
with $\real w>0$ \citep[p. 501-507]{Jackman2009}. The model represented
by Eq.~\ref{eq:p(y=00003Dk)} applies to all ordinal levels since
$\Psi_{\mu,\tau,\nu}(-\infty)=0$ and $\Psi_{\mu,\tau,\nu}(\infty)=1$.

\begin{figure}
\begin{centering}
\includegraphics[width=1\columnwidth]{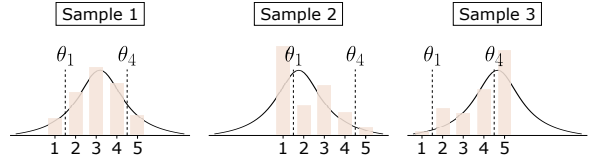}
\par\end{centering}
\caption{\label{fig:fixed-thresholds}Examples of sample histograms of ordinal
responses with five levels and possible adequate latent distributions.
Extreme thresholds are fixed at $\theta_{1}=1.5$ and $\theta_{4}=4.5$.}
\end{figure}

The model has $K+2$ parameters: the latent variables $\mu$, $\tau$,
and $\nu$ and the $K-1$ thresholds that map the latent variable
into the ordinal responses. There are infinite possible combinations
for these parameters that result in the same ordinal probabilities,
since we can ``drag,'' ``compress,'' or ``expand'' the distribution,
by changing the whole set of parameters (see Fig.~\ref{fig:ordinal}),
while keeping the probabilities associated with each level. To solve
this problem, Kruschke \citep{Kruschke2015} suggests fixing the extreme
thresholds, $\theta_{1}$ and $\theta_{K-1}$, at meaningful values
with respect to the response scale, specifically $\theta_{1}=1.5$
and $\theta_{K-1}=K-0.5$, so these fixed values anchor the estimation.
By doing it, the estimated parameters are interpreted according to
the meaning of the response options. Suppose we ask people to answer
an item stating ``I like robots'' using a scale with five response
levels, from ``totally disagree'' to ``totally agree'' and with a
middle answer saying ``I do not know.'' Fig.~\ref{fig:fixed-thresholds}
shows examples of histograms from three possible samples and possible
adequate latent distributions to each one of them, with the extreme
thresholds fixed at $1.5$ and $4.5$. For Sample~1, the answers
concentrate in the middle of the ordinal scale so the latent $\mu$
would be approximately $3$, suggesting that, on average, people are
not certain if they like robots or not. For Sample~2, negative answers
are more frequent and the latent $\mu$ would be smaller, indicating
that, on average, people do not like robots. Finally, the results
from Sample~3 suggest that people do like robots, with $\mu$ having
a larger value, since the higher response levels are more frequent.

Now, suppose we want to compare a variable in two different conditions
(or groups), such as the acceptance of the virtual agent in EX and
EXIM communication configurations. When measuring the same latent
variable (\textit{e.g.}, acceptance) using the same questionnaire
such as the Likert scale shown in Table~\ref{tab:likert_scales},
we assume that the latent variable for all groups have the same probability
density function but with different parameters. Since the thresholds
are related to the way we measure the variable (\textit{i.e.}, the
sentence in the Likert scale, such as ``I found the virtual agent
friendly''), we assume they are the same across all groups. Therefore,
what differs between groups is how much people agree or disagree with
the sentence and the variance of this feeling. For each group, we
consider that there are common latent $\mu$, $\tau$, and $\nu$
for all items of the scale, but a different set of thresholds for
each item, since they access the same latent variable in different
ways \citep{Liddell2018}.

After fixing the outer thresholds of a single item on each scale in
$\theta_{1}=1.5$ and $\theta_{K-1}=K-0.5$, as suggested by Liddell
and Kruschke, we need to estimate the remaining parameters. For multiple
items and multiple groups, the probability of each ordinal response
$y^{[i]}_{g}$ of the $i$th item and $g$th group is given by \citep{Liddell2018}
\begin{multline}
P(y^{[i]}_{g}=k\mid\mu_{g},\tau_{g},\nu_{g},\theta^{[i]}_{1},\dots,\theta^{[i]}_{K-1})\\
=\Psi_{\mu_{g},\tau_{g},\nu_{g}}\left(\theta^{[i]}_{k}\right)-\Psi_{\mu_{g},\tau_{g},\nu_{g}}\left(\theta^{[i]}_{k-1}\right),\label{eq:p(y_i=00003Dk)}
\end{multline}
where $\theta^{[i]}_{k}$ is the $k$th threshold of the $i$th item
(\textit{e.g.}, ``I found the virtual agent friendly''), and $\mu_{g}$,
$\tau_{g}$, and $\nu_{g}$ are the mean, the scale, and the normality
parameter of the latent variable (\textit{e.g.,} acceptance) in group
$g$ (\textit{e.g.}, EXIM).

The model states that the ordinal response $y^{[i]}_{g}$ comes from
a categorical distribution with probabilities given by Eq.~\ref{eq:p(y_i=00003Dk)}.
As mentioned before, we fix the outer thresholds only of the first
item of each scale in Table~\ref{tab:likert_scales}.\footnote{We analyse the subjective measures considering two separate groups,
instead of using the difference as we do for time and errors. This
is to maintain the meaning of the fixed thresholds and not to generate
more empty response levels in the sample data, since they cause negative
probabilities, as discussed at the end of this section and in Appendix~\ref{sec:limitations}.} Therefore, the goal of the Bayesian inference is to estimate the
parameters $\mu_{\mathrm{EX}}$, $\mu_{\mathrm{EXIM}}$, $\tau_{\mathrm{EX}}$,
$\tau_{\mathrm{EXIM}}$, $\nu_{\mathrm{EX}}$, and $\nu_{\mathrm{EXIM}}$
of each variable (acceptance, sociability, transparency, and perceived
efficiency) and the unfixed thresholds (\textit{i.e.}, $\left(\bigcup^{n_{i}}_{i=1}\left\{ \theta^{[i]}_{1},\ldots,\theta^{[i]}_{K-1}\right\} \right)\setminus\{\theta^{[1]}_{1},\theta^{[1]}_{K-1}\}$
with $n_{i}$ being the number of items) associated with the items
of each scale. After the estimation, we analyse the difference between
the groups EX and EXIM.

\begin{figure}
\begin{centering}
\includegraphics[width=1\columnwidth]{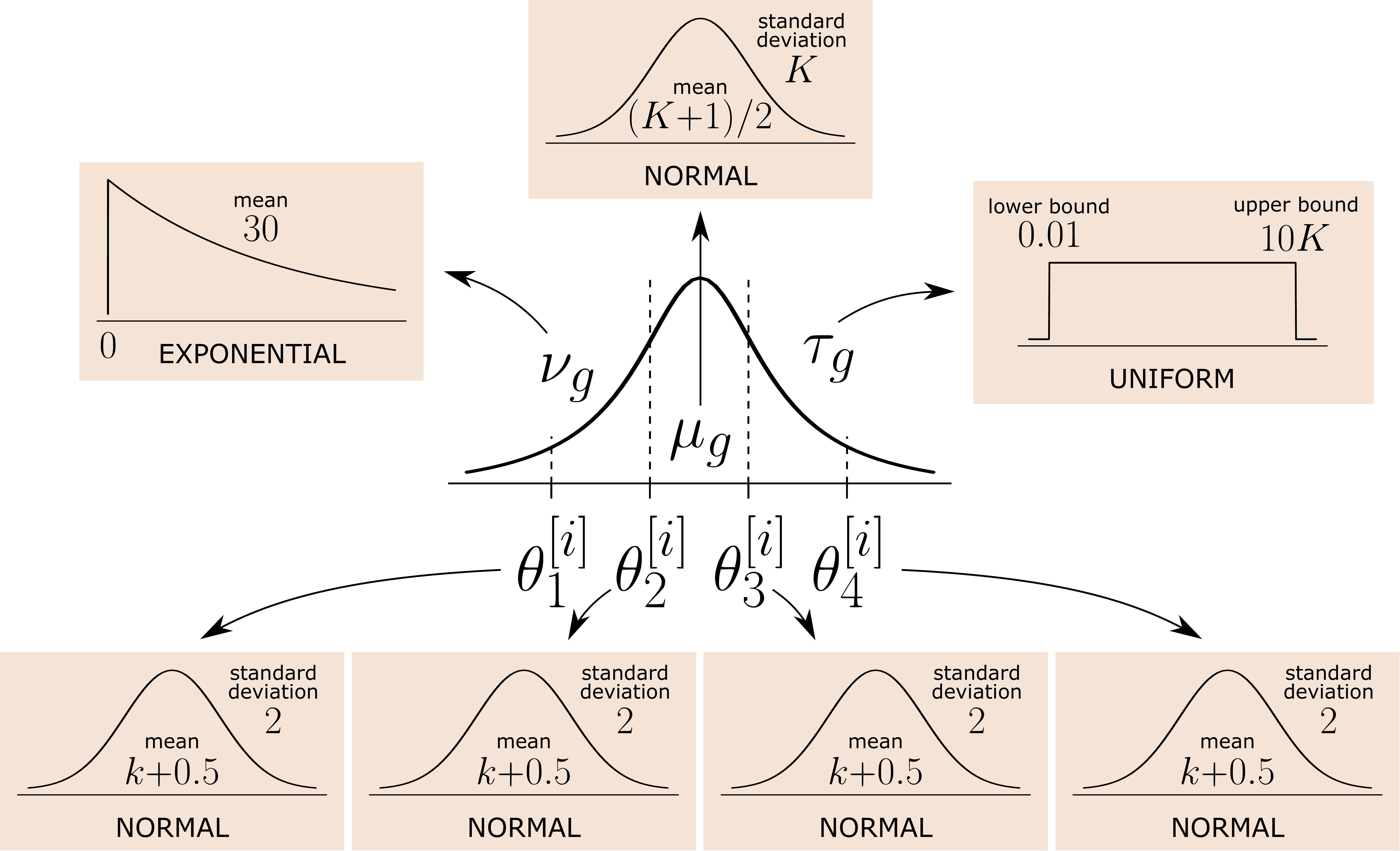}
\par\end{centering}
\caption{\label{fig:subj-priors}Bayesian estimation for the ordinal variables.
A \textit{t} distribution describes the latent variable and we estimate
its parameters $\mu_{g}$, $\tau_{g}$, and $\nu_{g}$ and the free
thresholds $\theta^{[i]}_{k}$, $k\in\{1,2,3,4\}$, translating the
latent variable into the ordinal responses of each item $i$ in the
Likert scales. The diagram shows the prior distributions for each
parameter, where $K=5$ is the number of ordinal response levels in
our scales.}
\end{figure}

Liddell and Kruschke \citep{Liddell2018} suggest using the priors
shown in Fig.~\ref{fig:subj-priors} for the parameters, with $\mu_{g}$
and $\tau_{g}$ in the neighbourhood of the data, whereas the free
thresholds follow normal distributions with considerable standard
deviation.

There is nothing in the model to specify that the thresholds are in
ascending order, \textit{i.e.}, $\theta_{1}<\theta_{2}<\dots<\theta_{K-1}$.
Therefore, if $\theta_{k-1}>\theta_{k}$, the probability of the ordinal
level $k$ is negative (see Eq.~\ref{eq:p(y_i=00003Dk)}), which
violates the first probability axiom. Kruschke works around this limitation
through implementation, by considering only non-negative probabilities.
However, his solution only works if the data sample has at least one
answer in every level $k$, which can be difficult to obtain with
small samples. So, together with Kruschke's proposed implementation,
we decided to add one extra observation to each empty level we encounter
in our samples, and use the updated data set in the Bayesian inference.
In our tests with simulated data, when we added the extra observations,
we observed that the estimations of the scale parameter $\tau$ gave
more credibility to values greater than the real ones. This increase
in the values of $\tau$ makes it more difficult to validate our work
hypotheses, since the effect size we calculate has $\tau$ in the
denominator, as described in the next section. \emph{Therefore, our
strategy to mitigate empty levels due to small samples is conservative}.
More details about our tests can be found in Appendix~\ref{sec:limitations}.

\section{Results\label{sec:Results}}

\begin{figure*}
\begin{centering}
\includegraphics[width=1\textwidth]{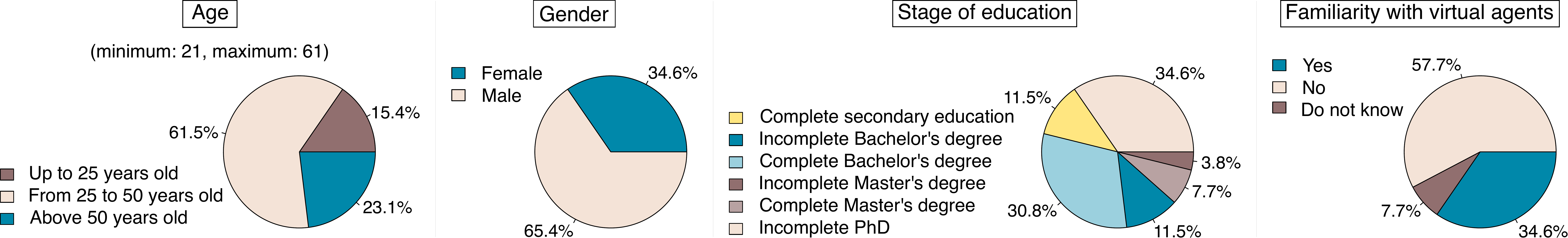}
\par\end{centering}
\caption{\label{fig:sample-summary}Summary of the profile of the interaction
experiment participants.}
\end{figure*}

Thirty volunteers participated in the experiments but four were excluded
due to significant deviation from the experimental protocol. Volunteer
4 did not finish interacting with the second virtual agent due to
a technical problem, volunteer 9 did not complete the questionnaire
after the first interaction, volunteer 10 asked for the researchers'
help during the task, and volunteer 22 completed the task atypically,
hindering the objective measures. Therefore, the final sample size
is of 26 participants, summarised in Fig.~\ref{fig:sample-summary}.

\subsection{\label{subsec:Bayesian-analysis-results}Bayesian analysis results}

Effect size is a measure quantifying the strength of the presence
of an effect. It is calculated considering the null value, which represents
an absence of effect \citep{Cohen1988}. We calculate the effect size
$d_{\mathrm{obj}}$ for the objective measures as
\begin{equation}
d_{\mathrm{obj}}=\frac{(\mu-\mu_{0})}{\tau},\label{eq:d_obj}
\end{equation}
considering the null value $\mu_{0}=0$ (absence of an effect) and
using the estimated mean $\mu$ and scale $\tau$ of the difference
between the configurations $\Delta t$ and $\Delta e$, as explained
in Section~\ref{subsec:Objective-measures}. Since the error and
time differences are defined as in Eq.~\ref{eq:time and error differences},
positive effect sizes favour hypotheses H1 and H2, whereas negative
effect sizes go against them. For the analysis of subjective measures,
which estimates the parameters of each condition (\textit{i.e.}, EX
and EXIM) separately, the calculated effect size is
\begin{align}
d_{\mathrm{sub}} & =\frac{\left(\mu_{\mathrm{EXIM}}-\mu_{\mathrm{EX}}\right)}{\sqrt{0.5\left(\tau^{2}_{\mathrm{EX}}+\tau^{2}_{\mathrm{EXIM}}\right)}},\label{eq:d_sub}
\end{align}
using the estimated mean and scale parameter of each group (EX e EXIM)
\citep{Kruschke2013,Kruschke2015}. Therefore, positive effect sizes
also favour hypotheses H3 to H6, whereas negative effect sizes go
against those hypotheses.

We define a ROPE from $-0.1$ to $0.1$ around the null value ($d_{\mathrm{obj}}=d_{\mathrm{sub}}=0$)
in the effect size posterior. This interval covers values up to half
of a small effect size, according to Cohen's convention \citep{Cohen1988},
which is used because we do not have a clear understanding yet of
what a significant effect size means in our context.

\subsubsection{Implementation details}

We generated the posterior distributions using Markov Chain Monte
Carlo (MCMC) methods and JAGS (Just Another Gibbs Sampler) system.\footnote{For more details, check Chapters 7 and 8 of \citep{Kruschke2015}.}
All scripts were written using R language and based on examples provided
in the works by Kruschke \citep{Kruschke2015} and Liddell \& Kruschke
\citep{Liddell2018}.

The MCMC sample contains a large number of parameter combinations,
allowing the generation of posterior distributions for each parameter
and other distributions such as the difference between parameters
in each group and the effect size, calculated using Eqs.~\ref{eq:d_obj}~and~\ref{eq:d_sub}.

\subsubsection{Objective measures}

For the sake of conciseness, we only show the distributions of the
effect size for each measure. For the posterior distributions of mean
$\mu$, scale $\tau$, and normality $\nu$, please refer to the Supplementary
Material accompanying the paper. The Supplementary Material also shows
some credible \textit{t} distributions superimposed on the data of
each variable to check model adequacy. As there is no critical deviation
(\textit{e.g.}, strong asymmetry or multimodal distribution) between
the data and the estimated \textit{t} distributions, we conclude that
the estimations fit the data well enough and the chosen model is adequate.

\begin{figure*}
\begin{centering}
\begin{minipage}[t]{0.32\textwidth}%
\begin{center}
\subfloat[\label{fig:results-time}Posterior distribution of the effect size
in the time difference $\Delta t$ (seconds).]{\centering{}%
\noindent\begin{minipage}[t]{1\columnwidth}%
\begin{center}
\includegraphics[height=0.75\columnwidth]{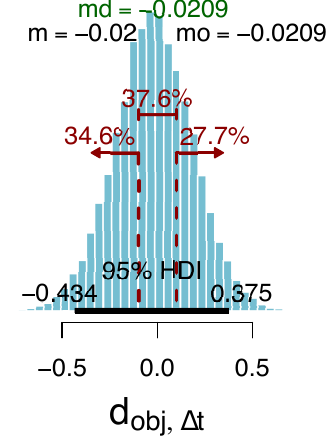}
\par\end{center}%
\end{minipage}}
\par\end{center}%
\end{minipage}~~%
\begin{minipage}[t]{0.32\textwidth}%
\begin{center}
\subfloat[\label{fig:results-errors}Posterior distribution of the effect size
in the difference in number of errors $\Delta e$.]{\centering{}%
\noindent\begin{minipage}[t]{1\columnwidth}%
\begin{center}
\includegraphics[height=0.75\columnwidth]{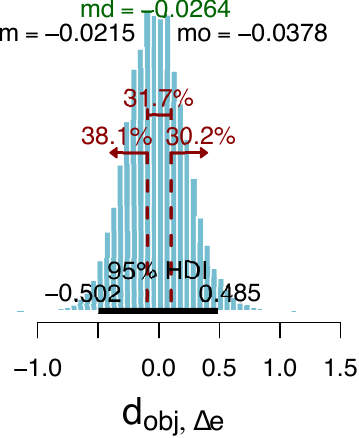}
\par\end{center}%
\end{minipage}}
\par\end{center}%
\end{minipage}~~%
\begin{minipage}[t]{0.32\textwidth}%
\begin{center}
\subfloat[\label{fig:results-errors-without}Posterior distribution of the effect
size in the difference in number of errors $\Delta e'$ without outliers.]{\centering{}%
\noindent\begin{minipage}[t]{1\columnwidth}%
\begin{center}
\includegraphics[height=0.75\columnwidth]{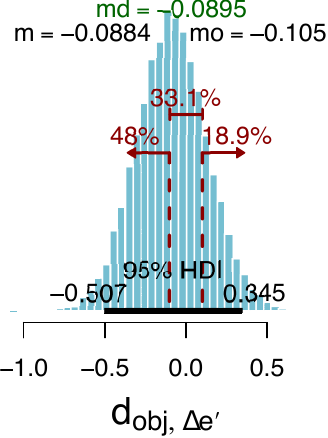}
\par\end{center}%
\end{minipage}}
\par\end{center}%
\end{minipage}
\par\end{centering}
\caption{\label{fig:results-objective}Results of the Bayesian inference for
the objective measures (time and number of errors). The figures show
the distributions of the effect size $d_{\mathrm{obj}}$ (Eq.~\ref{eq:d_obj})
calculated with the null value $\mu_{0}=0$. Mean (\textit{m}), median
(\textit{md}), mode (\textit{mo}), and the limits of the $95\%$ HDI
are annotated. Dashed vertical lines indicate the ROPE, together with
the percentages of the distribution below, between and above it.}
\end{figure*}

\paragraph{Time}

Fig.~\ref{fig:results-time} shows the distribution of the effect
size for the time difference $\Delta t$, in seconds, between the
two communication configurations. The distribution is centred close
to the null value, but it has a large $95\%$ HDI, including almost
medium positive and negative effect sizes (\textit{i.e.}, $d_{\mathrm{obj}}=0.5$
\citep{Cohen1988}). Positive effect sizes would favour hypothesis
H1, whereas negative effect sizes would go against it. Therefore,
this estimation does not allow us to reach a conclusion using the
decision rule illustrated in Fig.~\ref{fig:decision-rope} about
the time difference between the two communication configurations.

\paragraph{Error}

Owing to two possible outliers ($\Delta e=-11$ and $\Delta e=26$),\footnote{These two cases seem to have occurred because the participants did
not understand that the first colour shown on the screen (black for
one participant, white for the other) was already part of the password
and kept indicating the next color repeatedly, causing multiple errors
to be registered.} we have made the analysis for the difference in the number of errors
with and without them. The effect size distributions are shown in
Figs.~\ref{fig:results-errors}~and~\ref{fig:results-errors-without}.
With the outliers, more credibility is given for small values of the
normality parameter $\nu$, increasing the weight in the tails of
the latent \textit{t} distribution to try to accommodate the outliers.
Without them, the estimations of the mean and the scale parameter
became more precise (narrower $95\%$ HDI) and the effect size posterior
is slightly ``compressed'' to the left, reducing the percentage of
the distribution above the ROPE upper limit, and giving less credibility
to values favourable to our hypothesis. However, again we do not have
enough precision to draw strong conclusions about the existence or
not of difference in the number of errors between the two communication
configurations.

\subsubsection{Subjective measures}

\begin{figure*}
\begin{centering}
\begin{minipage}[t]{0.5\textwidth}%
\begin{center}
\subfloat[\label{fig:results-acceptance}Posterior distribution of the effect
size $d_{\mathrm{sub,acc}}$ in the acceptance of the virtual agents.]{\begin{centering}
\includegraphics[height=0.5\columnwidth]{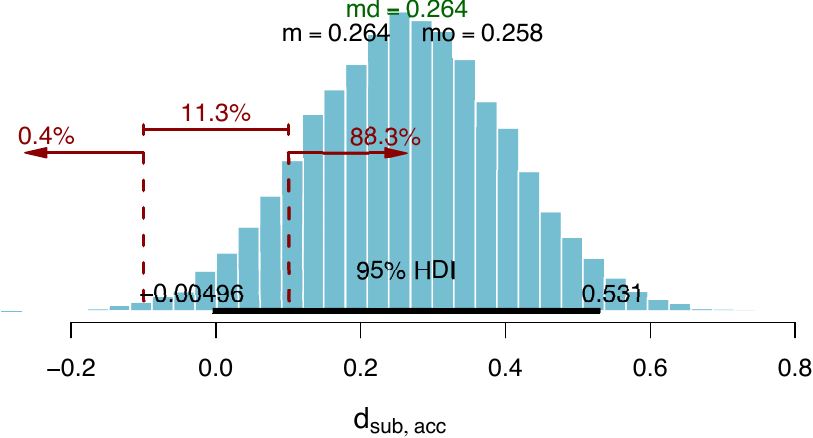}
\par\end{centering}
\centering{}}
\par\end{center}%
\end{minipage}%
\begin{minipage}[t]{0.5\textwidth}%
\begin{center}
\subfloat[\label{fig:results-sociability}Posterior distribution of the effect
size $d_{\mathrm{sub,soc}}$ in the sociability of the virtual agents.]{\begin{centering}
\includegraphics[height=0.5\columnwidth]{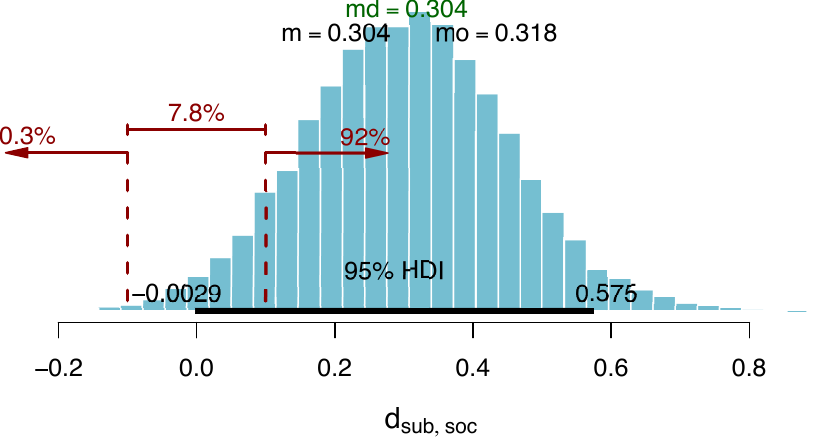}
\par\end{centering}
}
\par\end{center}%
\end{minipage}
\par\end{centering}
\begin{centering}
\begin{minipage}[t]{0.5\textwidth}%
\begin{center}
\subfloat[\label{fig:results-transparency}Posterior distribution of the effect
size $d_{\mathrm{sub,tra}}$ in the transparency of the virtual agents.]{\begin{centering}
\includegraphics[height=0.5\columnwidth]{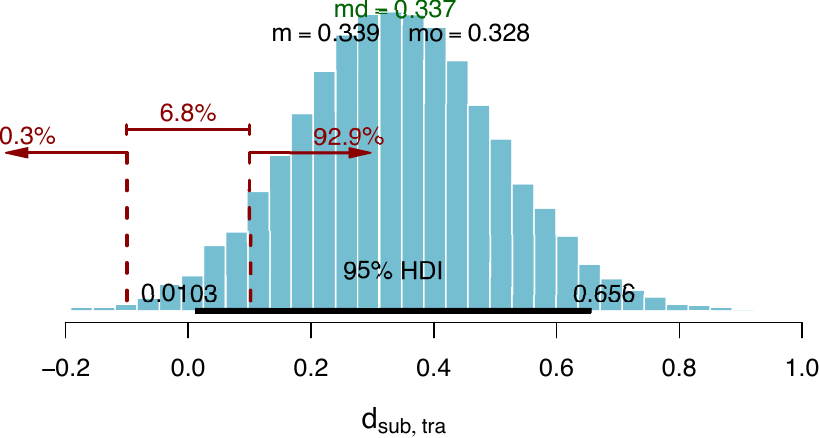}
\par\end{centering}
}
\par\end{center}%
\end{minipage}%
\begin{minipage}[t]{0.5\textwidth}%
\begin{center}
\subfloat[\label{fig:results-efficiency}Posterior distribution of the effect
size $d_{\mathrm{sub,eff}}$ in the perceived efficiency of the interactions.]{\begin{centering}
\includegraphics[height=0.5\columnwidth]{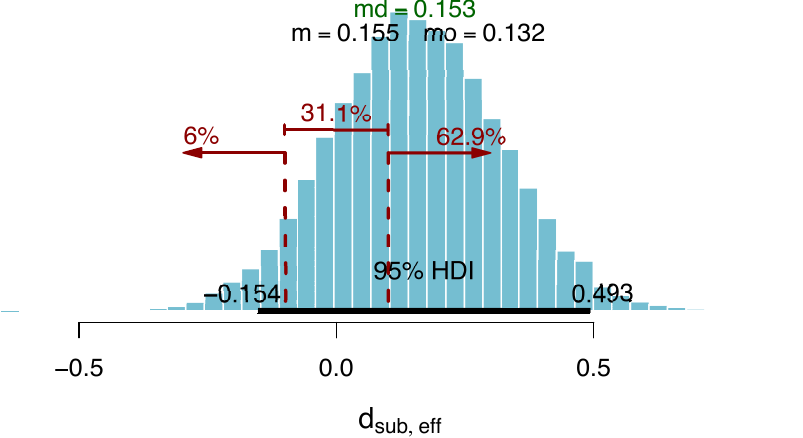}
\par\end{centering}
}
\par\end{center}%
\end{minipage}
\par\end{centering}
\caption{\label{fig:results-subjective}Results of the Bayesian estimation
of acceptance, sociability, transparency and perceived efficiency
in the two communication configurations. The figures show the distribution
of the effect sizes. Mean (\textit{m}), median (\textit{md}), mode
(\textit{mo}), and the limits of the $95\%$ HDI are annotated. Dashed
vertical lines indicate the ROPE, together with the percentages of
the distribution below, between and above it.}
\end{figure*}

We obtained posterior distributions for the latent parameters of each
group separately and then generated posteriors for the difference
between the means and the scale parameters of each group and the effect
size. Positive effect sizes calculated using Eq.~\ref{eq:d_sub}
favour our hypotheses, and their distributions are shown in Fig.~\ref{fig:results-subjective}.
Other distributions, including the estimations of the thresholds $\theta_{k}$,
and comparisons between the data and the estimations are available
in the Supplementary Material accompanying the paper. As the estimations
fit the data appropriately, we conclude the model is adequate.

\paragraph{Acceptance}

When completing the questionnaire with the Likert scales for the acceptance
of the virtual agent, no participants chose ordinal level 2 for item
4 in the EX group. Furthermore, no participants from the EXIM group
chose ordinal level 1 for items 1--3 and 5, ordinal level 2 for items
1, 2, and 4, and ordinal level 3 for item 5. Therefore, those levels
were empty in the data set and we added one extra answer in the EX
group (ordinal level 2) and eight in EXIM group (ordinal levels 1,
2, and 3) to avoid negative probabilities, as explained in Section~\ref{subsec:Subjective-measures}.
Thus, the estimated scales $\tau_{\mathrm{acc}}$ might be slightly
greater than the real ones, especially for the EXIM group, and the
effect size slightly lower (see Appendix~\ref{sec:limitations} for
more information). The Supplementary Material shows data histograms
indicating all the levels for which we added extra answers.

Fig.~\ref{fig:results-acceptance} shows the effect size posterior
of the acceptance of the virtual agents, with its median ($\mathrm{md}=0.264$)
indicating small to medium effects (\textit{i.e.}, $0.2$ to $0.5$
\citep{Cohen1988}), but without enough precision to draw a conclusion
using the $95\%$ HDI and ROPE.\footnote{The width of the 95\% HDI of the subjective variables is smaller than
the 95\% HDI of the objective variables. Consequently, the estimation
of subjective variables is more precise. This is because we assume
in the ordinal model that the latent parameters are the same for all
$Q\in\left\{ 4,5,6\right\} $ items from the Likert scales (\emph{i.e.},
each item measures the same phenomenon). Consequently, we use all
$26Q$ observations related to all 26 participants to estimate $\mu$,
$\tau$, and $\nu$, resulting in more precise estimations.} However, $88.3\%$ of the distribution is above the ROPE upper limit,
suggesting high credibility that there is an effect favourable to
our hypotheses;\textit{ }\textit{\emph{namely}}, that the EXIM virtual
agent is more accepted than the EX one.

\paragraph{Sociability}

For the sociability data set, we added one extra answer only to level
1 of item 1 in group EXIM, as no participant chose that response.
Again, the $95\%$ HDI of the effect size posterior, shown in Fig.~\ref{fig:results-sociability},
is not narrow enough to allow us a strong conclusion, but $92\%$
of the distribution is above the ROPE upper limit, suggesting that
the EXIM virtual agent was perceived as more sociable than the one
from the EX configuration.

\paragraph{Transparency}

We added one extra answer on level 1 of each group to overcome the
lack of participant responses in this level in item 1 for EX and item
3 for EXIM. The effect size estimation, shown in Fig.~\ref{fig:results-transparency},
is once again not precise to fulfil Kruschke's decision criterion,
but it indicates that the EXIM virtual agent might have been seen
as more transparent than the EX virtual agent, with $92.9\%$ of the
distribution above the ROPE upper limit.

\paragraph{Perceived efficiency}

Finally, for the perceived efficiency dataset, we added one extra
answer in level 2 of item 3 of EX group and one in level 1 of item
3 of group EXIM because those were empty due to the lack of response.
Fig.~\ref{fig:results-efficiency} shows the effect size posterior,
whose median ($\mathrm{md}=0.153$) indicates a less than small effect
(\textit{i.e.}, lower than $0.2$ \citep{Cohen1988}), with $62.9\%$
of the distribution above the ROPE upper limit.

\section{Discussions}

\subsection{Dealing with technical errors during the experiments}

The system for human kinematic chain recognition \citep{Campos2020}
sometimes failed to detect the participant and the experimenter intervened
to give additional instructions or to restart the system, sometimes
remotely. Also, some participants did not understand that they would
interact with two virtual agents and left the room after the first
interaction and questionnaire. In these cases, the experimenter asked
them to go back and continue. As long as these interventions did not
happen during the task execution and interrupted the interaction flow,
we took note of the occurrence and let the experiment continue and
the participant was not excluded from the analysis. We excluded the
system initialisation times and the time to solve the aforementioned
technical problems in our analyses.

All participations were recorded with the participants' knowledge
and formal consent. After the experiments, we watched the videos and
adjusted the data. For instance, for Phase 1, we discarded errors
caused by wrong detection of pointing gestures and some delay in the
sound signals indicating a correct or wrong colour (sometimes, a delay
happened and the participant kept pointing while waiting for the sound
signal, so the system counted two gestures instead of one). Errors
indirectly attributed to the system limitations, such as when a participant
points to a second object after indicating the correct one, but the
system fails to recognise it, were not discarded because these interpretations
were subjective; therefore, we decided to follow a more conservative
approach.

\subsection{\label{subsec:General-discussion}General discussion}

The results shown in Section~\ref{sec:Results} were obtained from
a sample of 26 participants. From the four participants excluded from
this analysis, three completed all the steps of the interaction and
the questionnaires, so they are included in the following discussion,
which aims to discuss the experimental protocol and how it could be
improved.

From the 29 participants that interacted with both virtual agents,
21 of them said they preferred to interact with the EXIM virtual agent,
which combined explicit and implicit communications. Participants
smiled at or talked to the virtual agents during the interactions
and most comments made about them were positive, either on the questionnaires
or to the researcher conducting the experiment. People seemed to have
positive reactions to them, which was also observed in the acceptance
analysis. One participant commented
\begin{quotation}
``\textit{I found the interaction very interesting, and, specially
after interacting with Luna }{[}the EXIM virtual agent{]}\textit{,
I noticed that the simple fact of the virtual agent to }`\textit{look}'\textit{
at my direction made a difference on how I felt with respect to the
task.}''
\end{quotation}
Another person commented that ``\textit{the voice is irritant}'' and
another one said to the experimenter that they did not like ``\textit{these
virtual agents}''.\footnote{All comments were translated from Portuguese.}

Acceptance, sociability, and transparency are variables more related
to the virtual agents, whereas time, errors, and perceived efficiency
of the interaction depend more directly on the task. Seven comments
on questionnaires reported difficulties in understanding the task,
which was also mentioned by other participants directly to the experimenter,
suggesting that instructions might not have been clear enough. From
the final sample of 26 participants, extra errors occurred due to
failure to complete the task before the timeout, which accounted for
$7.7\%$ of the errors in Phase 1 and $16.1\%$ in Phase 2, indicating
difficulties in completing the tasks. The videos also suggest that
people found Phase 2 more difficult, taking a long time to find the
counting images fixed in the environment and to understand what to
do. Six participants added what seems to be generic values, such as
1 for all objects, in at least one of the configurations, suggesting
that they did not understand the task or did not find the images.
Two people mentioned that the space used for the experiment was visually
cluttered, which might have created difficulties for participants
to find the counting images. The objective measures might have been
influenced by these factors.

On the EXIM configuration, we used implicit communications not only
to make Luna and Sofia more pleasant, sociable, and transparent. Indeed,
we also hoped they would help participants during the task execution,
since the virtual agents looked at the correct object they believe
people would point to in Phase 1. Moreover, they used people's gaze
to estimate their attention focus and give hints with the correct
answers in Phase 2. In fact, based on our interpretation of the experiment
recordings, we believe that at least four people might not have seen
the counting images and added correct values only trusting the information
provided by the virtual agent. Even when the system detected a wrong
gaze direction, the virtual agent's own gaze complemented the communication
and the participant could infer which object it was referring to.
Four other people neither understood nor considered the virtual agent's
hint and added wrong values despite being prompted with the correct
answer.

In Phase 1, the videos suggest that some participants might not have
seen the virtual agents' implicit communications, as they did not
seem to look at the screen displaying the virtual agent while executing
the task. On the other hand, other people clearly noticed that the
virtual agent looked at them because they played around with its gaze
for a moment, moving their bodies to see the virtual agent following
them. People may also not have attributed meaning to its gaze, seeing
it but not interpreting it. One participant seemed not to have understood
that one colour was from the password and, after the interaction,
told the experimenter that the virtual agent kept looking the other
way, without realising it was looking at the correct object. After
all, it was not necessarily a collaborative task, meaning that it
did not need to be done collaboratively, although the virtual agent
could help. We believe that these aspects may have influenced the
perceived efficiency of the interaction, with some people attributing
little or no credit to the virtual agent for the completion of the
tasks.

In summary, variables more related to the task, such as time, number
of errors, and perceived efficiency of the interaction, did not seem
to be affected by the communication type. This may be attributed to
the fact that the tasks were not necessarily collaborative and perhaps
too simple, making external help unnecessary for their successful
conclusion. We decided to investigate this further in a follow-up
study, which is presented in the next section.

\section{\label{sec:Follow-up-study}Follow-up study}

Our results differ from the work by Breazeal \textit{et al.} \citep{Breazeal2005a},
where they observed effects on performance measures in a task where
people guided the robot to push some buttons. In their scenario, the
task was designed in a way that it was mandatory for the human and
the robot to work together to complete the task. In contrast, we have
designed tasks in which the robot or virtual agent behave as companions,
such that the human could finish the task without help and the artificial
agents' communications could be disregarded. One of the main reasons
behind our design choice was that, when artificial agents are optional
for the successful task completion, humans will tend to use them only
if they add some value. However, as shown in Section~\ref{sec:Results},
task-related metrics were not affected by the communication type,
which induced a new hypothesis regarding the role of the task difficulty
in our study. Therefore, we conducted a follow-up study to investigate
the impact of the task difficulty in the comparison of different communication
configurations.

For this study, we hypothesise the following:
\begin{lyxlist}{00.00.0000}
\item [{\textit{H7:}}] There is a statistically-relevant difference between
EX and EXIM configurations in task-related outcomes for a difficult
task but not for an easy one.
\end{lyxlist}
This is because the difficult task would encourage more interaction
with the artificial agent. Consequently, the implicit communications
added in EXIM configuration would be helpful enough to improve the
task performance and the perceived efficiency of the interaction.

\subsection{Experimental design}

To test hypothesis H7, we propose an interaction inspired by the first
phase of the previous experiment using a humanoid robot instead of
a virtual agent. We compare EX and EXIM configurations in a task with
two difficulty levels (easy and difficult), resulting in four experimental
conditions.

The task consists of replicating a sequence of colours shown on the
screen using coloured blocks.\footnote{Each colour is associated with a symbol to account for colour-blind
participants.} The sequence is shown for a few seconds on the screen, and the participant
should try to memorise it and then place the blocks one at a time
in the correct order, while the screen shows the task progress. The
participant can ask to see the sequence again at any moment during
the task, but points are lost for doing it. A score system is introduced
to encourage participants to try to complete the task using their
memory or the robot's help instead of asking to see the sequence repeatedly.
In the easy conditions, the sequences contain four colours and are
shown on the screen for four seconds, whereas in the difficult conditions,
they are eight colours long and shown on the screen for two seconds.
Considering the human mental storage capacity and the time needed
for consolidation in memory \citep{Luck1997,Cowan2001,Vogel2006},
a stimuli of four seconds should be more than enough to memorise a
set of four colours. On the other hand, a set of eight items exceeds
the short term memory storage limit and allowing less processing time
should make it more difficult to memorise it.

We implemented a human-robot communication infrastructure using the
NAO H25 V5 robot.\footnote{\url{http://doc.aldebaran.com/2-1/home_nao.html}}
The robot gives instructions at the beginning and at the end of each
task and responds to commands to show the sequence. In the EX configuration,
where the robot employs only an explicit communication, the robot
uses voice and there are sounds indicating whether the added block
is from the correct or wrong group. Also, the screen application shows
the robot's speech synchronised with its voice, and there are signs
accompanying the feedback sounds. The human explicitly communicates
by clicking on buttons on the screen application. In EXIM configurations,
along with the explicit communication, the robot also uses gaze and
facial expressions, and time is used as an implicit human communication.
When implicit communication is available, the robot looks at the next
correct colour group, and reacts when a block is added to the designated
space using a happy expression if it is correct and a sad one otherwise.
We implemented the robot gaze using a geometric approach similar to
the one proposed in \citep{Campos2020} to interpret human communications,
and created the happy and sad expressions using colour patterns in
the robot's eyes as recommended in \citep{Johnson2013}. Finally,
longer pauses (eight seconds) between added blocks are considered
as an indication that the human is uncertain about the next colour,
so the robot looks at the person, as to call their attention, and
look at the correct group again. In both communication configurations,
the robot looks at the human's face when talking to them.

\begin{figure}
\begin{centering}
\includegraphics[width=1\columnwidth]{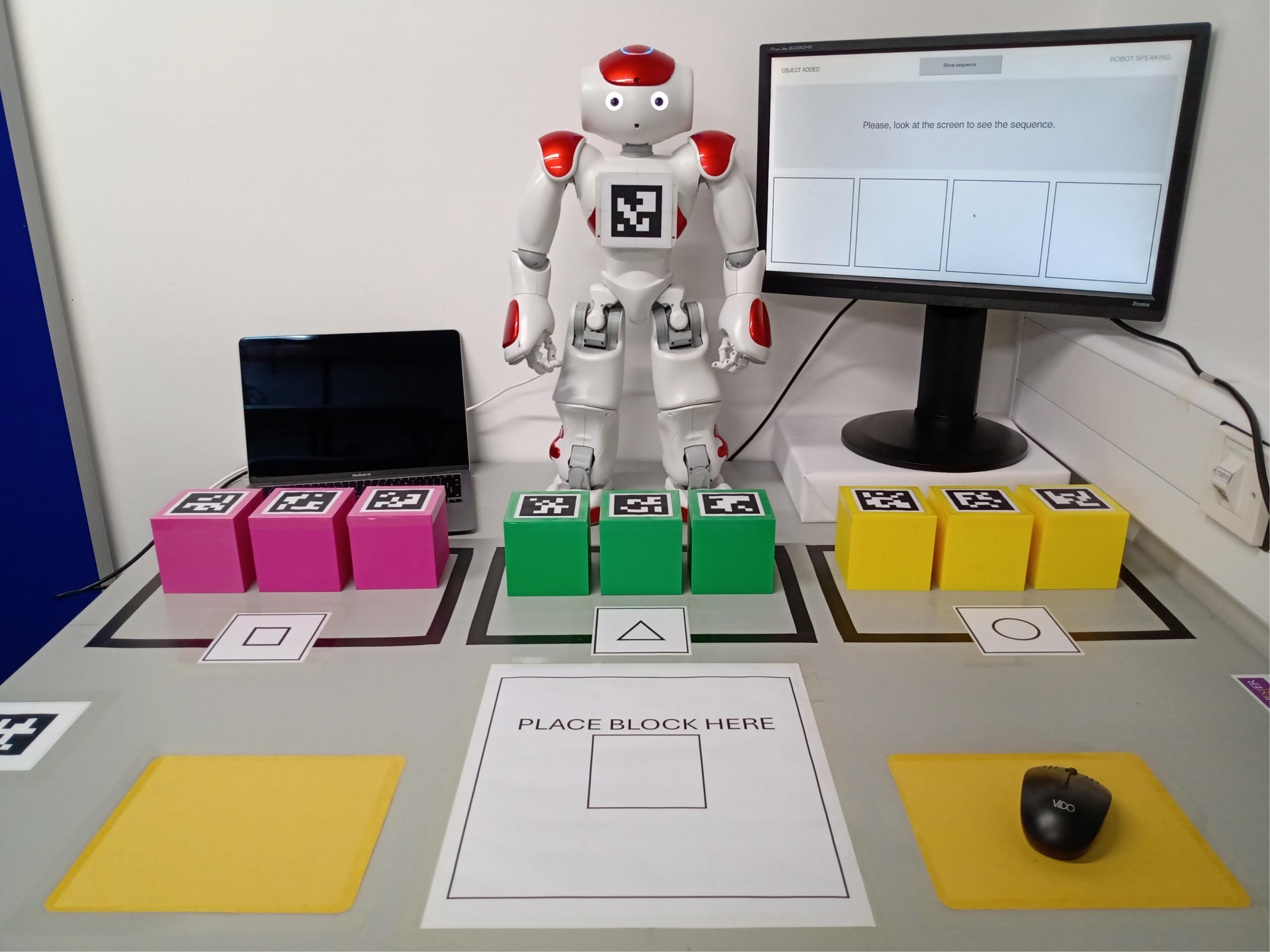}
\par\end{centering}
\begin{centering}
\includegraphics[width=1\columnwidth]{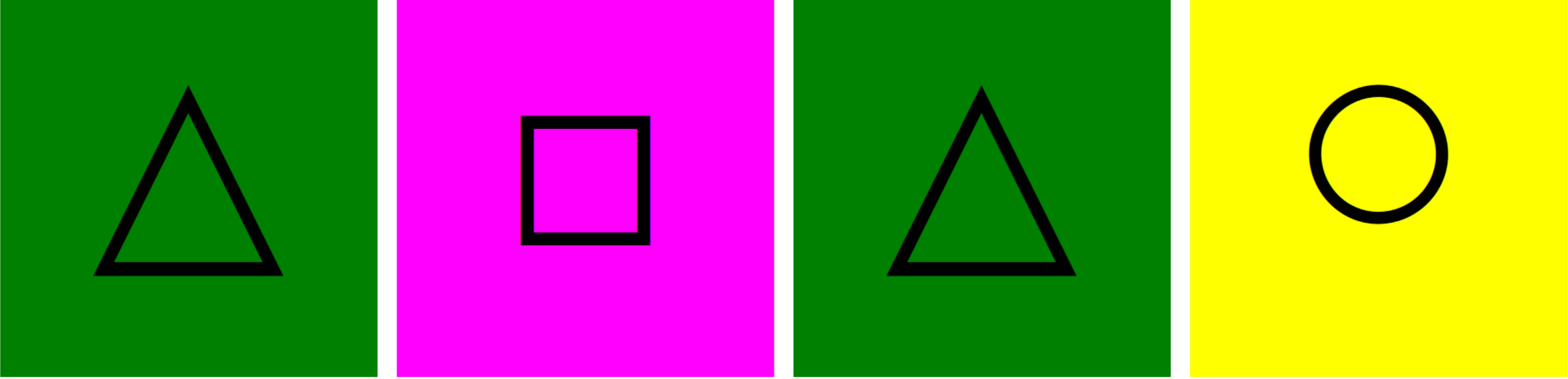}
\par\end{centering}
\caption{\label{fig:setup-experiment2}Setup for the follow-up experiment and
example of a colour sequence.}
\end{figure}

Fig.~\ref{fig:setup-experiment2} shows a picture of the experimental
setup and an example of a four-colour sequence for an easy condition
that would be shown on the computer screen. There are three groups
of blocks, separated by colour. The sequence in each task is randomly
generated at the beginning. Every participant goes through all conditions
in a random order, and they are asked to fill a questionnaire at the
end of each round. We consider the same measurements as the previous
study: time, number of errors, robot acceptance, sociability, transparency,
and the perceived efficiency of the interaction. Objective variables
were obtained automatically by the system. For the subjective variables,
we used a shorter version of the scales presented in Table~\ref{tab:likert_scales},
with only two items each, and replaced ``VA'' with ``NAO''. For the
acceptance scale, we used items 2 and 4; for sociability, items 1
and 5; items 1 and 4 for transparency; and items 1 and 2 from the
perceived efficiency scale.

We also added a practice round before the task conditions, so participants
could familiarise themselves with the task, the sounds, and the screen
application. The robot did not participate in the practice round,
and the instructions were given only through text on the screen. At
the end of the four conditions, the participant was asked to fill
a final questionnaire with demographics and a question to check whether
the instructions were clear enough, as this became a concern after
analysing the results from the previous experiment, discussed in Section~\ref{subsec:General-discussion}.

\subsection{Results}

Twenty eight people participated in the study, but three were excluded
due to a technical error during the experiment that prevented them
from finishing the four conditions. The final sample comprised 25
participants, with an average age of $27.04$ years, with $52\%$
self-identified as male, $44\%$ as female, and $4\%$ as non-binary/third
gender. The most common education level was an incomplete Doctorate
degree or equivalent ($40\%$) and all participants had at least an
incomplete Bachelor's degree or equivalent. The majority of participants
indicated they have familiarity with robots ($76\%$).

\begin{figure*}
\begin{centering}
\begin{minipage}[t]{0.32\textwidth}%
\begin{center}
\subfloat[\label{fig:comparison-time}Time]{\begin{centering}
\includegraphics[width=0.95\columnwidth]{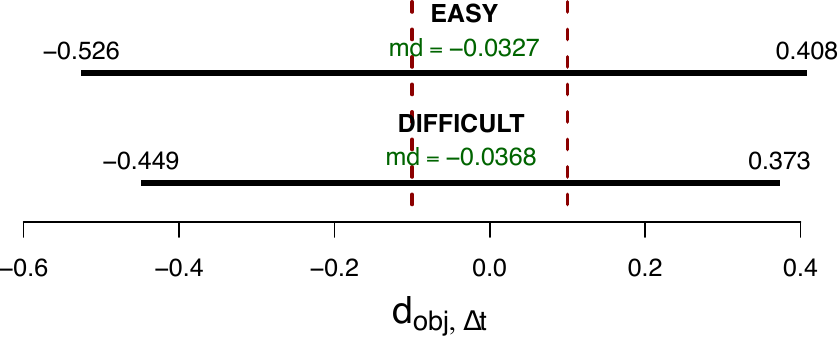}
\par\end{centering}
}
\par\end{center}%
\end{minipage}~~%
\begin{minipage}[t]{0.32\textwidth}%
\begin{center}
\subfloat[\label{fig:comparison-errors}Number of errors, 96.7\% (difficult)
of the distribution above ROPE]{\begin{centering}
\includegraphics[width=0.95\columnwidth]{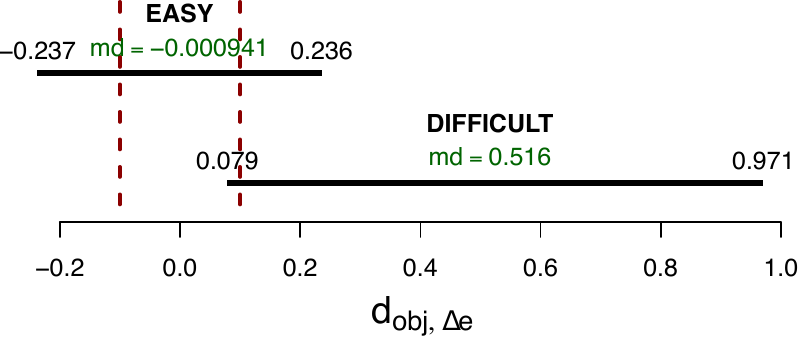}
\par\end{centering}
}
\par\end{center}%
\end{minipage}~~%
\begin{minipage}[t]{0.32\textwidth}%
\begin{center}
\subfloat[\label{fig:comparison-acceptance}Robot acceptance, 94.1\% (easy)
and 84.6\% (difficult) of the distribution above ROPE]{\begin{centering}
\includegraphics[width=0.95\columnwidth]{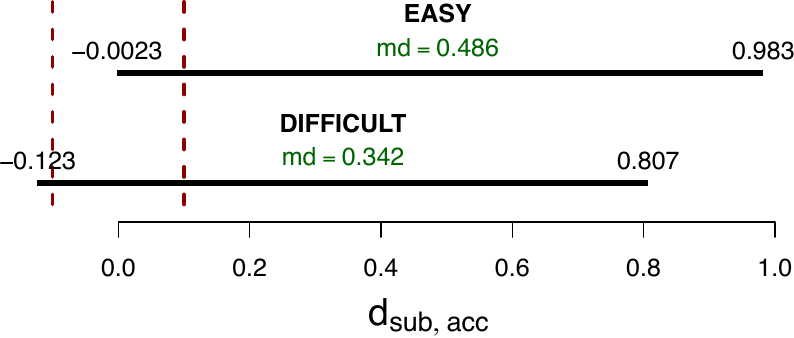}
\par\end{centering}
}
\par\end{center}%
\end{minipage}
\par\end{centering}
\begin{centering}
\begin{minipage}[t]{0.32\textwidth}%
\begin{center}
\subfloat[\label{fig:comparison-sociability}Robot sociability, 97\% (easy)
and 94.6\% (difficult) of the distribution above ROPE]{\begin{centering}
\includegraphics[width=0.95\columnwidth]{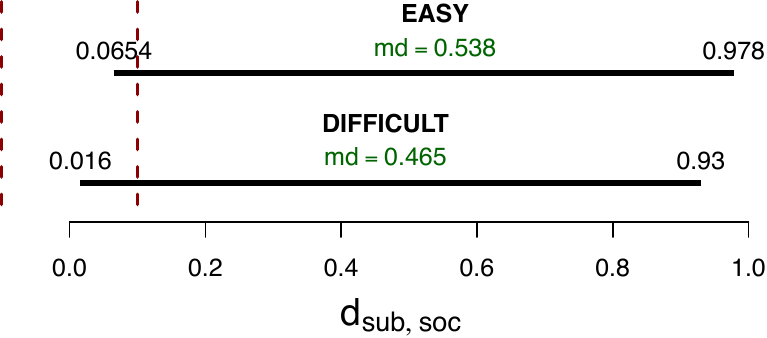}
\par\end{centering}
}
\par\end{center}%
\end{minipage}~~%
\begin{minipage}[t]{0.32\textwidth}%
\begin{center}
\subfloat[\label{fig:comparison-transparency}Robot transparency, 97.4\% (difficult)
of the distribution above ROPE]{\begin{centering}
\includegraphics[width=0.95\columnwidth]{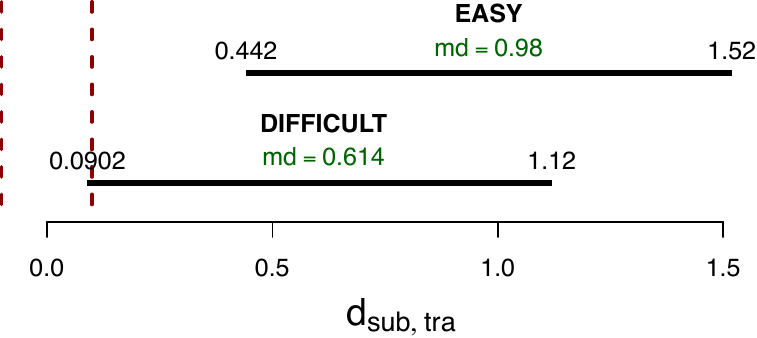}
\par\end{centering}
}
\par\end{center}%
\end{minipage}~~%
\begin{minipage}[t]{0.32\textwidth}%
\begin{center}
\subfloat[\label{fig:comparison-efficiency}Perceived efficiency of the interaction]{\begin{centering}
\includegraphics[width=0.95\columnwidth]{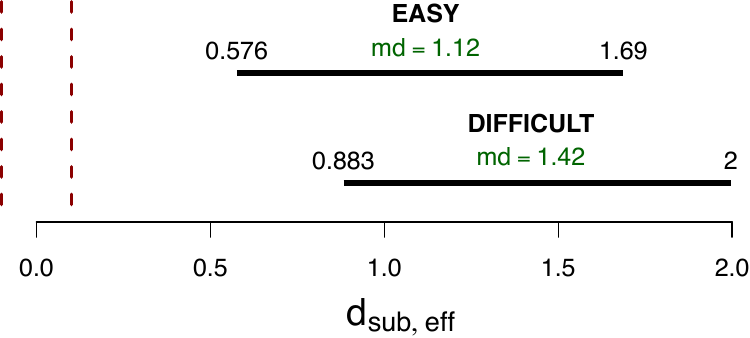}
\par\end{centering}
}
\par\end{center}%
\end{minipage}
\par\end{centering}
\caption{\label{fig:results-experiment2}Results of the Bayesian inference
of the follow-up experiment to assess hypothesis~H7. Each figure
shows a summary of the effect size distributions comparing EX and
EXIM configurations for the easy and difficult tasks. Effects sizes
were calculated using Eqs.~\ref{eq:d_obj} and \ref{eq:d_sub}. Median
and the limits of the 95\% HDI are annotated, and dashed vertical
lines indicate the ROPE.}
\end{figure*}

We use the same Bayesian approach as in the previous experiment, detailed
in Section~\ref{subsec:experiment-analysis}, and calculate effect
sizes as described in Section~\ref{subsec:Bayesian-analysis-results}.
We compare the EX and EXIM configurations for the easy and difficult
tasks, where positive effect sizes indicate results favouring the
EXIM configuration over the EX one. If hypothesis H7 holds, the difference
between EX and EXIM in task-related outcomes (time, number of errors,
and perceived efficiency of the interaction) will be more evident
for the difficult task than for the easy one, which will be expressed
by higher effect sizes for the difficult task. Fig.~\ref{fig:results-experiment2}
shows the results of the experiment.

For the difference in time, the task difficulty did not seem to have
impacted the comparison between the two communication configurations,
as the HDI for both effect sizes distributions include the null value
and range from effects in both directions. The difference in the number
of errors, however, was significantly higher for the difficult task
than for the easy one, as shown in Fig.~\ref{fig:comparison-errors}.
While $60.8\%$ of the effect size distribution for the easy task
is inside the ROPE, $96.7\%$ of the distribution for the difficult
one is above the upper limit of the ROPE, which means that the impact
of the communication configuration was stronger and more evident for
a more difficult task.

Compared with the results from the previous experiments, the HDIs
of the effect size distributions for the subjective measures for this
second experiment are wider. This is expected as we used fewer items
in the scales for the questionnaires. Thus, there is less data to
use to estimate the parameters, even though the samples are of similar
sizes. The results suggest that the impressions on the robot, \textit{i.e.},
its acceptance, sociability, and transparency, were better in the
EXIM configuration for both levels of task difficulty, which corroborates
the results from the experiments described in Section~\ref{sec:Experimental-design}.
For the acceptance results, shown in Fig.~\ref{fig:comparison-acceptance},
the HDI for the effect size in the difficult task included the entire
ROPE but $84.6\%$ of the distribution is above its upper limit, still
suggesting better acceptance when combining communication types. For
the robot transparency, the EXIM configuration seemed to be better
especially for the easy task. Different from the previous experiment,
the perceived efficiency of the interaction was more affected by the
communication configuration in this follow-up study, for both the
easy and difficult tasks, with an effect size slightly higher for
the difficult one.

\subsection{Discussion}

\subsubsection{System errors during experiment execution}

During the task execution, an error is counted when a block from the
wrong group is detected at the designated space. However, sometimes
the same block would be detected twice, without being added again
by the participant. If it was a correct block and the sequence had
the same group colour twice in a row, this would accidentally help
the execution of the task. Otherwise, it would be counted as an extra
error, despite not being an error from the participant. We checked
the recordings of the experiments looking for these events, and removed
the errors caused by this. However, we did not change the dataset
when system errors helped the participant complete the task to avoid
false positives, because we cannot say whether an actual human error
associated with that particular event would have happened immediately
after the system error. We also had some problems with occlusions:
depending on the participant's position, they would occlude the added
block from the camera's field of view. With the pause in the interaction
flow while the system waited for the detection, the participant would
eventually move, and the block would be detected, continuing with
the task. This would increase the time to complete the task and sometimes
would make the participant remove the block and try one from a different
group. In these cases, whenever the first undetected block was from
the correct group and the one detected after was from a wrong one,
we removed the error from the dataset, as the participant only tried
a different group because the system would not recognise the correct
one. The occlusions might have affected the time comparison and also
the robot acceptance, as some participants mentioned that the robot
sometimes took some time to recognise the blocks.

\subsubsection{The role of task difficulty}

Our hypothesis for this second experiment was that the impact of different
communication configurations on outcomes more related to the task
would be more evident in a difficult task than in an easier one. This
was clearly observed for the number of errors during task execution
and only slightly for the perceived efficiency of the interaction.
In contrast, we did not observe an evident difference in execution
time between the two task difficulty levels.

We observed a higher perceived efficiency of the interaction when
using the EXIM configuration for both easy and difficult tasks, with
$100\%$ of the distributions being above the upper limit of the ROPE.
Although the difference was slightly more evident for the difficult
task, no strong conclusion can be drawn about the relationship between
perceived interaction efficiency and task difficulty. The results
from this experiment and the previous one suggest that there might
be another factor impacting the evaluation of perceived efficiency,
and not only the task difficulty, such as the type of task or the
embodiment of the artificial agent, as we used a virtual agent in
the first experiment and a humanoid robot in the second one.

In the final questionnaire, participants were asked whether they thought
the task instructions were clear enough, and all answered positively.
With this, we conclude that the task instructions did not impact our
evaluations in this second experiment.

\subsubsection{{The embodiment effect}}

{One clear difference between the first experiment
and the follow-up experiment is the agent's embodiment: the first
experiment used a simple virtual agent (see Fig.~\ref{fig:virtual_agents}),
and the second used a small humanoid (see Fig.~\ref{fig:setup-experiment2}).
Since we had different results in the two experiments, we decided
to conduct a further investigation with our complete dataset to gain
insights into the effect of the agent's embodiment on both task- and
agent-related outcomes. To isolate the embodiment effect as best as
possible across the two experiments, we use only a subset of our aggregated
dataset that corresponds to very similar activities across both experiments,
more specifically the one related to the tasks of replicating a sequence
of four colours. From the first experiment, we use only the time and
number of errors of the first phase and the subset of items in questionnaires
that we also used in the follow-up experiment to ensure that effects
are measured using similar mechanisms (}{\emph{i.e.}}{,
same questionnaire items) across both experiments. Furthermore, only
the easy conditions of the second experiment is used since the difficult
conditions had longer colour sequences. Nonetheless, it is important
to emphasise that, in the first experiment dataset, it is not possible
to separate human perceptions between phase 1 and phase 2, as the
questionnaire was applied after a different communication condition
that included both phases. Therefore, the questionnaire measured the
combined effect of the two phases in the first experiment, which might
be a confounding factor.}

\begin{figure*}
\begin{centering}
\begin{minipage}[t]{0.32\textwidth}%
\begin{center}
\subfloat[\label{fig:embodiment- EX-time}Time]{\centering{}{\includegraphics[totalheight=0.55\textwidth]{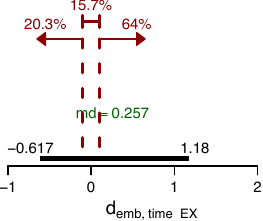}}}
\par\end{center}%
\end{minipage}{~~}%
\begin{minipage}[t]{0.32\textwidth}%
\begin{center}
\subfloat[\label{fig:embodiment- EX-errors}Number of errors]{\begin{centering}
{\includegraphics[totalheight=0.55\textwidth]{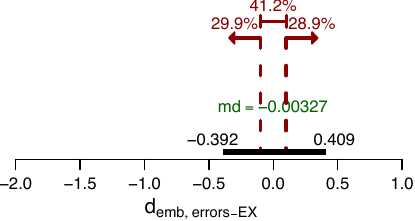}}
\par\end{centering}
{}}
\par\end{center}%
\end{minipage}{~~}%
\begin{minipage}[t]{0.32\textwidth}%
\begin{center}
\subfloat[\label{fig:eembodiment- EX-acceptance}Robot acceptance]{\begin{centering}
{\includegraphics[totalheight=0.55\textwidth]{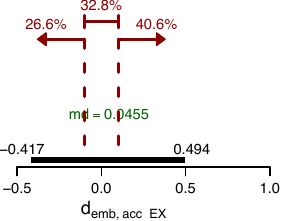}}
\par\end{centering}
{}}
\par\end{center}%
\end{minipage}
\par\end{centering}
\begin{centering}
\begin{minipage}[t]{0.32\textwidth}%
\begin{center}
\subfloat[\label{fig:embodiment- EX-sociability}Robot sociability]{\begin{centering}
{\includegraphics[totalheight=0.55\textwidth]{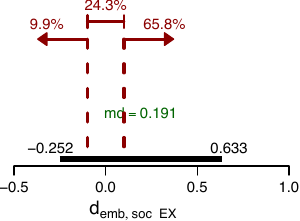}}
\par\end{centering}
{}}
\par\end{center}%
\end{minipage}{~~}%
\begin{minipage}[t]{0.32\textwidth}%
\begin{center}
\subfloat[\label{fig:embodiment- EX-transparency}Robot transparency]{\begin{centering}
{\includegraphics[totalheight=0.55\textwidth]{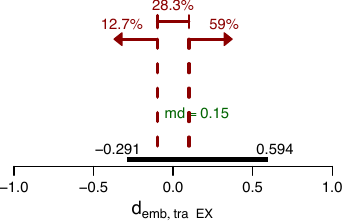}}
\par\end{centering}
{}}
\par\end{center}%
\end{minipage}{~~}%
\begin{minipage}[t]{0.32\textwidth}%
\begin{center}
\subfloat[\label{fig:embodiment- EX-efficiency}Perceived efficiency of the
interaction]{\begin{centering}
{\includegraphics[totalheight=0.55\textwidth]{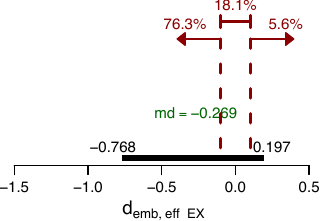}}
\par\end{centering}
{}}
\par\end{center}%
\end{minipage}
\par\end{centering}
{\caption{\label{fig:results-embodiment-EX}Results of the Bayesian inference
of the effect of embodiment in the variables measured in the EX communication
configuration. Each figure shows a summary of the effect size distributions
comparing the virtual agent and the humanoid embodiments. Effects
sizes were calculated using Eq.~\ref{eq:d-emb}. Median and the limits
of the 95\% HDI are annotated, and dashed vertical lines indicate
the ROPE.}
}
\end{figure*}

\begin{figure*}
\begin{centering}
\begin{minipage}[t]{0.32\textwidth}%
\begin{center}
\subfloat[\label{fig:embodiment- EXIM-time}Time]{\centering{}{\includegraphics[totalheight=0.55\textwidth]{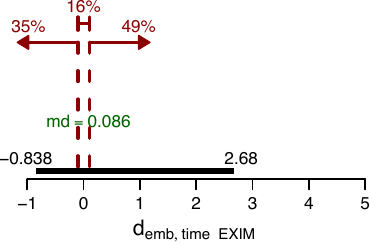}}}
\par\end{center}%
\end{minipage}{~~}%
\begin{minipage}[t]{0.32\textwidth}%
\begin{center}
\subfloat[\label{fig:embodiment- EXIM-errors}Number of errors]{\begin{centering}
{\includegraphics[totalheight=0.55\textwidth]{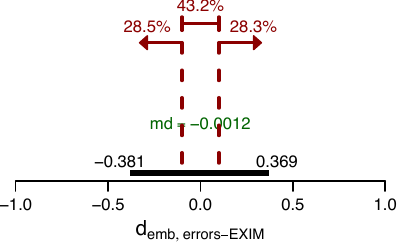}}
\par\end{centering}
{}}
\par\end{center}%
\end{minipage}{~~}%
\begin{minipage}[t]{0.32\textwidth}%
\begin{center}
\subfloat[\label{fig:embodiment- EXIM-acceptance}Robot acceptance]{\begin{centering}
{\includegraphics[totalheight=0.55\textwidth]{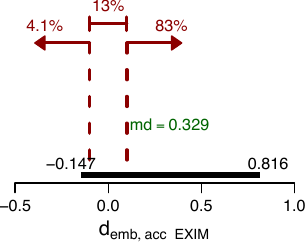}}
\par\end{centering}
{}}
\par\end{center}%
\end{minipage}
\par\end{centering}
\begin{centering}
\begin{minipage}[t]{0.32\textwidth}%
\begin{center}
\subfloat[\label{fig:embodiment- EXIM-sociability}Robot sociability]{\begin{centering}
{\includegraphics[totalheight=0.55\textwidth]{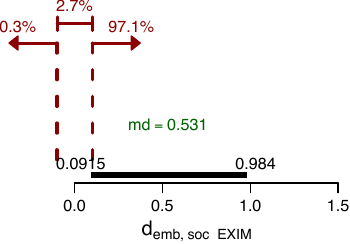}}
\par\end{centering}
{}}
\par\end{center}%
\end{minipage}{~~}%
\begin{minipage}[t]{0.32\textwidth}%
\begin{center}
\subfloat[\label{fig:embodiment- EXIM-transparency}Robot transparency]{\begin{centering}
{\includegraphics[totalheight=0.55\textwidth]{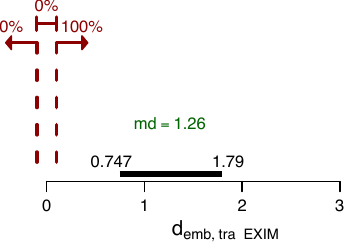}}
\par\end{centering}
{}}
\par\end{center}%
\end{minipage}{~~}%
\begin{minipage}[t]{0.32\textwidth}%
\begin{center}
\subfloat[\label{fig:embodiment- EXIM-efficiency}Perceived efficiency of the
interaction]{\begin{centering}
{\includegraphics[totalheight=0.55\textwidth]{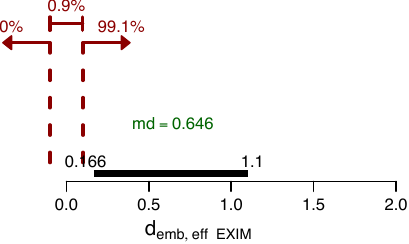}}
\par\end{centering}
{}}
\par\end{center}%
\end{minipage}
\par\end{centering}
{\caption{\label{fig:results-embodiment-EXIM}Results of the Bayesian inference
of the effect of embodiment in the variables measured in the EXIM
communication configuration. Each figure shows a summary of the effect
size distributions comparing the virtual agent and the humanoid embodiments.
Effects sizes were calculated using Eq.~\ref{eq:d-emb}. Median and
the limits of the 95\% HDI are annotated, and dashed vertical lines
indicate the ROPE.}
}
\end{figure*}

{In this analysis, we compare the use of the virtual
agent (VA) and the humanoid (H) separately for both communication
configurations (EX and EXIM), meaning we do not use the difference
in time and number of errors as in the previous analyses. Apart from
that, we follow the same procedure we used before; namely, we estimate
the parameters of the }{\emph{t}}{-distributions
associated with each variable, calculate an effect size}
\begin{equation}
d_{\mathrm{emb}}=\frac{\left(\mu_{\mathrm{H}}-\mu_{\mathrm{VA}}\right)}{\sqrt{0.5\left(\tau^{2}_{\mathrm{H}}+\tau^{2}_{\mathrm{VA}}\right)}}\label{eq:d-emb}
\end{equation}
{for all variables, and check the location of the
HDI with respect to a ROPE of length 0.2 around zero. Fig.~\ref{fig:results-embodiment-EX}
shows the distributions of the effect sizes for each variable in the
EX configuration, and Fig.~\ref{fig:results-embodiment-EXIM} shows
the results in the EXIM configuration. Positive effect sizes mean
higher values for the H condition than for the AV condition.}

{The time and number of errors associated with the
first phase of the first experiment are the variables we can isolate
more effectively for comparison with the second experiment, since
the tasks of replicating a sequence of colours by either pointing
to or placing coloured blocks are very similar. Therefore, any observed
effects should be mainly attributed to the embodiment. For task execution
time, in both EX and EXIM configurations, the HDI in the effect size
includes the ROPE and gives more credibility to positive effects,
meaning that if there is an effect of embodiment, it is more likely
to favour the virtual agent. For the number of errors, the HDI again
includes the ROPE for both EX and EXIM, and only smaller-than-medium
effects were observed (following Cohen's convention of 0.5 as a medium
effect size \citep{Cohen1988}).}

{The subjective measures have confounders that make
it difficult to isolate the embodiment effect. Any differences in
the subjective measures could be attributed to the agent's embodiment
or to the effect of phase 2 on people's perceptions in experiment
1, since we could not isolate those measures to phase 1 only. Nonetheless,
we would not expect a strong impact of phase 2 on the acceptance,
sociability, and transparency of the agent. On the other hand, since
participants considered the task in phase 2 more difficult to complete,
it is reasonable to expect that the perceived efficiency of the interaction
(a task-related outcome) would be smaller for the VA condition.}

{For the agent's acceptance in the EX configuration,
the HDI includes the ROPE, with up to medium effects. However, in
the EXIM configuration, although the HDI still contains the ROPE,
positive effects are more likely than negative ones, indicating higher
acceptance of the humanoid than of the virtual agent.}

{Sociability for H seems higher than for VA, especially
in the EXIM configuration. Sociability is a measure of how social
the agent is perceived to be, and one of the questions in the questionnaires
concerns whether the agent's behaviour is similar to a person's. Facial
expressions are used to express emotions in human interactions, so
we would expect that the agent would have been perceived as more social
(}{\emph{i.e.}}{, more similar to
a human) with the effective use of facial expressions, which happened
only for the VA, as facial expressions of H were not correctly interpreted.
Still, the sociability of the humanoid embodiment was higher, which
could mean it might have an important positive impact on how social
the agent is perceived, even when implicit communication through facial
expressions is unclear.}

{Transparency relates to how clear the agent is, and
our comparison shows large effects in favour of H for the EXIM configuration.
In both experiments, facial expressions were used as feedback, while
gaze helped with task execution. As mentioned before, facial expressions
were clearer for the VA. In contrast, we believe that gaze was clearer
for the H condition due to the location of the regions of interest
where the agent should look, and because the communication included
an actual physical movement of the robot's head instead of a virtual
one created using a perspective effect on a flat screen, as in experiment
1. This suggests that clear, useful communications could have more
impact on how transparent the agent is perceived than clear implicit
communication in general.}

{For the perceived efficiency of the interaction,
our comparison shows opposite effects in EX and EXIM configurations.
For EX, the interaction with the VA seems to be perceived as more
efficient, as 76.3\% of the distribution was below the lower limit
of the ROPE. This happened even with the inclusion of phase 2 data
from experiment 1, in which the VA did not help in a task that was
considered more difficult. Even with this confounding factor that
could bias the comparison against the VA, the interaction with it
was still perceived as more efficient for this configuration. On the
other hand, in the EXIM configuration, we observed a strong effect
favouring H for the perceived efficiency, as 99.1\% of the distribution
was above the upper limit of the ROPE.}

{Based on the previous analyses, we conclude that
the embodiment might have affected the impact of communication in
the perceived interaction efficiency. Indeed, although there was no
clear difference between EX and EXIM in the first experiment,}\footnote{{Fig.~\ref{fig:results-efficiency} shows that the
HDI completely contains the ROPE and only 62.9\% of the distribution
is above the ROPE's upper limit, which slightly favours the EXIM condition
but does not provide a robust conclusion.}}{{} the follow-up experiment showed substantial differences
for both task difficulty levels, as the HDIs excluded the ROPEs with
large effect sizes. The result from this last analysis shows that
communication also affects how embodiment influences perceived efficiency:
when using only EX, the VA was perceived as more efficient, but when
using EXIM, the perceived efficiency was higher for H. In summary,
the combination of embodiment and communication configuration seems
to have an important influence on the perceived efficiency of the
interaction. The estimated mean of perceived efficiency for the combination
communication/embodiment is smaller for EX+H (median $2.54$, $95\%$-HDI
$[1.77,3.38]$) and greater for EXIM+H (median $4.62$, $95\%$-HDI
$[3.88,5.54]$).}

\section{{Power analysis}}

{An important limitation in both our experiments is
the relatively small sample sizes, although they are comparable to
those in the literature. Small sample sizes impact the sensitivity
of the experiment to detect effects and to obtain precise estimates.
We conducted a retrospective power analysis following the procedure
presented in \citep{Kruschke2015}. In this analysis, we generate
a number of simulated samples using the posterior distributions of
the parameters we estimated, run the same analysis again with the
simulated samples, and check whether the goals were achieved or not.
Power is then defined as the proportion of times the goal was achieved
for the simulated samples \citep{Kruschke2015}. In our study, hypotheses
are validated if the $95\%$-HDI of the effect size's posterior distribution
is entirely above the ROPE around the null value (goal: $\mathrm{HDI}\succ\mathrm{ROPE}$).
We calculated this using 1000 simulated samples and the power for
each variable and each experiment are shown in Table~\ref{tab:power}.
We also show the probability that the $95\%$-HDI in the effect size
distribution excludes the ROPE entirely in any direction (goal: $\mathrm{HDI}\cap\mathrm{ROPE}=\varnothing$).}

\begin{table*}
\centering{}{\caption{\label{tab:power}Power of our experiments for the goals of the HDI
on the posterior effect size distribution excluding the ROPE in any
direction (\emph{$\mathrm{HDI}\cap\mathrm{ROPE}=\varnothing$}) and
being entirely above it (\emph{$\mathrm{HDI}\succ\mathrm{ROPE}$}).}
}%
\begin{tabular*}{1\textwidth}{@{\extracolsep{\fill}}lcccc}
\hline 
 & \multirow{2}{*}{\textbf{{Goal}}} & \multirow{2}{*}{\textbf{{Experiment 1}}} & \multicolumn{2}{c}{\textbf{{Experiment 2}}}\tabularnewline
 &  &  & \textbf{{Easy}} & \textbf{{Difficult}}\tabularnewline
\hline 
\hline 
\noalign{\vskip\doublerulesep}
\multirow{2}{*}{\textbf{{Time}}} & {$\mathrm{HDI}\cap\mathrm{ROPE}=\varnothing$} & {7.3\%} & {8.6\%} & {8.4\%}\tabularnewline[\doublerulesep]
\noalign{\vskip\doublerulesep}
 & {$\mathrm{HDI}\succ\mathrm{ROPE}$} & {2.6\%} & {4.2\%} & {3\%}\tabularnewline
\hline 
\multirow{2}{*}{\textbf{{Number of errors}}} & {$\mathrm{HDI}\cap\mathrm{ROPE}=\varnothing$} & {4.7\%} & {0.7\%} & {41\%}\tabularnewline
 & {$\mathrm{HDI}\succ\mathrm{ROPE}$} & {2.6\%} & {0.5\%} & {41\%}\tabularnewline
\hline 
\multirow{2}{*}{\textbf{{Acceptance}}} & {$\mathrm{HDI}\cap\mathrm{ROPE}=\varnothing$} & {26.3\%} & {7.2\%} & {6.2\%}\tabularnewline
 & {$\mathrm{HDI}\succ\mathrm{ROPE}$} & {25.5\%} & {7.2\%} & {6.1\%}\tabularnewline
\hline 
\multirow{2}{*}{\textbf{{Sociability}}} & {$\mathrm{HDI}\cap\mathrm{ROPE}=\varnothing$} & {38.3\%} & {29.6\%} & {25.3\%}\tabularnewline
 & {$\mathrm{HDI}\succ\mathrm{ROPE}$} & {36.7\%} & {29.5\%} & {25.2\%}\tabularnewline
\hline 
\multirow{2}{*}{\textbf{{Transparency}}} & {$\mathrm{HDI}\cap\mathrm{ROPE}=\varnothing$} & {32.4\%} & {43.9\%} & {16.8\%}\tabularnewline
 & {$\mathrm{HDI}\succ\mathrm{ROPE}$} & {31.7\%} & {43.9\%} & {16.8\%}\tabularnewline
\hline 
\multirow{2}{*}{\textbf{{Perceived efficiency of the interaction}}} & {$\mathrm{HDI}\cap\mathrm{ROPE}=\varnothing$} & {12.8\%} & {75.7\%} & {90.3\%}\tabularnewline
 & {$\mathrm{HDI}\succ\mathrm{ROPE}$} & {11\%} & {75.7\%} & {90.3\%}\tabularnewline
\hline 
\end{tabular*}
\end{table*}

{The values shown in Table~\ref{tab:power} mean,
for example, that a sample of 25 participants would result in the
$95\%$-HDI falling completely above the ROPE $41\%$ of times for
the number of errors in the difficult condition of Experiment 2. This
implies that, with the statistical power of a sample of 25 participants,
our experiments would unequivocally support our hypothesis (in the
sense that $95\%$-HDI is above the ROPE) regarding the impact of
using explicit and implicit communication over using only explicit
communication $41\%$ of the time. Notice, however, that there is
a 96.7\% chance that our data confirm our hypothesis, as 96.7\% of
the posterior distribution is above the ROPE (see Fig.~14(b)). The
increase in the power between the easy and difficult conditions for
number of errors and perceived efficiency of the interaction can be
explained by an effect more pronounced in the difficult one. Also,
our follow-up experiment was much more powerful to detect a difference
between the perceived efficiency in the two communication configurations
than the first experiment, achieving a power of $90.3\%$ for the
difficult condition, showing that even with a relatively small sample,
the results regarding this variable for this experiment are robust.
For the first experiment, although the power was never high, it was
higher for the agent-related outcomes (acceptance, sociability, and
transparency) than the task-related ones, suggesting at least a more
pronounced effect on the agent-related variables.}

\section{Conclusions}
In this work, we have investigated the effects of combining explicit
and implicit communications on performance and on people's perceptions
while interacting with artificial agents. Following the HRI literature,
we have first hypothesised that using a combination of explicit and
implicit communications between a human and a virtual agent would
reduce the time and the number of errors in task execution and increase
the acceptance of the virtual agent, its sociability and transparency,
and the perceived efficiency of the interaction. The experimental
results for the particular task in this study suggest that combining
explicit and implicit communications improve the subjective measures
related to the virtual agent (acceptance, sociability, and transparency),
but not task-related outcomes (time, number of errors, and perceived
efficiency of the interaction).

To further understand the difference in the effect of the mixed explicit-implicit
communication modalities in task-related and artificial-agent subjective-related
metrics, we hypothesised that this dichotomy could be caused by tasks
that are easily completed without the need of collaboration with the
artificial agent. We tested this in a follow-up experiment, manipulating
the communication configuration and also the difficulty level of a
second HRI task. This follow-up experiment shows that the time to
complete the task was still not affected by the combination of implicit
and explicit communication modalities. However, there was a clear
difference between the two communication configurations in the perceived
efficiency of the interaction, for both levels of task difficulty,
showing that the perceived efficiency of the interaction increased
when using a combination of explicit and implicit communication modalities.
Furthermore, the number of errors was significantly higher for the
EX configuration than the EXIM one in the difficult task, but not
for the easy one. This supports the additional hypothesis that a combination
of implicit and explicit communication modalities have more impact
in task-related metrics when tasks are difficult. The experimental
results from the follow-up experiments also suggest an improvement
in agent-related outcomes when using the EXIM configuration, which
was slightly more evident for the easy task. Although the estimations
were relatively less precise to draw a strong conclusion about this
trend, as the questionnaires contained fewer items in the follow-up
experiment, it supports the results from the first experiment that
the EXIM configuration increases the acceptance of artificial agent,
its sociability and transparency.

{As we used a virtual agent in the first experiment
and a humanoid in the follow-up experiment, we conducted further analysis
on a subset of our dataset to investigate the effect of embodiment.
We compared the same task- and agent-related outcomes by considering
only the task of replicating a four-colour sequence from both experiments.
However, we cannot fully separate the human perceptions about the
virtual agent between the two tasks in the first experiment, as participants
answered agent-related questions only after the second task. Our results
suggest that there is no strong indication that time and number of
errors were affected by embodiment. The agent's acceptance seemed
to be the same for both embodiments in the EX configuration, although
slightly better for the humanoid in the EXIM configuration. Sociability
and transparency results favoured the humanoid, with medium to strong
effects in the EXIM configuration. For perceived interaction efficiency,
our analyses show a combined effect of embodiment and communication
configuration, with the best case being the humanoid with explicit
and implicit modalities and the worst case the humanoid with explicit
modalities only.}

From the results of both experiments, we conclude that combining explicit
and implicit communication is likely to improve subjective perceptions
about virtual agents and robots. The impact of different communication
configurations in the number of errors might be more sensitive to
the task difficulty, whereas the time to complete it might be affected
more by other factors than the modalities the artificial agents use
to communicate. {The perceived interaction efficiency
seems to be affected by a combination of embodiment and communication
configuration.} With this knowledge, communication can be used purposefully
to help achieve interaction goals. For example, in social scenarios,
combining explicit and implicit communication modalities might help
a robot be perceived as more sociable. If an agent is designed to
help reduce the time needed to execute a task, our results suggest
that simply manipulating the communication might not be enough; other
routes should be considered.

The impact of combining communication modalities on the subjective
perceptions towards artificial agents is important, regardless of
task performance. For example, for socially interactive robots, which
are designed to socially interact with people \citep{Fong2003}, these
perceptions might be even more important than performance indicators.
Artificial agents, virtual or physical ones, need to be well accepted
in every scenario where humans are expected to interact with them.
For long-term applications, the robot's behaviour should encourage
the human to keep and repeat the interaction. Increasing the acceptance
of artificial agents and enabling more natural communication by making
them sociable and transparent is important to increase the human willingness
to collaborate and interact with them. The transparency of the agent
is also important both in more social environments, as a way of building
trust, for example, and in any collaborative scenarios, where a clear
understanding of the other agent's intentions is essential.

{Finally, our study helps shed light on the various
aspects at play in different communication configurations in interactions
between humans and artificial agents. The results from our analyses
show that task-related outcomes, in particular, are more nuanced than
the literature has suggested.}

The supplementary material with an extended version of the results
of both experiments can be found on arXiv.\footnote{\url{https://arxiv.org/abs/2409.18745}}
Source codes used in both experiments are available on GitHub.\cprotect\footnote{\url{https://github.com/Adorno-Lab/human_artificial_agent_interaction_experiments}}
The informed consent asked from participants in the first experiment
did not include data sharing. Therefore, these cannot be made publicly
available. Anonymised data from the follow-up experiment can be found
on Figshare.\footnote{\url{https://doi.org/10.48420/33213606}}

\subsection{Limitations of the work}

One limitation of our work is that, in both experiments, the artificial
agent only acted as a helping tool for task execution rather than
being a required collaborator. This is a valid scenario that is also
worth exploring, as it is relevant to social applications, such as
companion robots, as well as supervisory applications. However, the
results from this research should not be directly generalised to scenarios
where the robot has a more active role in task completion.

For the follow-up experiment, we proposed a task inspired by the first
phase of the first experiment. However, the tasks were slightly different,
as were the artificial agents and their communication modalities.
Although those experiments revealed some degree of generality in terms
of subjective metrics, as both interaction with the virtual agent
and humanoid when using EXIM modalities improved acceptance, sociability,
and transparency, completion time across both experiments was not
affected, from a statistical point of view, by the communication configurations.
This invariance with respect to communication configurations highlights
that other effects might be at play when it comes to the completion
time, warranting further investigation. The decrease in the number
of errors when combining modalities only for a difficult task and
the increase in perception of efficiency for both difficulty levels
are robust for the particular combination of humanoid embodiment and
task, which is an insightful result.

{The results of the analysis of the embodiment effect
suggest an interesting combined effect of embodiment and communication
on the perceived efficiency of the interaction. However, there are
still confounding factors that we could not isolate in our dataset
to conduct a more rigorous comparison.}

{Our estimates for the effect size are generally imprecise,
as they have large HDIs. Except for the effect size of the number
of errors in the easy condition of Experiment 2 ($0.3\%$), the probabilities
of obtaining HDIs smaller than 0.2 with our samples in both experiments
were zero, which reflects the low statistical power of our experiments.
Nevertheless, stronger conclusions can still be drawn when the HDI
does not overlap with the ROPE, such as in the case of the perceived
interaction efficiency for both task difficulty levels in Experiment
2, despite the HDI being large (}{\emph{i.e.}}{,
imprecise). This is because, although the estimate is imprecise, the
effect size of those variables is very large, allowing us to measure
it with statistically significant confidence. Furthermore, the posterior
distributions we generated can still be incorporated in future research
by using them to inform prior distributions in new estimations, which
should help narrow the HDIs and increase the certainty of future results.}

{Finally, we estimated how large the samples would
have to be to obtain $80\%$ power to detect an HDI smaller than 0.2
and 0.1 for the effect size distributions, as well as to avoid empty
levels in the subjective measures responses and consequently avoid
adding extra data to our samples, as described in Appendix~\ref{sec:limitations}.
For all variables, the sample size required would have to be larger
than 100, which suggests that our protocol might not have been appropriate
to measure what we intended with a reasonable sample size. Increasing
measurement reliability (}{\emph{e.g.}}{,
improving the infrastructure to prevent errors caused by the system,
including occlusions that led to increased task time), controlling
for other aspects that might influence the variables, and amplifying
the effect using another interaction protocol are strategies to increase
power without necessarily having to increase the sample size substantially
\citep{Kruschke2015}.}

\subsection{Future work}

Further studies should investigate other factors that might affect
the time needed to complete a task while interacting with an artificial
agent, as this might be an important outcome in some scenarios and
did not seem to be affected by the communication type in either of
our experiments. Another possibility is to investigate the impact
of communication configurations in other types of tasks. We would
expect the increase in transparency to translate to performance improvement
in collaborative tasks where the artificial agent has a more crucial
role. The results of our second experiment also raised the question
of whether a more difficult task, where the human might be more focused
on completing it than on the interaction, could reduce the impact
of communication type on the subjective perceptions of the artificial
agents. {Also, other experiments could be conducted
to investigate the relationship between embodiment and communication
configuration in subjective measures, especially perceived interaction
efficiency.} Furthermore, the effect of each implicit modality should
be investigated to better understand whether adding only one modality
would be enough to achieve the results we observed when adding multiple
modalities, and whether one modality is more important than the others.

Additionally, a promising research avenue is to extend our cross-sectional
studies to longitudinal ones to assess whether long interaction times
affect subjective metrics in the class of HRI tasks studied in the
present study. We hypothesise that humans will find the artificial
agent more transparent and more likely to accept it when explicit
and implicit communication modalities are employed, because repeated
interactions will make them more familiar with the nuances of implicit
communication. We also expect that the perceived efficiency of the
interaction will increase substantially for particularly difficult
tasks in longitudinal studies. This is because humans tend to get
better at the task through practice, and the EXIM configuration objectively
helps them make fewer mistakes during execution when the task is sufficiently
difficult. Therefore, those two effects might be confounding variables
that result in a positive net effect on the perceived efficiency of
the interaction.

Finally, {instead of providing results in the form
of }{\emph{p}}{-values, as the vast
majority of the literature does, which makes incorporating results
into future studies harder, we contribute posterior distributions
obtained through a rigorous Bayesian analysis that can be easily incorporated
into future research.} When designing further HRI experiments to assess
differences between explicit and explicit-plus-implicit communication
configurations, the posterior distributions obtained from this study
can be used to inform prior distributions in future Bayesian analyses,
leading to more precise estimations, and to construct a data-generating
distribution for a prospective power analysis to inform the sample
size needed for more robust conclusions.

\appendix

\section{\label{sec:limitations}Ordinal model limitations and adjustments
due to empty levels}

On the ordinal model, described in Section~\ref{subsec:Subjective-measures},
there is nothing to specify that the thresholds are in ascending order,
that is, $\theta_{1}<\theta_{2}<\dots<\theta_{K-1}$. However, if
we have inverted thresholds ($\theta_{k-1}>\theta_{k}$), the probability
of the $k$th ordinal level becomes negative, violating the first
probability axiom. The MCMC posterior distribution sample generator
also does not prevent inverted thresholds from being generated. To
solve that problem, Kruschke includes a condition that if inverted
thresholds are generated at the $k$th level, the corresponding calculated
probability would be zero and then the thresholds are discarded \citep[Section 23.2.1]{Kruschke2015}.
This implementation solution works \textit{as long as there is at
least one answer in each level of the data sample}. When the data
contains an empty level, there is nothing in the mathematical model
or in the implementation that prevents the generation of inverted
thresholds, and hence negative probabilities are generated.

This model limitation is specially problematic when we have a small
data sample, which increases the chance of an empty level. In his
implementation, Kruschke decided to compress out the empty levels.
For example, if there are five levels of response (1 to 5) but level
2 does no occur in the data sample, Kruschke's implementation changes
the data set, considering only four levels (1 to 4). Then, this updated
data set is used in the Bayesian inference. We can easily see that
this change can alter the parameter estimation; if the original data
set, with empty level 2, came from a latent distribution with mean
$\mu=4$, the compression would shift the estimated mean to a value
smaller than $4$.\footnote{We discussed the matter with Prof. John Kruschke in private communication.
According to him, to keep the original number of levels even when
there are empty levels in the data sample, it would be necessary to
change the model (using more restricted priors for the thresholds,
for example) and the mechanism to select possible thresholds during
the MCMC sample generation.}

\begin{table*}
\begin{centering}
\caption{\label{tab:empty-levels}Advantages and drawbacks of each method considered
for the analysis of ordinal data containing empty levels.}
\par\end{centering}
\centering{}%
\begin{tabular}[b]{>{\raggedright}m{0.3\textwidth}>{\raggedright}m{0.3\textwidth}>{\raggedright}b{0.3\textwidth}}
 & {\small\textbf{Advantages}} & {\small\textbf{Drawbacks}}\tabularnewline[\doublerulesep]
\hline 
\noalign{\vskip0.015\columnwidth}
{\small\textbf{Keep empty levels}} & {\small Original data} & {\small Possible negative probabilities}\tabularnewline[0.015\columnwidth]
\hline 
\noalign{\vskip0.015\columnwidth}
\multirow{2}{0.3\textwidth}{{\small\textbf{Compress empty levels}}} & \multirow{2}{0.3\textwidth}{{\small No negative probabilities}} & {\small Changed data}\tabularnewline
 &  & {\small Estimation bias}\tabularnewline[0.015\columnwidth]
\hline 
\noalign{\vskip0.015\columnwidth}
\multirow{2}{0.3\textwidth}{{\small\textbf{Restricted prior distributions for thresholds $\theta_{k}$}}} & {\small Original data} & \multirow{2}{0.3\textwidth}{{\small Estimation bias}}\tabularnewline
 & {\small No negative probabilities} & \tabularnewline[0.015\columnwidth]
\hline 
\noalign{\vskip0.015\columnwidth}
\multirow{2}{0.3\textwidth}{{\small\textbf{Add extra data}}} & \multirow{2}{0.3\textwidth}{{\small No negative probabilities}} & {\small Changed data}\tabularnewline
 &  & {\small Inflated estimation of the latent scale parameter $\tau$}\tabularnewline[0.015\columnwidth]
\hline 
\end{tabular}
\end{table*}

Because Kruschke's data compression strategy could bias our estimation
and the ordinal levels are related to the actual response options
in the questionnaires, we chose not to follow it. Therefore, we consider
some alternatives to deal with the empty levels problem, summarised
in Table~\ref{tab:empty-levels}. One option is to use more restricted
priors for the thresholds, more specifically by reducing the standard
deviation. That solves the problem of negative probabilities but also
adds an estimation bias, since restricted priors mean a strong initial
belief about the parameter value. Another reason against the restricted
priors is that centring them at the same values for all items already
confronts the idea that different items access the same latent variable
through different thresholds. Therefore, considerable standard deviation
should be used to allow greater variability. Moreover, we do not have
enough information to place them at different locations for each item. 

Another alternative is to add extra data to eliminate empty levels,
and we consider three ways of doing it: 1) adding one extra answer
in all levels, empty or not, to avoid shifting the estimation of the
mean $\mu$; 2) adding one extra answer only in empty levels; 3) and
adding one extra answer in each empty level and adding extra answers
in non empty levels to keep the probability (frequency of occurrence)
of each level as close to the original as possible, adding a maximum
of $K$ new answers in the data sample of each item. In our tests,
when we added extra data, we observed an inflated estimation of the
scale parameter $\tau$, with greater credibility given to values
higher than the real one.

Since all aforementioned alternatives to the solution of empty levels
change the estimated distribution, a sensible choice must consider
two main aspects: the results should be mathematical coherent (no
negative probabilities) and the effects of the change in the data
sample should not favour our hypotheses (that is, we seek a conservative
solution). Using simulated data, we have found the best consistent
results by adding one extra datum to each empty level. With this method,
only the estimation of the scale parameter $\tau$ was significantly
hindered, making it more difficult to validate our hypotheses. The
parameter $\tau$ appears on the denominator of the effect size; therefore,
greater scales imply smaller effect sizes.

Another important aspect of the model is how the estimations strongly
depend on the value of the fixed parameters. If we fix the latent
mean and/or scale parameter, it would be more difficult to interpret
the estimations. Consequently, following Kruschke's suggestion \citep{Kruschke2015},
a better option is to fix the extreme thresholds of one of the items
considering the response levels in the questionnaires. We arbitrarily
chose to fix the extreme thresholds of the first item of each scale
as shown in Table~\ref{tab:likert_scales}.

Unlike the objective measures, which do not use an ordinal model and
for which we pair the observations for each participant and use the
difference between the communication configurations, we estimate the
parameters of each group separately for the subjective measures. This
is because considering five response levels (see Section~\ref{subsec:Measurements}),
the difference between communication configurations would assume values
from $-4$ to $4$, which do not have a direct relation with the original
five response options in the questionnaire. Remember that we fix the
extreme thresholds in $1.5$ and $K-0.5$ to enable us to interpret
the estimation of the latent parameters considering the actual response
options presented to participants, as discussed in Section~\ref{subsec:Subjective-measures}
(see Fig.~\ref{fig:fixed-thresholds}).  Moreover, more levels would
increase the chance of empty ones and require more extra data in those
empty levels, causing more change in the final sample and the estimations.

\bibliographystyle{sn-basic}
\bibliography{mylibrary}

@article{Agrigoroaie2020,
  title = {In the {{Wild HRI Scenario}}: {{Influence}} of {{Regulatory Focus Theory}}},
  author = {Agrigoroaie, Roxana and Ciocirlan, Stefan-Dan and Tapus, Adriana},
  year = 2020,
  month = apr,
  journal = {Frontiers in Robotics and AI},
  volume = {7},
  number = {April},
  pages = {1--11},
  issn = {2296-9144},
  doi = {10.3389/frobt.2020.00058}
}

@article{Ajoudani2018,
  title = {Progress and Prospects of the Human-Robot Collaboration},
  author = {Ajoudani, Arash and Zanchettin, Andrea Maria and Ivaldi, Serena and {Albu-Sch{\"a}ffer}, Alin and Kosuge, Kazuhiro and Khatib, Oussama},
  year = 2018,
  journal = {Autonomous Robots},
  volume = {42},
  number = {5},
  pages = {957--975},
  publisher = {Springer US},
  issn = {0929-5593},
  doi = {10.1007/s10514-017-9677-2}
}

@inproceedings{Andrist2017,
  title = {Looking {{Coordinated}}: {{Bidirectional Gaze Mechanisms}} for {{Collaborative Interaction}} with {{Virtual Characters}}},
  shorttitle = {Looking {{Coordinated}}},
  booktitle = {Proceedings of the 2017 {{CHI Conference}} on {{Human Factors}} in {{Computing Systems}}},
  author = {Andrist, Sean and Gleicher, Michael and Mutlu, Bilge},
  year = 2017,
  month = may,
  pages = {2571--2582},
  publisher = {ACM},
  address = {Denver Colorado USA},
  doi = {10.1145/3025453.3026033},
  urldate = {2025-05-27},
  isbn = {978-1-4503-4655-9},
  langid = {english}
}

@article{Apraiz2025,
  title = {The User Experience in Industrial Human-Robot Interaction: {{A}} Comparative Analysis of {{Unimodal}} and {{Multimodal}} Interfaces for Disassembly Tasks},
  shorttitle = {The User Experience in Industrial Human-Robot Interaction},
  author = {Apraiz, Ainhoa and Lasa, Ganix and Mazmela, Maitane and {Arana-Arexolaleiba}, Nestor and Elguea, {\'I}{\~n}igo and Escallada, Oscar and Osa, Nagore and Etxabe, Amaia},
  year = 2025,
  month = oct,
  journal = {Robotics and Computer-Integrated Manufacturing},
  volume = {95},
  pages = {103045},
  issn = {07365845},
  doi = {10.1016/j.rcim.2025.103045},
  urldate = {2026-06-23},
  langid = {english}
}

@inproceedings{Asfour2006,
  title = {{{ARMAR-III}}: {{An Integrated Humanoid Platform}} for {{Sensory-Motor Control}}},
  booktitle = {2006 6th {{IEEE-RAS International Conference}} on {{Humanoid Robots}}},
  author = {Asfour, T. and Regenstein, K. and Azad, P. and Schroder, J. and Bierbaum, A. and Vahrenkamp, N. and Dillmann, R.},
  year = 2006,
  month = dec,
  pages = {169--175},
  publisher = {IEEE},
  doi = {10.1109/ICHR.2006.321380},
  isbn = {1-4244-0199-2}
}

@article{Bauer2008,
  title = {Human-{{Robot Collaboration}}: A {{Survey}}},
  author = {Bauer, Andrea and Wollherr, Dirk and Buss, Martin},
  year = 2008,
  month = mar,
  journal = {International Journal of Humanoid Robotics},
  volume = {05},
  number = {01},
  pages = {47--66},
  issn = {0219-8436},
  doi = {10.1142/S0219843608001303},
  isbn = {0219843608001}
}

@article{Bavelas1986,
  title = {"{{I Show How You Feel}}": {{Motor Mimicry}} as a {{Communicative Act}}},
  shorttitle = {"{{I}} Show How You Feel"},
  author = {Bavelas, Janet B. and Black, Alex and Lemery, Charles R. and Mullett, Jennifer},
  year = 1986,
  month = feb,
  journal = {Journal of Personality and Social Psychology},
  volume = {50},
  number = {2},
  pages = {322--329},
  issn = {1939-1315, 0022-3514},
  doi = {10.1037/0022-3514.50.2.322},
  urldate = {2024-09-20},
  langid = {english}
}

@inproceedings{Baxter2016,
  title = {From Characterising Three Years of {{HRI}} to Methodology and Reporting Recommendations},
  booktitle = {2016 11th {{ACM}}/{{IEEE International Conference}} on {{Human-Robot Interaction}} ({{HRI}})},
  author = {Baxter, Paul and Kennedy, James and Senft, Emmanuel and Lemaignan, Severin and Belpaeme, Tony},
  year = 2016,
  month = mar,
  volume = {2016-April},
  pages = {391--398},
  publisher = {IEEE},
  issn = {21672148},
  doi = {10.1109/HRI.2016.7451777},
  isbn = {978-1-4673-8370-7}
}

@article{Belkaid2021,
  title = {Mutual Gaze with a Robot Affects Human Neural Activity and Delays Decision-Making Processes},
  author = {Belkaid, Marwen and Kompatsiari, Kyveli and De Tommaso, Davide and Zablith, Ingrid and Wykowska, Agnieszka},
  year = 2021,
  month = sep,
  journal = {Science Robotics},
  volume = {6},
  number = {58},
  issn = {2470-9476},
  doi = {10.1126/scirobotics.abc5044}
}

@article{Breazeal2003,
  title = {Toward Sociable Robots},
  author = {Breazeal, Cynthia},
  year = 2003,
  month = mar,
  journal = {Robotics and Autonomous Systems},
  volume = {42},
  number = {3-4},
  pages = {167--175},
  issn = {09218890},
  doi = {10.1016/S0921-8890(02)00373-1}
}

@inproceedings{Breazeal2005a,
  title = {Effects of Nonverbal Communication on Efficiency and Robustness in Human-Robot Teamwork},
  booktitle = {2005 {{IEEE}}/{{RSJ International Conference}} on {{Intelligent Robots}} and {{Systems}}},
  author = {Breazeal, Cynthia and Kidd, C.D. and Thomaz, A.L. and Hoffman, Guy and Berlin, Matt},
  year = 2005,
  publisher = {IEEE},
  doi = {10.1109/IROS.2005.1545011},
  isbn = {0-7803-8912-3}
}

@inproceedings{Bruce2002,
  title = {The Role of Expressiveness and Attention in Human-Robot Interaction},
  booktitle = {Proceedings 2002 {{IEEE International Conference}} on {{Robotics}} and {{Automation}} ({{Cat}}. {{No}}.{{02CH37292}})},
  author = {Bruce, Allison and Nourbakhsh, Illah and Simmons, Reid},
  year = 2002,
  volume = {4},
  pages = {4138--4142},
  publisher = {IEEE},
  issn = {10504729},
  doi = {10.1109/ROBOT.2002.1014396},
  isbn = {0-7803-7272-7}
}

@inproceedings{Buschmeier2018,
  title = {Communicative {{Listener Feedback}} in {{Human}}--{{Agent Interaction}}: {{Artificial Speakers Need}} to {{Be Attentive}} and {{Adaptive}}},
  booktitle = {Proceedings of the 17th {{International Conference}} on {{Autonomous Agents}} and {{Multiagent Systems}} ({{AAMAS}} 2018)},
  author = {Buschmeier, Hendrik and Kopp, Stefan},
  year = 2018,
  address = {Stockholm, Sweden},
  langid = {english}
}

@inproceedings{Campos2020,
  title = {Development of {{Human-Robot Communication Technologies}} for {{Future Interaction Experiments}}},
  booktitle = {2020 {{Latin American Robotics Symposium}} ({{LARS}}), 2020 {{Brazilian Symposium}} on {{Robotics}} ({{SBR}}) and 2020 {{Workshop}} on {{Robotics}} in {{Education}} ({{WRE}})},
  author = {Campos, Ana Christina Almada and Adorno, Bruno Vilhena},
  year = 2020,
  month = nov,
  pages = {1--6},
  publisher = {IEEE},
  doi = {10.1109/LARS/SBR/WRE51543.2020.9306965},
  isbn = {978-0-7381-1153-7}
}

@article{Carifio2008,
  title = {Resolving the 50-Year Debate around Using and Misusing {{Likert}} Scales},
  author = {Carifio, James and Perla, Rocco},
  year = 2008,
  month = dec,
  journal = {Medical Education},
  volume = {42},
  number = {12},
  pages = {1150--1152},
  issn = {03080110},
  doi = {10.1111/j.1365-2923.2008.03172.x},
  pmid = {19120943}
}

@article{Che2020,
  title = {Efficient and {{Trustworthy Social Navigation}} via {{Explicit}} and {{Implicit Robot-Human Communication}}},
  author = {Che, Yuhang and Okamura, Allison M. and Sadigh, Dorsa},
  year = 2020,
  journal = {IEEE Transactions on Robotics},
  eprint = {1810.11556},
  pages = {1--16},
  issn = {1552-3098},
  doi = {10.1109/TRO.2020.2964824},
  archiveprefix = {arXiv}
}

@article{Chen2013,
  title = {Robots for Humanity: Using Assistive Robotics to Empower People with Disabilities},
  shorttitle = {Robots for Humanity},
  author = {Chen, T. L. and Ciocarlie, M. and Cousins, S. and Grice, P. M. and Hawkins, K. and {Kaijen Hsiao} and Kemp, C. C. and {Chih-Hung King} and Lazewatsky, D. A. and Leeper, A. E. and {Hai Nguyen} and Paepcke, A. and Pantofaru, C. and Smart, W. D. and Takayama, L.},
  year = 2013,
  month = mar,
  journal = {IEEE Robot. Automat. Mag.},
  volume = {20},
  number = {1},
  pages = {30--39},
  issn = {1070-9932},
  doi = {10.1109/MRA.2012.2229950},
  urldate = {2024-05-31},
  copyright = {https://ieeexplore.ieee.org/Xplorehelp/downloads/license-information/IEEE.html},
  langid = {english}
}

@book{Cohen1988,
  title = {Statistical {{Power Analysis}} for the {{Behavioral Sciences}}},
  author = {Cohen, Jacob},
  year = 1988,
  month = may,
  publisher = {Lawrence Erlbaum Associates},
  isbn = {978-1-134-74270-7}
}

@article{Cowan2001,
  title = {The Magical Number 4 in Short-Term Memory: {{A}} Reconsideration of Mental Storage Capacity},
  shorttitle = {The Magical Number 4 in Short-Term Memory},
  author = {Cowan, Nelson},
  year = 2001,
  month = feb,
  journal = {Behav Brain Sci},
  volume = {24},
  number = {1},
  pages = {87--114},
  issn = {0140-525X, 1469-1825},
  doi = {10.1017/S0140525X01003922},
  urldate = {2025-11-18},
  copyright = {https://www.cambridge.org/core/terms},
  langid = {english}
}

@article{Fiore2013,
  title = {Toward Understanding Social Cues and Signals in Human-Robot Interaction: Effects of Robot Gaze and Proxemic Behavior},
  author = {Fiore, Stephen M. and Wiltshire, Travis J. and Lobato, Emilio J. C. and Jentsch, Florian G. and Huang, Wesley H. and Axelrod, Benjamin},
  year = 2013,
  journal = {Frontiers in Psychology},
  volume = {4},
  number = {NOV},
  pages = {1--15},
  issn = {1664-1078},
  doi = {10.3389/fpsyg.2013.00859},
  pmid = {24348434}
}

@article{Fong2003,
  title = {A Survey of Socially Interactive Robots},
  author = {Fong, Terrence and Nourbakhsh, Illah and Dautenhahn, Kerstin},
  year = 2003,
  month = mar,
  journal = {Robotics and Autonomous Systems},
  volume = {42},
  pages = {143--166},
  issn = {09218890},
  doi = {10.1016/S0921-8890(02)00372-X}
}

@inproceedings{Gockley2005,
  title = {Designing Robots for Long-Term Social Interaction},
  booktitle = {2005 {{IEEE}}/{{RSJ International Conference}} on {{Intelligent Robots}} and {{Systems}}},
  author = {Gockley, Rachel and Bruce, Allison and Forlizzi, Jodi and Michalowski, Marek and Mundell, Anne and Rosenthal, Stephanie and Sellner, Brennan and Simmons, Reid and Snipes, Kevin and Schultz, A.C. and {Jue Wang}},
  year = 2005,
  pages = {1338--1343},
  publisher = {IEEE},
  doi = {10.1109/IROS.2005.1545303},
  isbn = {0-7803-8912-3}
}

@inproceedings{Gockley2006,
  title = {Interactions with a Moody Robot},
  booktitle = {Proceeding of the 1st {{ACM SIGCHI}}/{{SIGART Conference}} on {{Human-Robot Interaction}} - {{HRI}} '06},
  author = {Gockley, Rachel and Forlizzi, Jodi and Simmons, Reid},
  year = 2006,
  pages = {186--193},
  publisher = {ACM Press},
  address = {New York, New York, USA},
  doi = {10.1145/1121241.1121274},
  isbn = {1-59593-294-1}
}

@article{Gombolay2015,
  title = {Decision-Making Authority, Team Efficiency and Human Worker Satisfaction in Mixed Human - Robot Teams},
  author = {Gombolay, Matthew C. and Gutierrez, Reymundo A. and Clarke, Shanelle G. and Sturla, Giancarlo F. and Shah, Julie A.},
  year = 2015,
  month = oct,
  journal = {Autonomous Robots},
  volume = {39},
  number = {3},
  pages = {293--312},
  issn = {0929-5593},
  doi = {10.1007/s10514-015-9457-9}
}

@article{Gombolay2017,
  title = {Computational Design of Mixed-Initiative Human - Robot Teaming That Considers Human Factors: Situational Awareness, Workload, and Workflow Preferences},
  author = {Gombolay, Matthew and Bair, Anna and Huang, Cindy and Shah, Julie},
  year = 2017,
  month = jun,
  journal = {The International Journal of Robotics Research},
  volume = {36},
  number = {5-7},
  pages = {597--617},
  issn = {0278-3649},
  doi = {10.1177/0278364916688255}
}

@article{Guznov2019,
  title = {Robot {{Transparency}} and {{Team Orientation Effects}} on {{Human-Robot Teaming}}},
  author = {Guznov, S. and Lyons, J. and Pfahler, M. and Heironimus, A. and Woolley, M. and Friedman, J. and Neimeier, A.},
  year = 2019,
  month = apr,
  journal = {International Journal of Human-Computer Interaction},
  pages = {650--660},
  publisher = {Taylor \& Francis},
  issn = {1044-7318},
  doi = {10.1080/10447318.2019.1676519}
}

@article{Harpe2015,
  title = {How to Analyze {{Likert}} and Other Rating Scale Data},
  author = {Harpe, Spencer E.},
  year = 2015,
  month = nov,
  journal = {Currents in Pharmacy Teaching and Learning},
  volume = {7},
  number = {6},
  pages = {836--850},
  publisher = {Elsevier},
  issn = {18771297},
  doi = {10.1016/j.cptl.2015.08.001}
}

@article{Heerink2010,
  title = {Assessing {{Acceptance}} of {{Assistive Social Agent Technology}} by {{Older Adults}}: The {{Almere Model}}},
  author = {Heerink, Marcel and Kr{\"o}se, Ben and Evers, Vanessa and Wielinga, Bob},
  year = 2010,
  month = dec,
  journal = {International Journal of Social Robotics},
  volume = {2},
  number = {4},
  pages = {361--375},
  issn = {1875-4791},
  doi = {10.1007/s12369-010-0068-5}
}

@article{Hirano2018,
  title = {How {{Do Communication Cues Change Impressions}} of {{Human-Robot Touch Interaction}}?},
  author = {Hirano, Takahiro and Shiomi, Masahiro and Iio, Takamasa and Kimoto, Mitsuhiko and Tanev, Ivan and Shimohara, Katsunori and Hagita, Norihiro},
  year = 2018,
  month = jan,
  journal = {International Journal of Social Robotics},
  volume = {10},
  pages = {21--31},
  publisher = {Springer Netherlands},
  issn = {1875-4791},
  doi = {10.1007/s12369-017-0425-8}
}

@article{Hoffman2021,
  title = {A {{Primer}} for {{Conducting Experiments}} in {{Human}}--{{Robot Interaction}}},
  author = {Hoffman, Guy and Zhao, Xuan},
  year = 2021,
  month = mar,
  journal = {J. Hum.-Robot Interact.},
  volume = {10},
  number = {1},
  pages = {1--31},
  issn = {2573-9522},
  doi = {10.1145/3412374},
  urldate = {2025-05-16},
  langid = {english}
}

@article{Hong2021,
  title = {A {{Multimodal Emotional Human}}--{{Robot Interaction Architecture}} for {{Social Robots Engaged}} in {{Bidirectional Communication}}},
  author = {Hong, Alexander and Lunscher, Nolan and Hu, Tianhao and Tsuboi, Yuma and Zhang, Xinyi and Franco Dos Reis Alves, Silas and Nejat, Goldie and Benhabib, Beno},
  year = 2021,
  month = dec,
  journal = {IEEE Trans. Cybern.},
  volume = {51},
  number = {12},
  pages = {5954--5968},
  issn = {2168-2267, 2168-2275},
  doi = {10.1109/TCYB.2020.2974688},
  urldate = {2026-06-23},
  copyright = {https://ieeexplore.ieee.org/Xplorehelp/downloads/license-information/IEEE.html},
  langid = {english}
}

@inproceedings{Huang2016,
  title = {Anticipatory Robot Control for Efficient Human-Robot Collaboration},
  booktitle = {{{ACM}}/{{IEEE International Conference}} on {{Human-Robot Interaction}}},
  author = {Huang, Chien Ming and Mutlu, Bilge},
  year = 2016,
  month = mar,
  volume = {2016-April},
  pages = {83--90},
  publisher = {IEEE},
  issn = {21672148},
  doi = {10.1109/HRI.2016.7451737},
  isbn = {978-1-4673-8370-7}
}

@article{Iwasaki2019,
  title = {"{{That Robot Stared Back}} at {{Me}}!": {{Demonstrating Perceptual Ability Is Key}} to {{Successful Human}} - {{Robot Interactions}}},
  author = {Iwasaki, Masaya and Zhou, Jian and Ikeda, Mizuki and Koike, Yuki and Onishi, Yuya and Kawamura, Tatsuyuki and Nakanishi, Hideyuki},
  year = 2019,
  month = sep,
  journal = {Frontiers in Robotics and AI},
  volume = {6},
  number = {September},
  pages = {1--12},
  issn = {2296-9144},
  doi = {10.3389/frobt.2019.00085}
}

@book{Jackman2009,
  title = {Bayesian {{Analysis}} for the {{Social Sciences}}},
  author = {Jackman, Simon},
  year = 2009,
  month = oct,
  series = {Wiley {{Series}} in {{Probability}} and {{Statistics}}},
  edition = {1},
  publisher = {John Wiley \& Sons, Ltd},
  address = {United Kingdom},
  doi = {10.1002/9780470686621},
  urldate = {2024-06-20},
  copyright = {http://doi.wiley.com/10.1002/tdm\_license\_1.1},
  isbn = {978-0-470-01154-6 978-0-470-68662-1},
  langid = {english}
}

@article{Johnson2013,
  title = {Imitating {{Human Emotions}} with {{Artificial Facial Expressions}}},
  author = {Johnson, David O. and Cuijpers, Raymond H. and Van Der Pol, David},
  year = 2013,
  month = nov,
  journal = {Int J of Soc Robotics},
  volume = {5},
  number = {4},
  pages = {503--513},
  issn = {1875-4791, 1875-4805},
  doi = {10.1007/s12369-013-0211-1},
  urldate = {2024-09-13},
  copyright = {http://www.springer.com/tdm},
  langid = {english}
}

@article{Kelter2020,
  title = {Bayesian Alternatives to Null Hypothesis Significance Testing in Biomedical Research: A Non-Technical Introduction to {{Bayesian}} Inference with {{JASP}}},
  author = {Kelter, Riko},
  year = 2020,
  month = dec,
  journal = {BMC Medical Research Methodology},
  volume = {20},
  number = {1},
  publisher = {BMC Medical Research Methodology},
  issn = {1471-2288},
  doi = {10.1186/s12874-020-00980-6},
  pmid = {32503439}
}

@article{Kelter2021,
  title = {Bayesian and Frequentist Testing for Differences between Two Groups with Parametric and Nonparametric Two-Sample Tests},
  author = {Kelter, Riko},
  year = 2021,
  month = nov,
  journal = {WIREs Computational Statistics},
  volume = {13},
  number = {6},
  pages = {1--29},
  issn = {1939-5108},
  doi = {10.1002/wics.1523}
}

@inproceedings{Knepper2017,
  title = {Implicit {{Communication}} in a {{Joint Action}}},
  booktitle = {Proceedings of the 2017 {{ACM}}/{{IEEE International Conference}} on {{Human-Robot Interaction}}},
  author = {Knepper, Ross A. and Mavrogiannis, Christoforos I. and Proft, Julia and Liang, Claire},
  year = 2017,
  month = mar,
  pages = {283--292},
  publisher = {ACM},
  address = {New York, NY, USA},
  issn = {21672148},
  doi = {10.1145/2909824.3020226},
  isbn = {978-1-4503-4336-7}
}

@article{Kruschke2013,
  title = {Bayesian Estimation Supersedes the t Test.},
  author = {Kruschke, John K.},
  year = 2013,
  journal = {Journal of Experimental Psychology: General},
  volume = {142},
  number = {2},
  pages = {573--603},
  issn = {1939-2222},
  doi = {10.1037/a0029146},
  pmid = {22774788}
}

@book{Kruschke2015,
  title = {Doing {{Bayesian Data Analysis}}: {{A Tutorial}} with {{R}}, {{JAGS}}, and {{Stan}}},
  author = {Kruschke, John K.},
  year = 2015,
  publisher = {Academic Press / Elsevier},
  address = {Burlington, MA},
  isbn = {978-0-12-405888-0}
}

@article{Kruschke2018,
  title = {The {{Bayesian New Statistics}}: {{Hypothesis}} Testing, Estimation, Meta-Analysis, and Power Analysis from a {{Bayesian}} Perspective},
  author = {Kruschke, John K. and Liddell, Torrin M.},
  year = 2018,
  month = feb,
  journal = {Psychonomic Bulletin \& Review},
  volume = {25},
  number = {1},
  pages = {178--206},
  publisher = {Psychonomic Bulletin \& Review},
  issn = {1069-9384},
  doi = {10.3758/s13423-016-1221-4},
  pmid = {28176294}
}

@article{Lazzeri2014,
  title = {Development and {{Testing}} of a {{Multimodal Acquisition Platform}} for {{Human-Robot Interaction Affective Studies}}},
  author = {Lazzeri, Nicole and Mazzei, Daniele and De Rossi, Danilo},
  year = 2014,
  journal = {Journal of Human-Robot Interaction},
  volume = {3},
  number = {2},
  issn = {2163-0364},
  doi = {10.5898/JHRI.3.2.Lazzeri}
}

@inproceedings{Lenz2010,
  title = {The {{BERT2}} Infrastructure: {{An}} Integrated System for the Study of Human-Robot Interaction},
  booktitle = {2010 10th {{IEEE-RAS International Conference}} on {{Humanoid Robots}}},
  author = {Lenz, Alexander and Skachek, Sergey and Hamann, Katharina and Steinwender, Jasmin and Pipe, Anthony G. and Melhuish, Chris},
  year = 2010,
  month = dec,
  pages = {346--351},
  publisher = {IEEE},
  doi = {10.1109/ICHR.2010.5686319},
  isbn = {978-1-4244-8688-5}
}

@article{Liddell2018,
  title = {Analyzing Ordinal Data with Metric Models: {{What}} Could Possibly Go Wrong?},
  author = {Liddell, Torrin M. and Kruschke, John K.},
  year = 2018,
  month = nov,
  journal = {Journal of Experimental Social Psychology},
  volume = {79},
  number = {August},
  pages = {328--348},
  publisher = {Elsevier},
  issn = {00221031},
  doi = {10.1016/j.jesp.2018.08.009}
}

@article{Likert1932,
  title = {A {{Technique}} for the {{Measurement}} of {{Attitudes}}},
  author = {Likert, Rensis},
  year = 1932,
  journal = {Archives of Psychology},
  address = {New York}
}

@inproceedings{Ljungblad2012,
  title = {Hospital {{Robot}} at {{Work}}: {{Something Alien}} or an {{Intelligent Colleague}}?},
  booktitle = {Proceedings of the {{ACM}} 2012 {{Conference}} on {{Computer Supported Cooperative Work}} - {{CSCW}} '12},
  author = {Ljungblad, Sara and Kotrbova, Jirina and Jacobsson, Mattias and Cramer, Henriette and Niechwiadowicz, Karol},
  year = 2012,
  pages = {177--186},
  publisher = {ACM Press},
  address = {New York, New York, USA},
  doi = {10.1145/2145204.2145233},
  isbn = {978-1-4503-1086-4}
}

@article{Luck1997,
  title = {The Capacity of Visual Working Memory for Features and Conjunctions},
  author = {Luck, Steven J. and Vogel, Edward K.},
  year = 1997,
  month = nov,
  journal = {Nature},
  volume = {390},
  number = {6657},
  pages = {279--281},
  issn = {0028-0836, 1476-4687},
  doi = {10.1038/36846},
  urldate = {2026-01-29},
  copyright = {http://www.springer.com/tdm},
  langid = {english}
}

@article{Mavridis2015,
  title = {A Review of Verbal and Non-Verbal Human-Robot Interactive Communication},
  author = {Mavridis, Nikolaos},
  year = 2015,
  month = jan,
  journal = {Robotics and Autonomous Systems},
  volume = {63},
  pages = {22--35},
  publisher = {Elsevier B.V.},
  issn = {09218890},
  doi = {10.1016/j.robot.2014.09.031}
}

@article{Meek2007,
  title = {Comparison of the t vs. {{Wilcoxon Signed-Rank Test}} for {{Likert Scale Data}} and {{Small Samples}}},
  author = {Meek, Gary E. and Ozgur, Ceyhun and Dunning, Kenneth},
  year = 2007,
  month = may,
  journal = {Journal of Modern Applied Statistical Methods},
  volume = {6},
  number = {1},
  pages = {91--106},
  issn = {1538-9472},
  doi = {10.22237/jmasm/1177992540}
}

@article{Meghdari2018,
  title = {Design {{Performance Characteristics}} of a {{Social Robot Companion}} "{{Arash}}" for {{Pediatric Hospitals}}},
  author = {Meghdari, Ali and Shariati, Azadeh and Alemi, Minoo and Nobaveh, Ali Amoozandeh and Khamooshi, Mobin and Mozaffari, Behrad},
  year = 2018,
  month = oct,
  journal = {International Journal of Humanoid Robotics},
  volume = {15},
  number = {05},
  pages = {1850019},
  issn = {0219-8436},
  doi = {10.1142/S0219843618500196}
}

@book{Montgomery2011,
  title = {Applied {{Statistics}} and {{Probability}} for {{Engineers}},{{Fifth Edition}}},
  author = {Montgomery, Douglas C. and Runger, George C.},
  year = 2011,
  publisher = {John Wiley \& Sons, Inc.},
  isbn = {978-0-470-05304-1}
}

@inproceedings{Mutlu2009,
  title = {Nonverbal Leakage in Robots: {{Communication}} of Intentions through Seemingly Unintentional Behavior},
  booktitle = {Proceedings of the 4th {{ACM}}/{{IEEE International Conference}} on {{Human Robot Interaction}} - {{HRI}} '09},
  author = {Mutlu, Bilge and Yamaoka, Fumitaka and Kanda, Takayuki and Ishiguro, Hiroshi and Hagita, Norihiro},
  year = 2009,
  pages = {69--76},
  publisher = {ACM Press},
  address = {New York, New York, USA},
  doi = {10.1145/1514095.1514110},
  isbn = {978-1-60558-404-1}
}

@article{Nanna1998,
  title = {Analysis of {{Likert}} Scale Data in Disability and Medical Rehabilitation Research},
  author = {Nanna, Michael J. and Sawilowsky, Shlomo S.},
  year = 1998,
  journal = {Psychological Methods},
  volume = {3},
  number = {1},
  pages = {55--67},
  issn = {1082-989X},
  doi = {10.1037//1082-989x.3.1.55}
}

@article{Natarajan2023,
  title = {Human-{{Robot Teaming}}: {{Grand Challenges}}},
  shorttitle = {Human-{{Robot Teaming}}},
  author = {Natarajan, Manisha and Seraj, Esmaeil and Altundas, Batuhan and Paleja, Rohan and Ye, Sean and Chen, Letian and Jensen, Reed and Chang, Kimberlee Chestnut and Gombolay, Matthew},
  year = 2023,
  month = aug,
  journal = {Curr Robot Rep},
  issn = {2662-4087},
  doi = {10.1007/s43154-023-00103-1},
  urldate = {2023-08-14},
  langid = {english}
}

@article{Rau2009,
  title = {Effects of Communication Style and Culture on Ability to Accept Recommendations from Robots},
  author = {Rau, P.L. Patrick and Li, Ye and Li, Dingjun},
  year = 2009,
  month = mar,
  journal = {Computers in Human Behavior},
  volume = {25},
  number = {2},
  pages = {587--595},
  publisher = {Elsevier Ltd},
  issn = {07475632},
  doi = {10.1016/j.chb.2008.12.025}
}

@article{Robins2018,
  title = {Kaspar, the Social Robot and Ways It May Help Children with Autism - an Overview},
  author = {Robins, Ben and Dautenhahn, Kerstin and Nadel, Jacqueline},
  year = 2018,
  journal = {Enfance},
  volume = {1},
  number = {1},
  pages = {91},
  issn = {0013-7545},
  doi = {10.3917/enf2.181.0091},
  isbn = {9782130803492}
}

@article{Sebanz2006,
  title = {Joint Action: Bodies and Minds Moving Together},
  author = {Sebanz, Natalie and Bekkering, Harold and Knoblich, G{\"u}nther},
  year = 2006,
  month = feb,
  journal = {Trends in Cognitive Sciences},
  volume = {10},
  number = {2},
  pages = {70--76},
  issn = {13646613},
  doi = {10.1016/j.tics.2005.12.009},
  pmid = {16406326}
}

@inproceedings{Shah2011,
  title = {Improved Human-Robot Team Performance Using Chaski, a Human-Inspired Plan Execution System},
  booktitle = {Proceedings of the 6th {{International Conference}} on {{Human-Robot Interaction}} - {{HRI}} '11},
  author = {Shah, Julie and Wiken, James and Williams, Brian and Breazeal, Cynthia},
  year = 2011,
  pages = {29--36},
  publisher = {ACM Press},
  address = {New York, New York, USA},
  doi = {10.1145/1957656.1957668},
  isbn = {978-1-4503-0561-7}
}

@article{Six2025,
  title = {Impact of {{Conversational}} and {{Animation Features}} of a {{Mental Health App Virtual Agent}} on {{Depressive Symptoms}} and {{User Experience Among College Students}}: {{Randomized Controlled Trial}}},
  shorttitle = {Impact of {{Conversational}} and {{Animation Features}} of a {{Mental Health App Virtual Agent}} on {{Depressive Symptoms}} and {{User Experience Among College Students}}},
  author = {Six, Stephanie and Schlesener, Elizabeth and Hill, Victoria and Babu, Sabarish V and Byrne, Kaileigh},
  year = 2025,
  month = apr,
  journal = {JMIR Ment Health},
  volume = {12},
  pages = {e67381-e67381},
  issn = {2368-7959},
  doi = {10.2196/67381},
  urldate = {2025-05-26},
  langid = {english}
}

@inproceedings{Takayama2009,
  title = {Influences on Proxemic Behaviors in Human-Robot Interaction},
  booktitle = {2009 {{IEEE}}/{{RSJ International Conference}} on {{Intelligent Robots}} and {{Systems}}},
  author = {Takayama, Leila and Pantofaru, Caroline},
  year = 2009,
  month = oct,
  pages = {5495--5502},
  publisher = {IEEE},
  doi = {10.1109/IROS.2009.5354145},
  isbn = {978-1-4244-3803-7}
}

@article{Tanaka2012,
  title = {Effect of a Human-Type Communication Robot on Cognitive Function in Elderly Women Living Alone},
  author = {Tanaka, Masaaki and Ishii, Akira and Yamano, Emi and Ogikubo, Hiroki and Okazaki, Masatsugu and Kamimura, Kazuro and Konishi, Yasuharu and Emoto, Shigeru and Watanabe, Yasuyoshi},
  year = 2012,
  journal = {Medical Science Monitor},
  volume = {18},
  number = {9},
  pages = {CR550-CR557},
  issn = {1234-1010},
  doi = {10.12659/MSM.883350},
  pmid = {22936190}
}

@article{Toh2016,
  title = {A Review on the Use of Robots in Education and Young Children},
  author = {Toh, Lai Poh Emily and Causo, Albert and Tzuo, Pei Wen and Chen, I. Ming and Yeo, Song Huat},
  year = 2016,
  journal = {Educational Technology and Society},
  volume = {19},
  number = {2},
  pages = {148--163},
  issn = {14364522}
}

@article{Unhelkar2017,
  title = {Challenges for {{Communication Decision-Making}} in {{Sequential Human-Robot Collaborative Tasks}}},
  author = {Unhelkar, Vaibhav V and Yang, X Jessie and Shah, Julie A},
  year = 2017,
  journal = {Workshop on Mathematical Models, Algorithms, and Human-Robot Interaction at Robotics: Science and Systems}
}

@article{Unhelkar2018,
  title = {Mobile {{Robots}} for {{Moving-Floor Assembly Lines}}: {{Design}}, {{Evaluation}}, and {{Deployment}}},
  author = {Unhelkar, Vaibhav V. and Dorr, Stefan and Bubeck, Alexander and Lasota, Przemyslaw A. and Perez, Jorge and Siu, Ho Chit and Boerkoel, James C. and Tyroller, Quirin and Bix, Johannes and Bartscher, Stefan and Shah, Julie A.},
  year = 2018,
  month = jun,
  journal = {IEEE Robotics \& Automation Magazine},
  volume = {25},
  number = {2},
  pages = {72--81},
  issn = {1070-9932},
  doi = {10.1109/MRA.2018.2815639}
}

@article{VanMaris2020,
  title = {Designing {{Ethical Social Robots}} - {{A Longitudinal Field Study With Older Adults}}},
  author = {{van Maris}, Anouk and Zook, Nancy and {Caleb-Solly}, Praminda and Studley, Matthew and Winfield, Alan and Dogramadzi, Sanja},
  year = 2020,
  month = jan,
  journal = {Frontiers in Robotics and AI},
  volume = {7},
  number = {January},
  issn = {2296-9144},
  doi = {10.3389/frobt.2020.00001}
}

@article{Vogel2006,
  title = {The Time Course of Consolidation in Visual Working Memory.},
  author = {Vogel, Edward K. and Woodman, Geoffrey F. and Luck, Steven J.},
  year = 2006,
  journal = {Journal of Experimental Psychology: Human Perception and Performance},
  volume = {32},
  number = {6},
  pages = {1436--1451},
  issn = {1939-1277, 0096-1523},
  doi = {10.1037/0096-1523.32.6.1436},
  urldate = {2026-01-29},
  langid = {english}
}

@article{Wagenmakers2018,
  title = {Bayesian Inference for Psychology. {{Part I}}: {{Theoretical}} Advantages and Practical Ramifications},
  author = {Wagenmakers, Eric-Jan and Marsman, Maarten and Jamil, Tahira and Ly, Alexander and Verhagen, Josine and Love, Jonathon and Selker, Ravi and Gronau, Quentin F. and {\v S}m{\'i}ra, Martin and Epskamp, Sacha and Matzke, Dora and Rouder, Jeffrey N. and Morey, Richard D.},
  year = 2018,
  month = feb,
  journal = {Psychonomic Bulletin \& Review},
  volume = {25},
  number = {1},
  pages = {35--57},
  issn = {1069-9384},
  doi = {10.3758/s13423-017-1343-3},
  pmid = {28779455}
}

@inproceedings{Zhang2010,
  title = {A {{Multimodal Real-Time Platform}} for {{Studying Human-Avatar Interactions}}},
  booktitle = {International {{Conference}} on {{Intelligent Virtual Agents}}},
  author = {Zhang, Hui and Fricker, Damian and Yu, Chen},
  year = 2010,
  pages = {49--56},
  issn = {03029743},
  doi = {10.1007/978-3-642-15892-6_6},
  isbn = {3-642-15891-9}
}

@inproceedings{Zhang2025,
  title = {Can You Pass That Tool?: {{Implications}} of {{Indirect Speech}} in {{Physical Human-Robot Collaboration}}},
  shorttitle = {Can You Pass That Tool?},
  booktitle = {Proceedings of the 2025 {{CHI Conference}} on {{Human Factors}} in {{Computing Systems}}},
  author = {Zhang, Yan and Ratnayake, Tharaka Sachintha and Sew, Cherie and Knibbe, Jarrod and Goncalves, Jorge and Johal, Wafa},
  year = 2025,
  month = apr,
  pages = {1--18},
  publisher = {ACM},
  address = {Yokohama Japan},
  doi = {10.1145/3706598.3713780},
  urldate = {2025-05-26},
  isbn = {979-8-4007-1394-1},
  langid = {english}
}



\section*{Supplementary material}

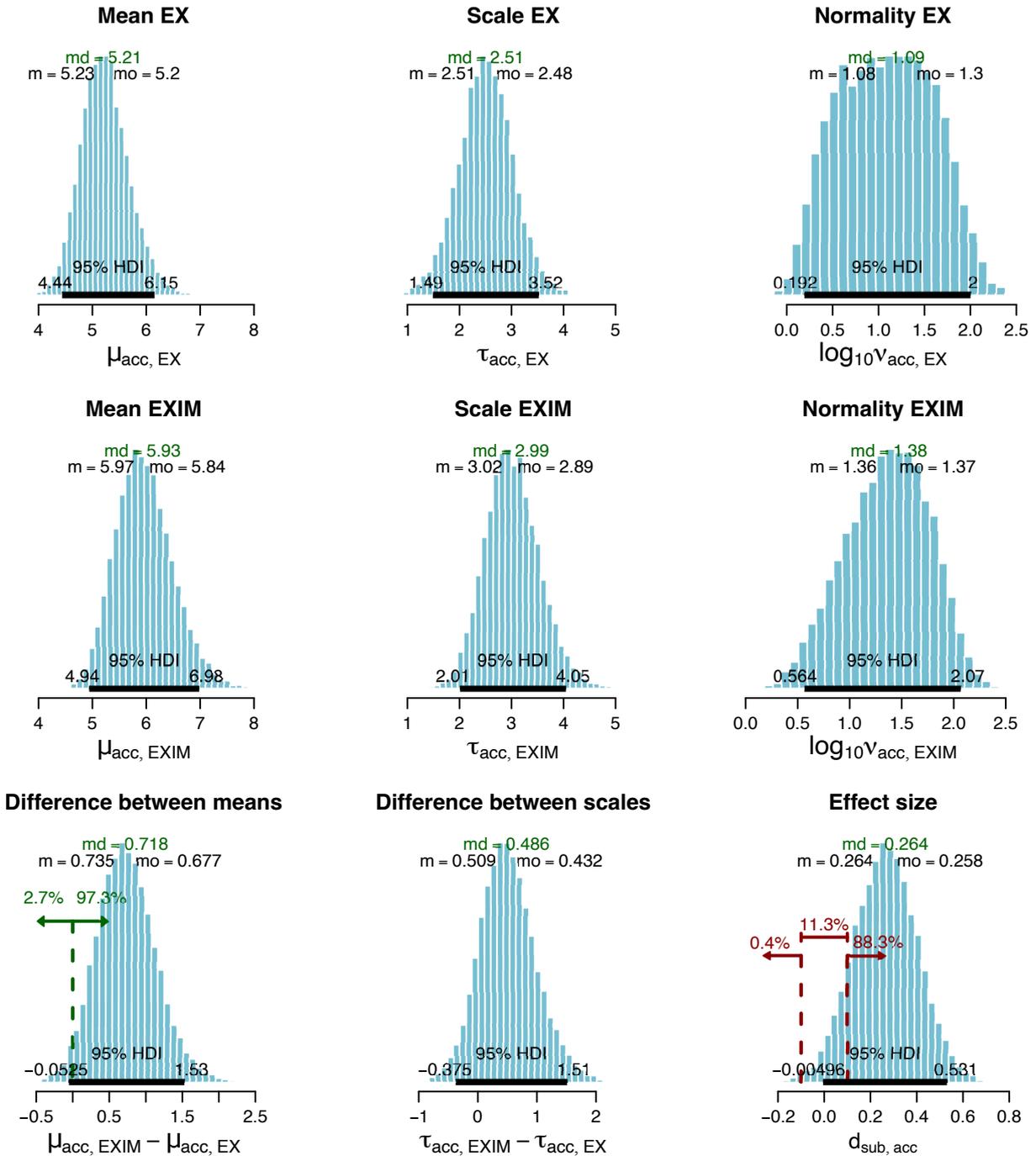

**Fig. 5** Results of the Bayesian inference of the acceptance of the virtual agents in EX and EXIM configurations. The first two rows show the posterior distributions of the mean $\mu$, scale $\tau$, and normality $\nu$ (in log scale) of the latent $t$ distribution of each group. On the left and center of the last row are the distributions of difference between the means and scales of the two groups, and on the right, the distribution of the effect size $d_{sub}$. Mean ($m$), median ($md$), mode ($mo$), and the limits of the 95% HDI are annotated in the distributions. Dashed vertical lines indicate the null value ($\mu_{EXIM} - \mu_{EX} = 0$) in the distribution of the difference between means and the ROPE in the effect size distribution together with the percentages of the distribution below, between and above the values associated with the ROPE and the null value.

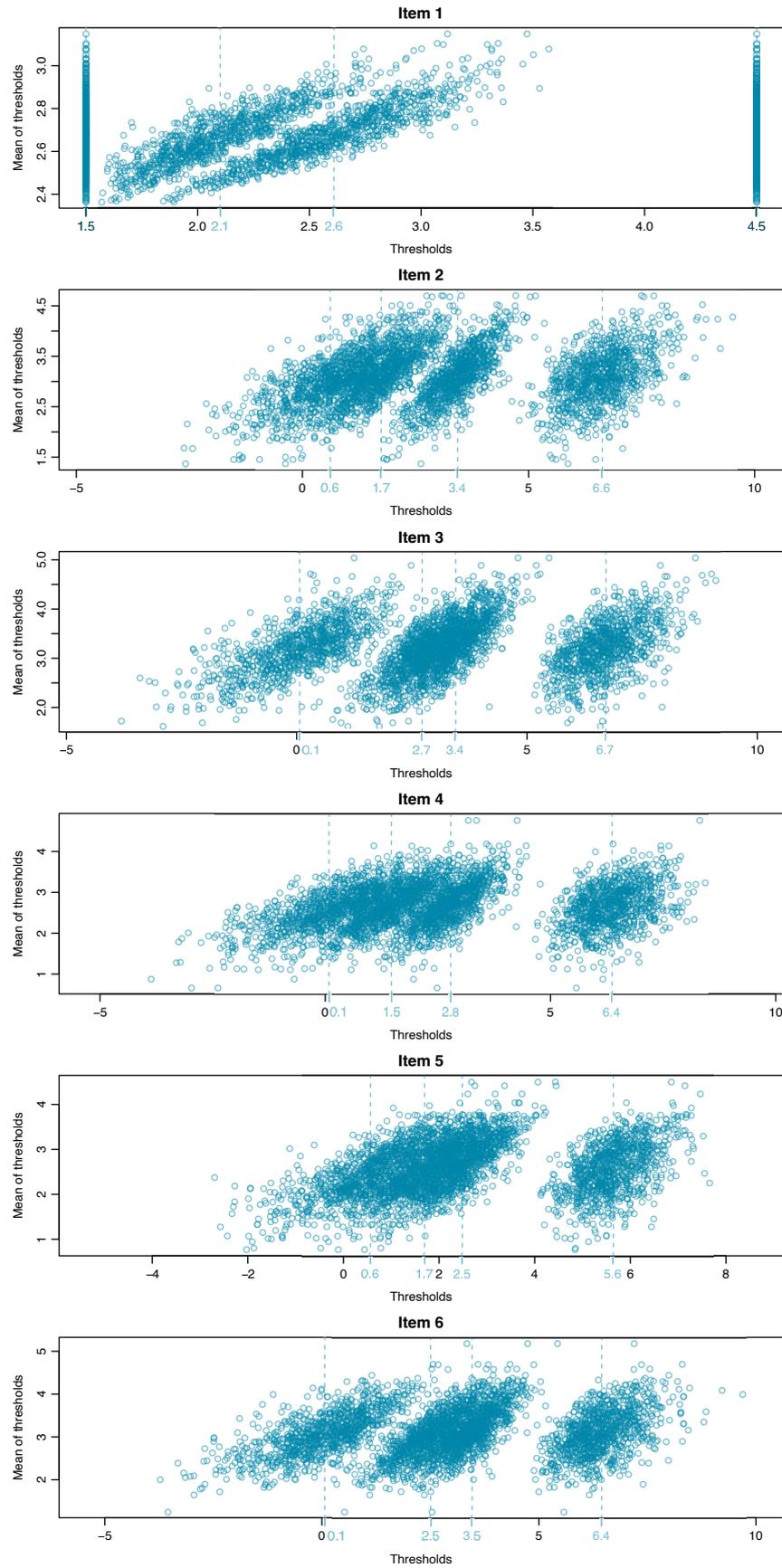

**Fig. 6** Posterior distributions of each item thresholds of the Likert scale for the acceptance of the virtual agents. Dashed lines indicate the means of the thresholds estimations.

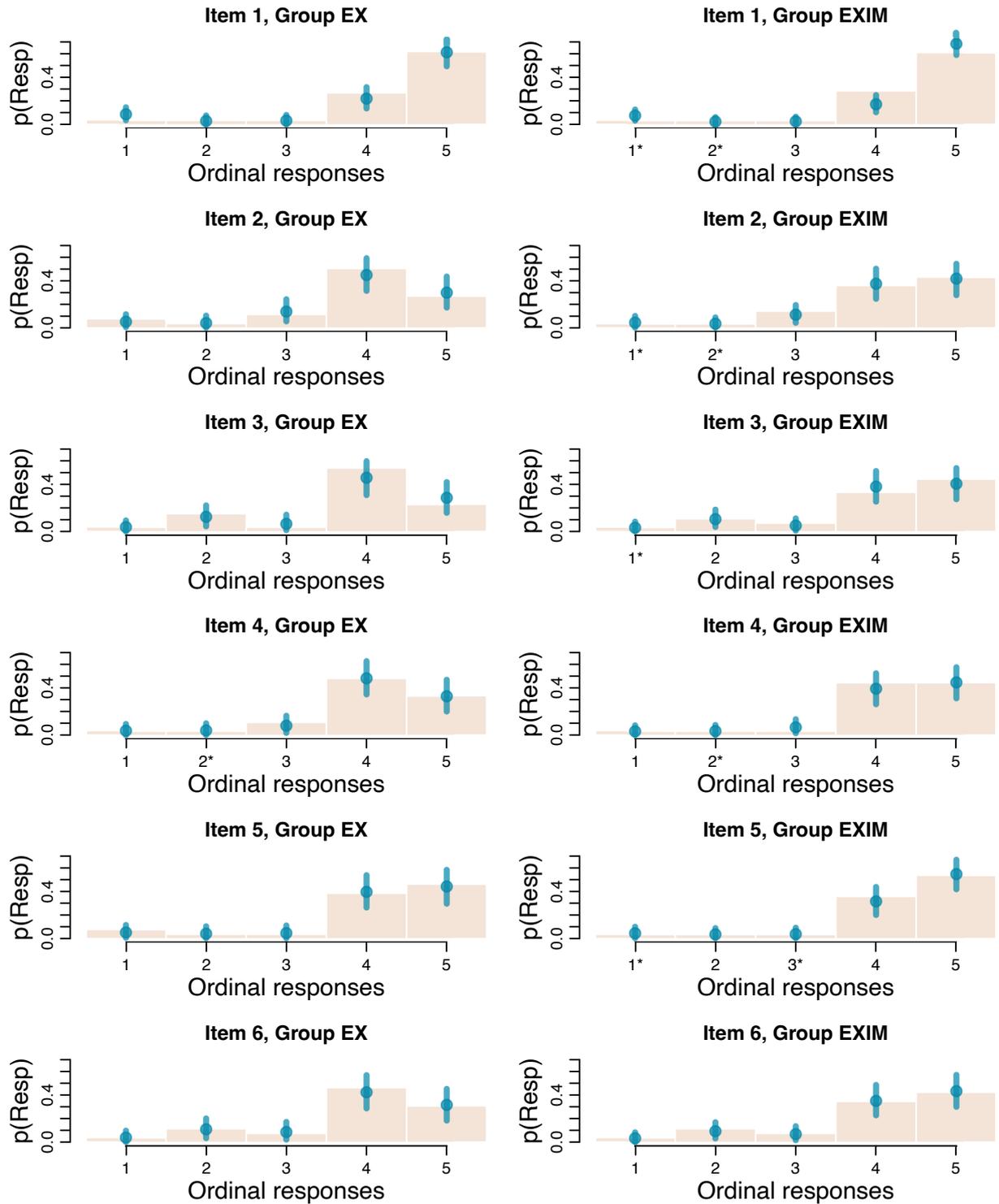

**Fig. 7** Acceptance data histograms superimposed with estimated probabilities to check model adequacy. Each blue dot indicates the estimated median and the vertical line represents the 95% HDI. Levels that had extra answers added are marked with an asterisk (*).

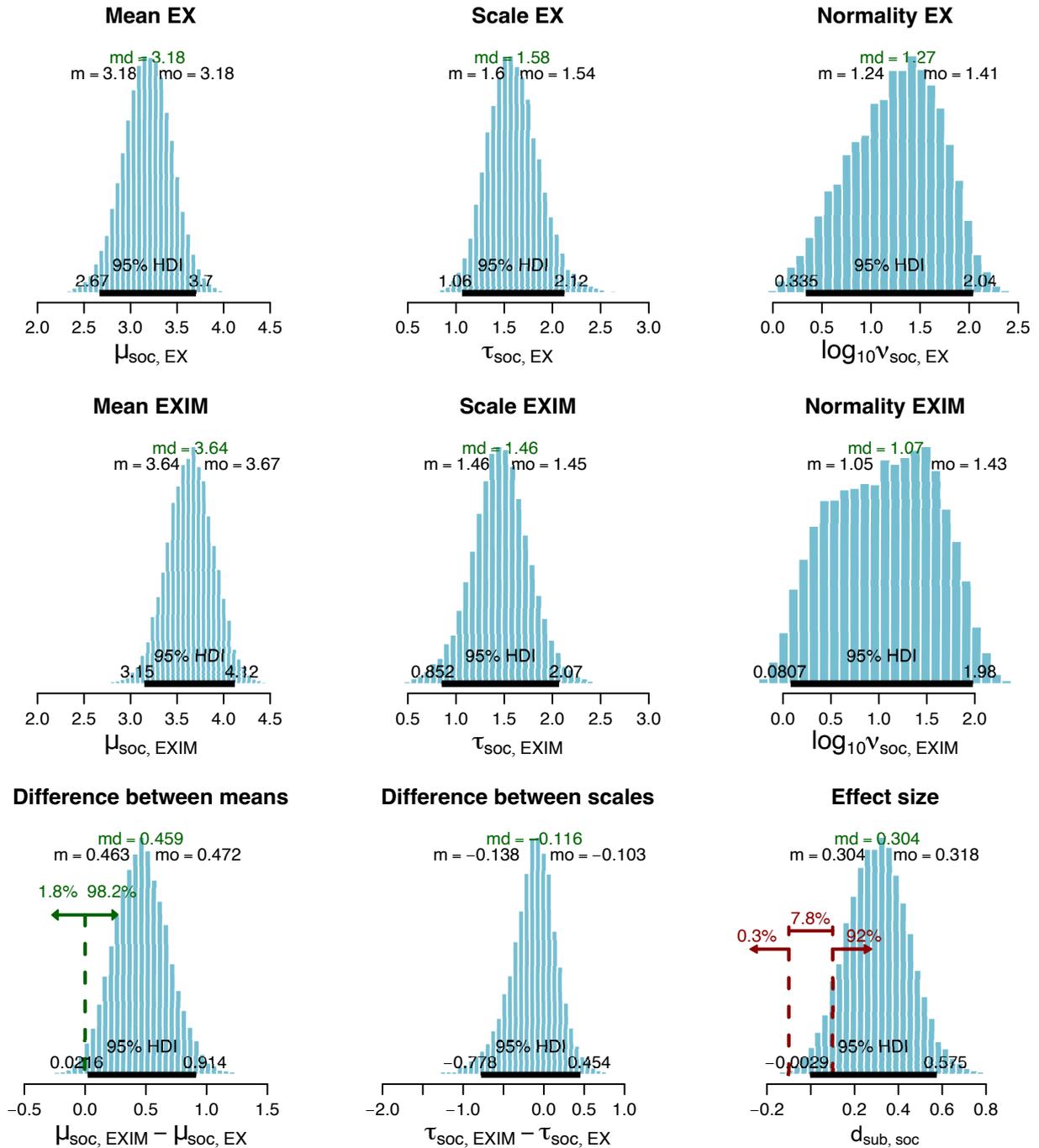

**Fig. 8** Results of the Bayesian inference of the sociability of the virtual agents in EX and EXIM configurations. The first two rows show the posterior distributions of the mean $\mu$, scale $\tau$, and normality $\nu$ (in log scale) of the latent $t$ distribution of each group. On the left and center of the last row are the distributions of difference between the means and scales of the two groups, and on the right, the distribution of the effect size $d_{sub}$. Mean ($m$), median ($md$), mode ($mo$), and the limits of the 95% HDI are annotated in the distributions. Dashed vertical lines indicate the null value ($\mu_{EXIM} - \mu_{EX} = 0$) in the distribution of the difference between means and the ROPE in the effect size distribution together with the percentages of the distribution below, between and above the values associated with the ROPE and the null value.

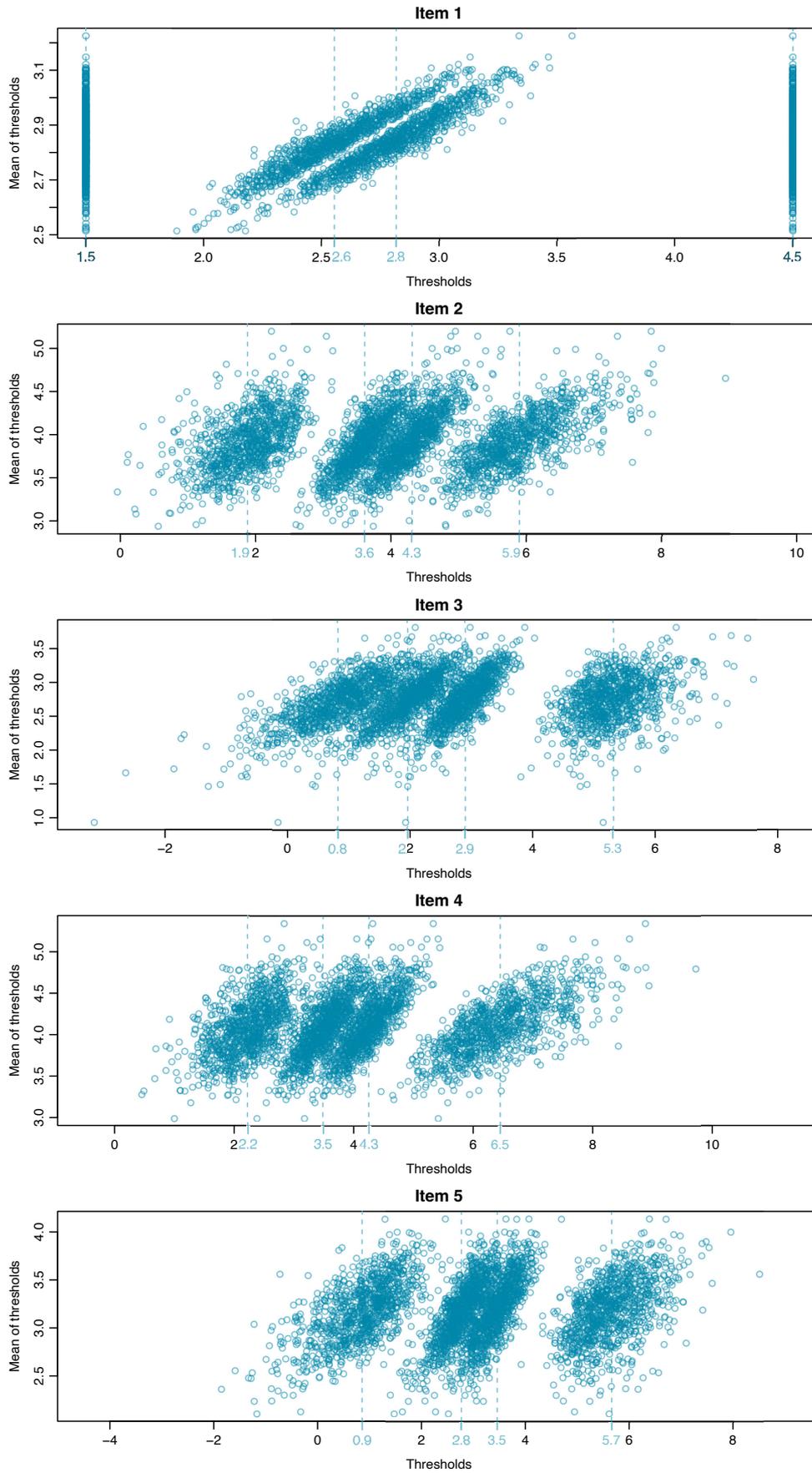

**Fig. 9** Posterior distributions of each item thresholds of the Likert scale for the sociability of the virtual agents. Dashed lines indicate the means of the thresholds estimations.

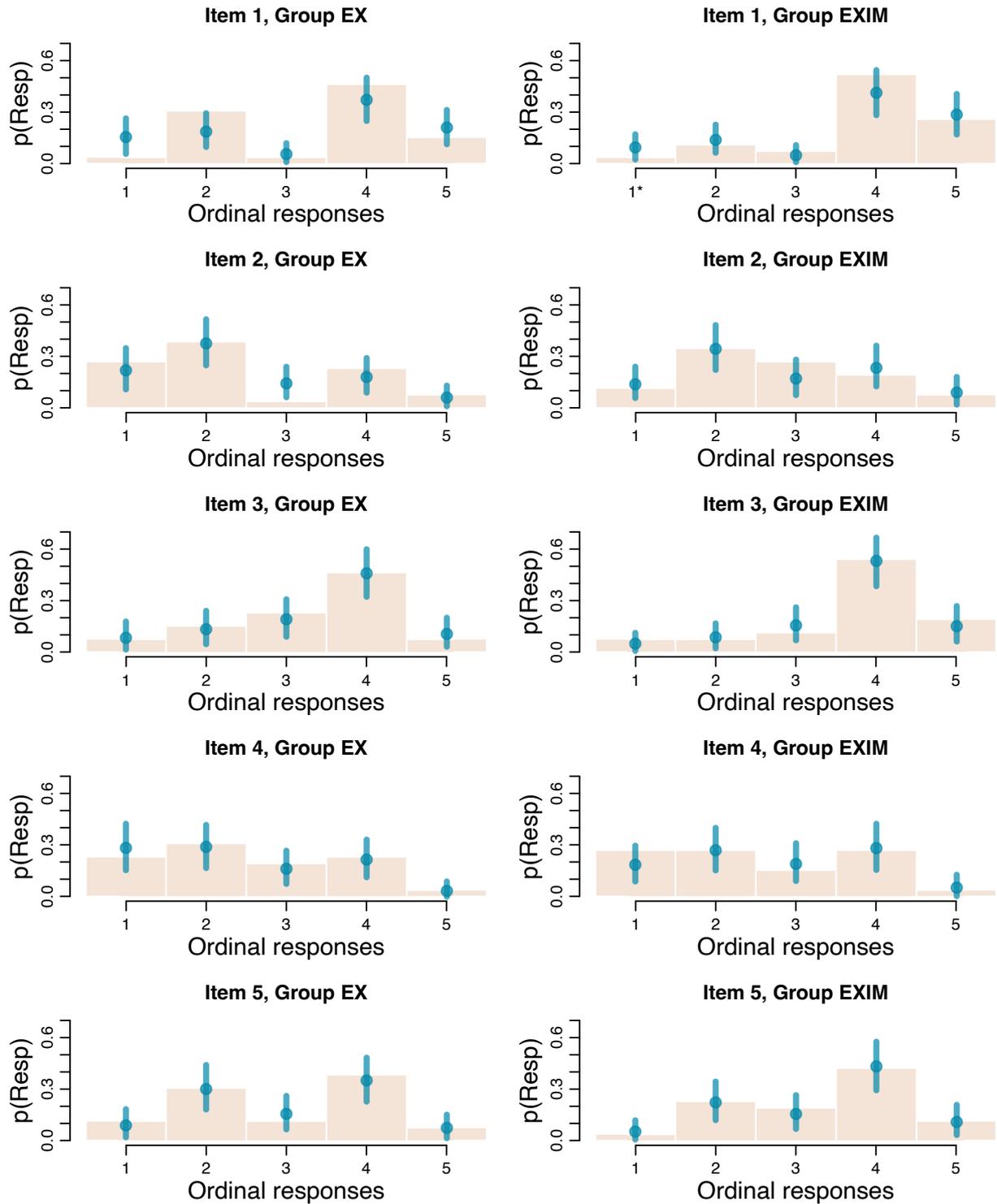

**Fig. 10** Sociability data histograms superimposed to estimated probabilities to check model adequacy. Each blue dot indicates the estimated median and the vertical line the 95% HDI. Levels that had extra answers added are marked with an asterisk (*).

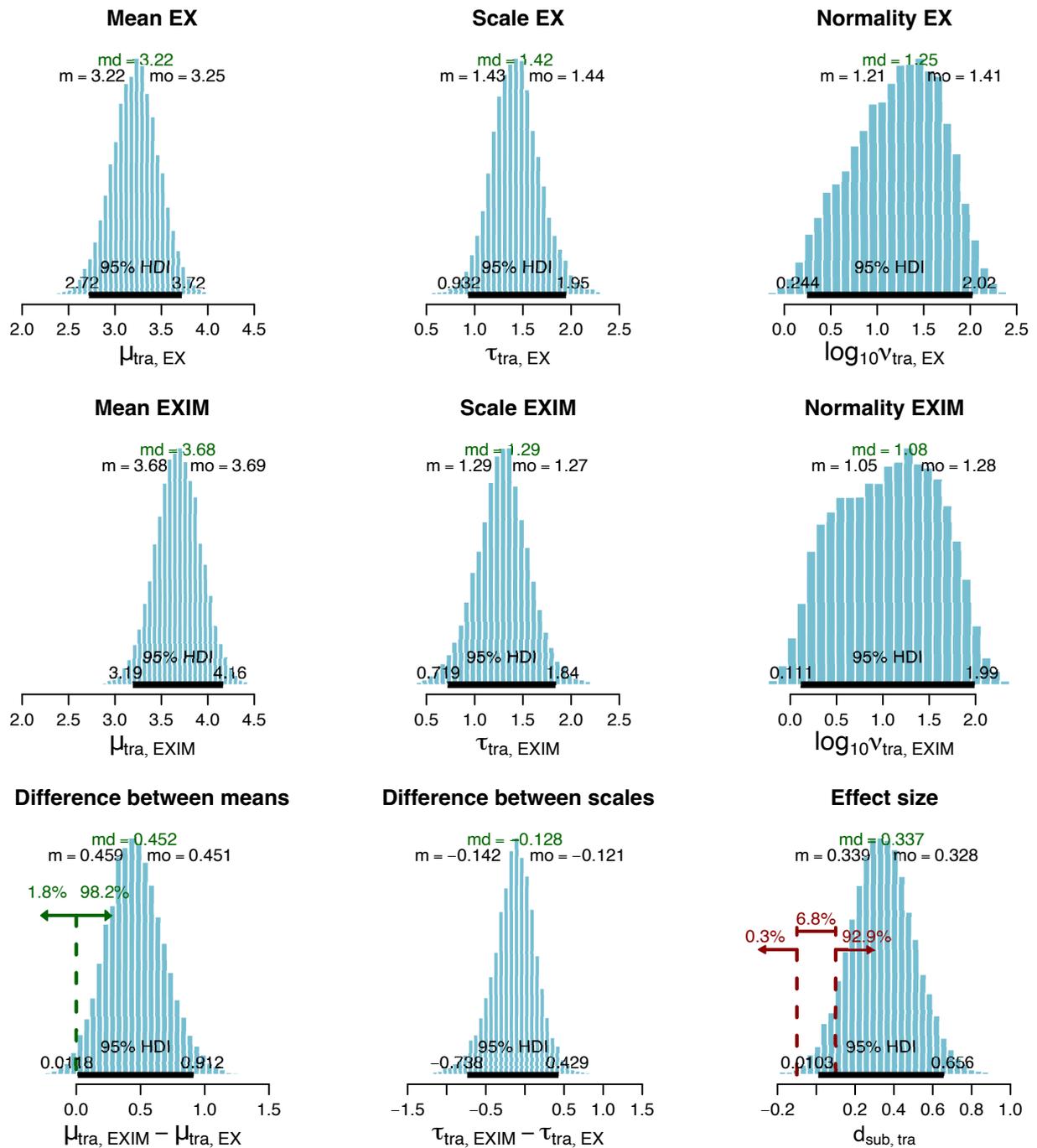

**Fig. 11** Results of the Bayesian inference of the transparency of the virtual agents in EX and EXIM configurations. The first two rows show the posterior distributions of the mean $\mu$, scale $\tau$, and normality $\nu$ (in log scale) of the latent $t$ distribution of each group. On the left and center of the last row are the distributions of difference between the means and scales of the two groups, and on the right, the distribution of the effect size $d_{\text{sub}}$. Mean $(m)$, median $(md)$, mode $(mo)$, and the limits of the 95% HDI are annotated in the distributions. Dashed vertical lines indicate the null value $(\mu_{\text{EXIM}} - \mu_{\text{EX}} = 0)$ in the distribution of the difference between means and the ROPE in the effect size distribution together with the percentages of the distribution below, between and above the values associated with the ROPE and the null value.

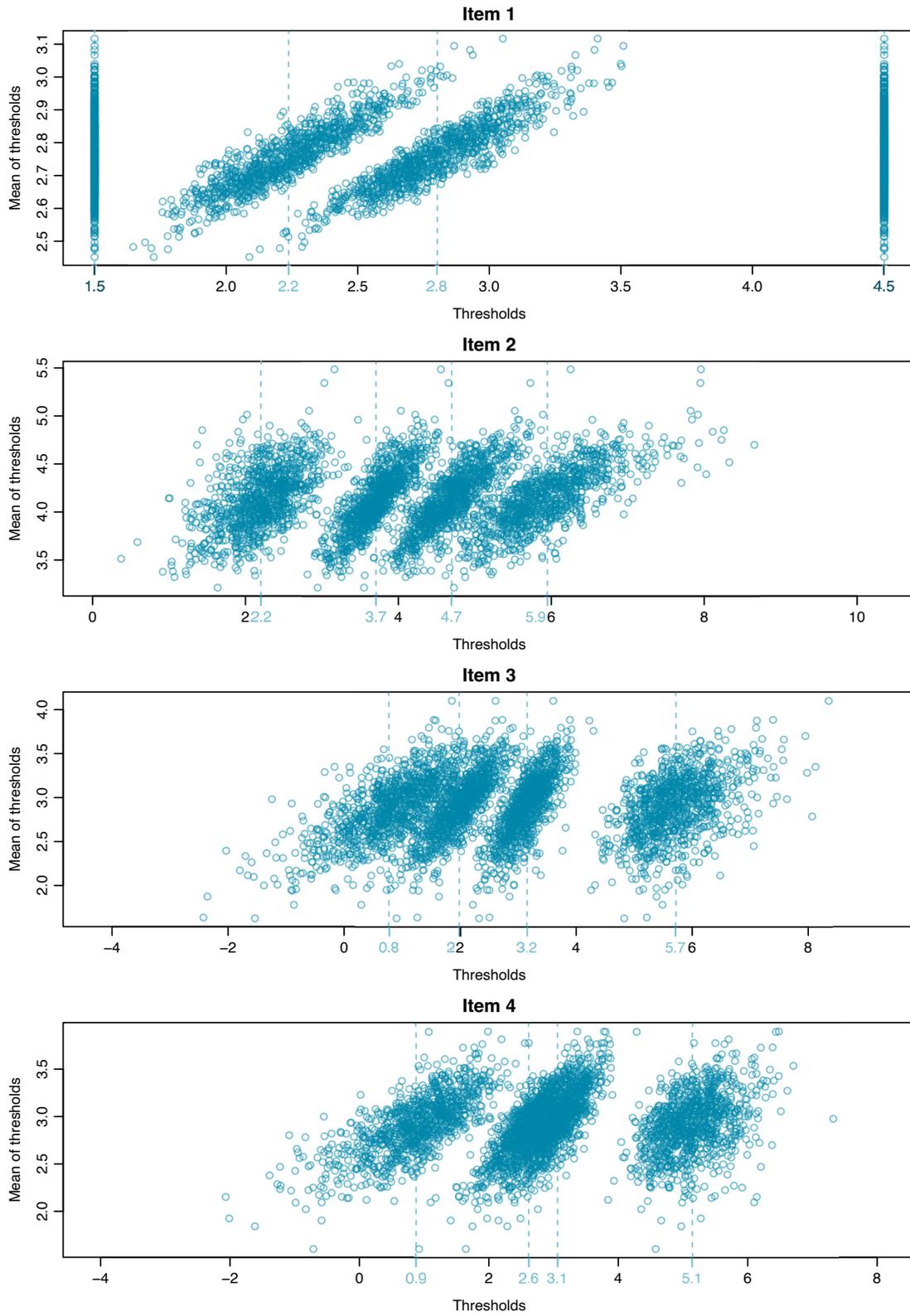

**Fig. 12** Posterior distributions of each item thresholds of the Likert scale for the transparency of the virtual agents. Dashed lines indicate the means of the thresholds estimations.

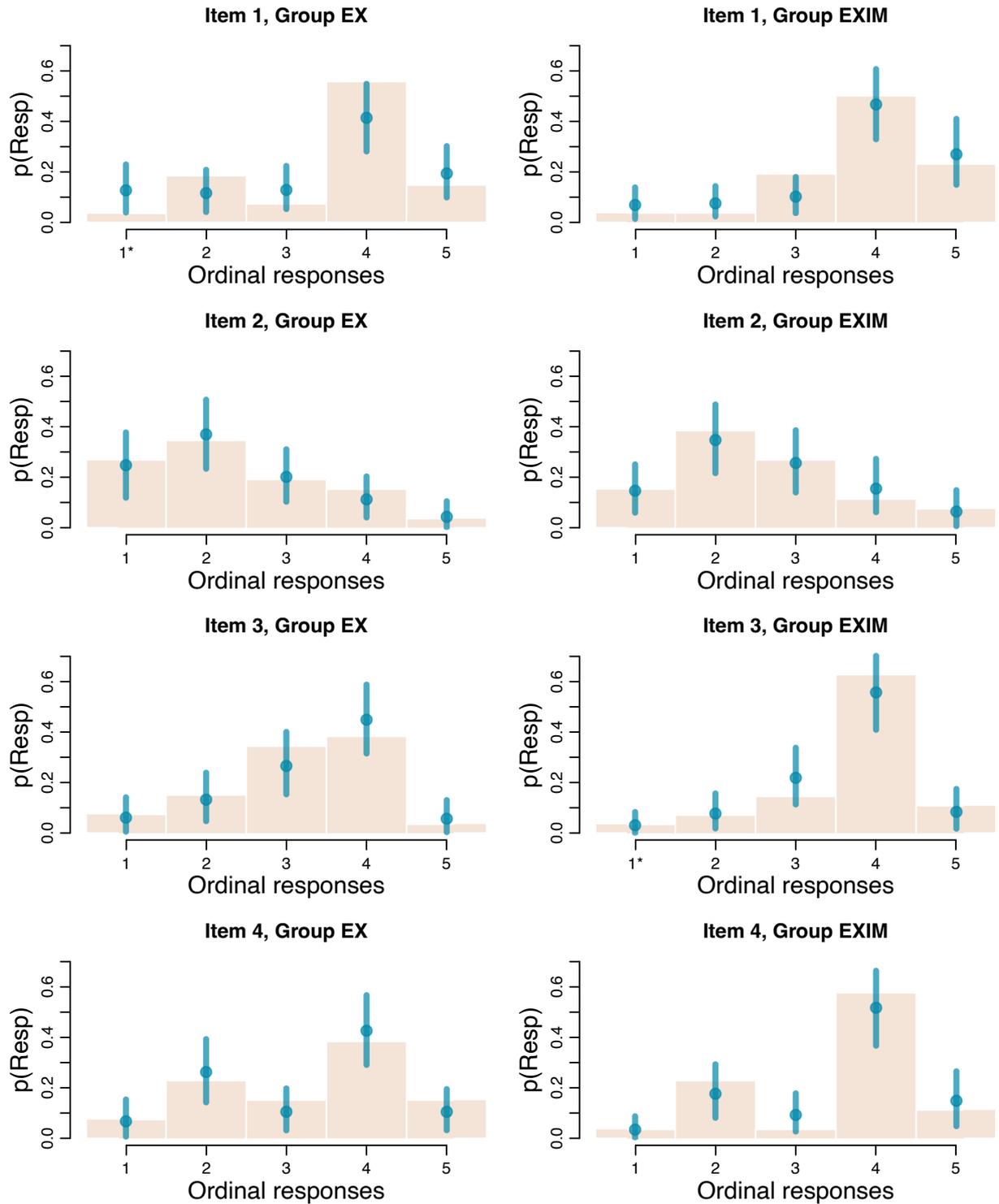

**Fig. 13** Transparency data histograms superimposed to estimated probabilities to check model adequacy. Each blue dot indicates the estimated median and the vertical line the 95% HDI. Levels that had extra answers added are marked with an asterisk (*).

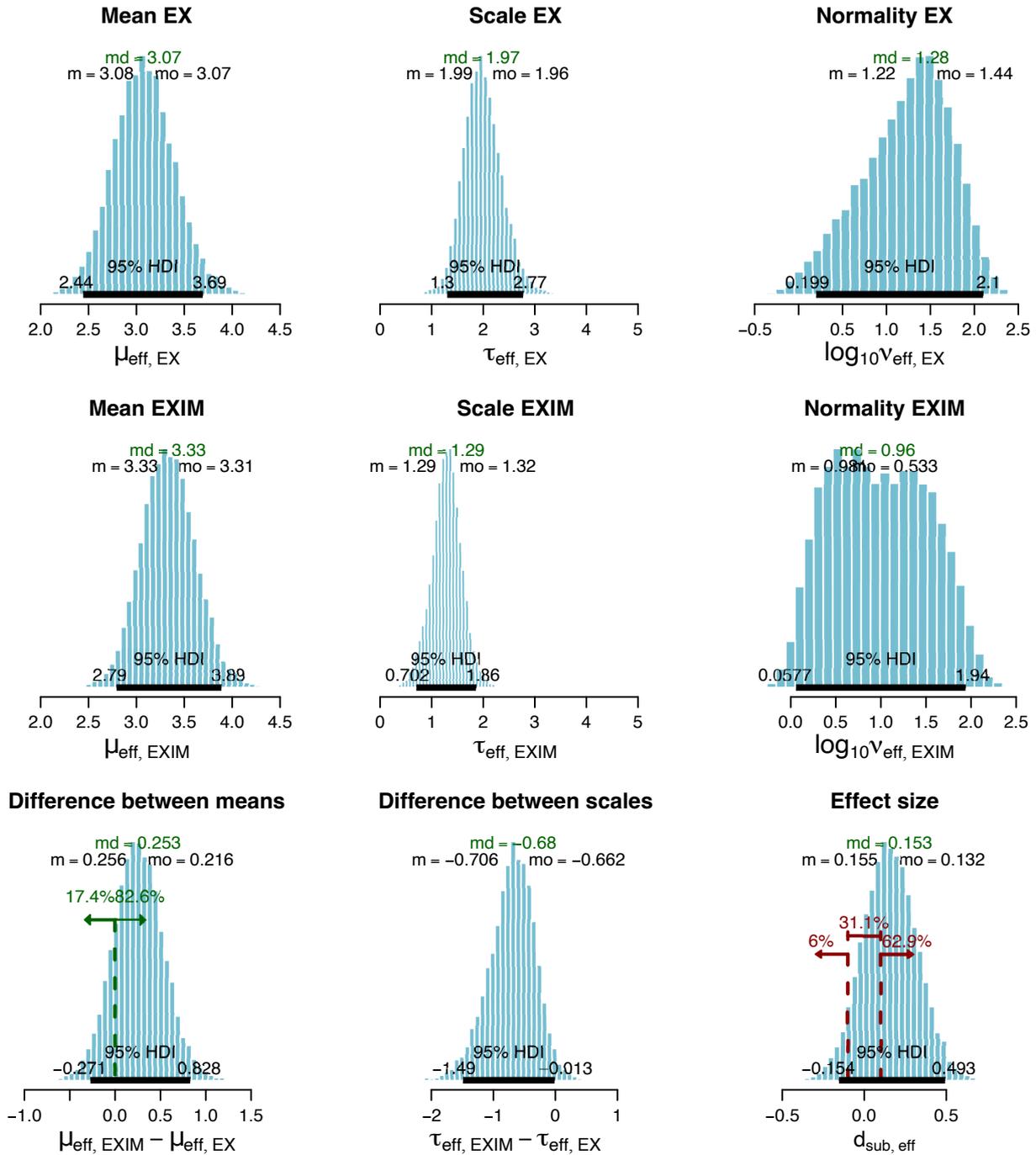

**Fig. 14** Results of the Bayesian inference of the perceived efficiency of the interactions in EX and EXIM configurations. The first two rows show the posterior distributions of the mean $\mu$, scale $\tau$, and normality $\nu$ (in log scale) of the latent $t$ distribution of each group. On the left and center of the last row are the distributions of difference between the means and scales of the two groups, and on the right, the distribution of the effect size $d_{\text{sub}}$. Mean ($m$), median ($md$), mode ($mo$), and the limits of the 95% HDI are annotated in the distributions. Dashed vertical lines indicate the null value ($\mu_{\text{EXIM}} - \mu_{\text{EX}} = 0$) in the distribution of the difference between means and the ROPE in the effect size distribution together with the percentages of the distribution below, between and above the values associated with the ROPE and the null value.

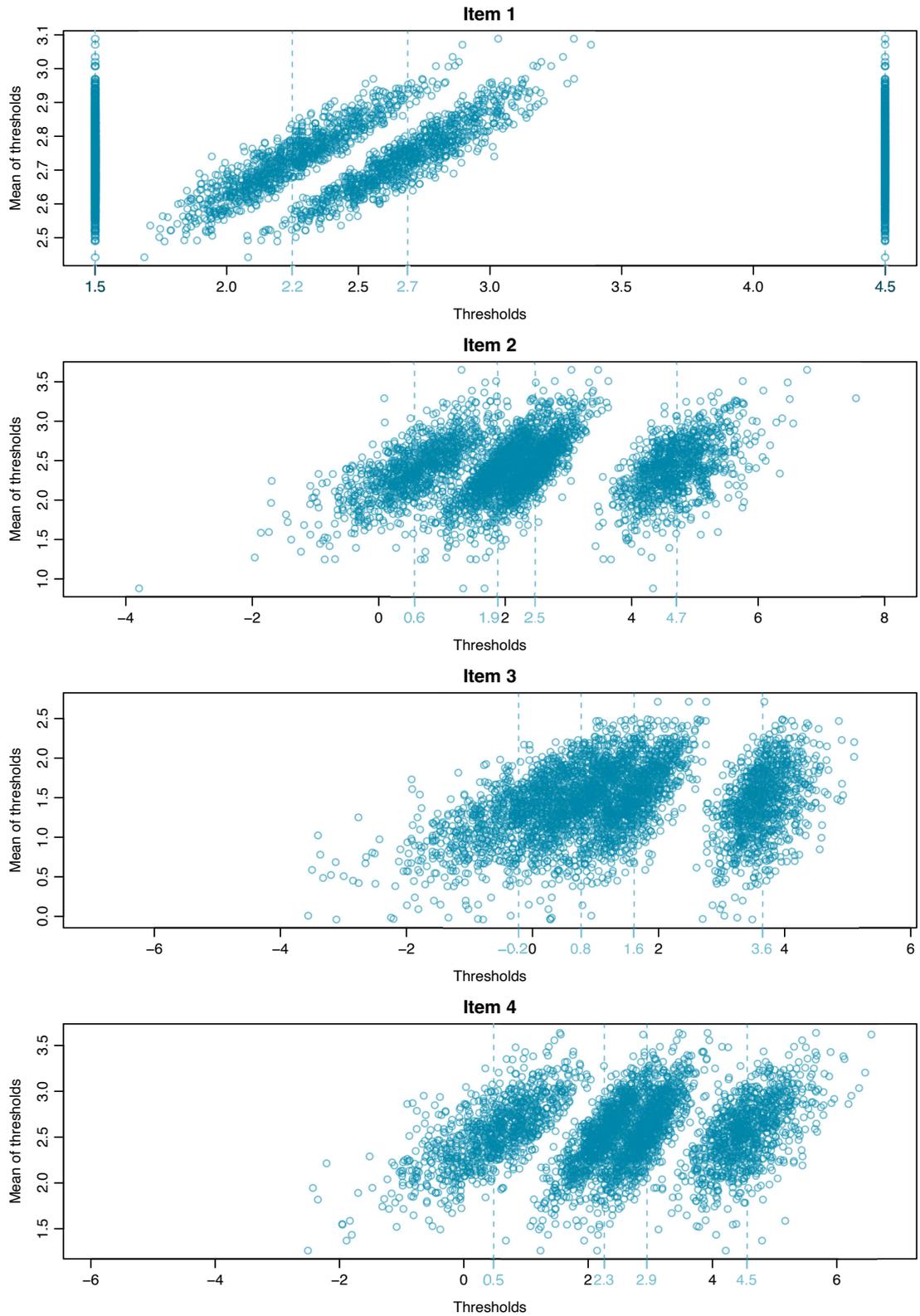

**Fig. 15** Posterior distributions of each item thresholds of the Likert scale for the perceived efficiency of the interactions. Dashed lines indicate the means of the thresholds estimations.

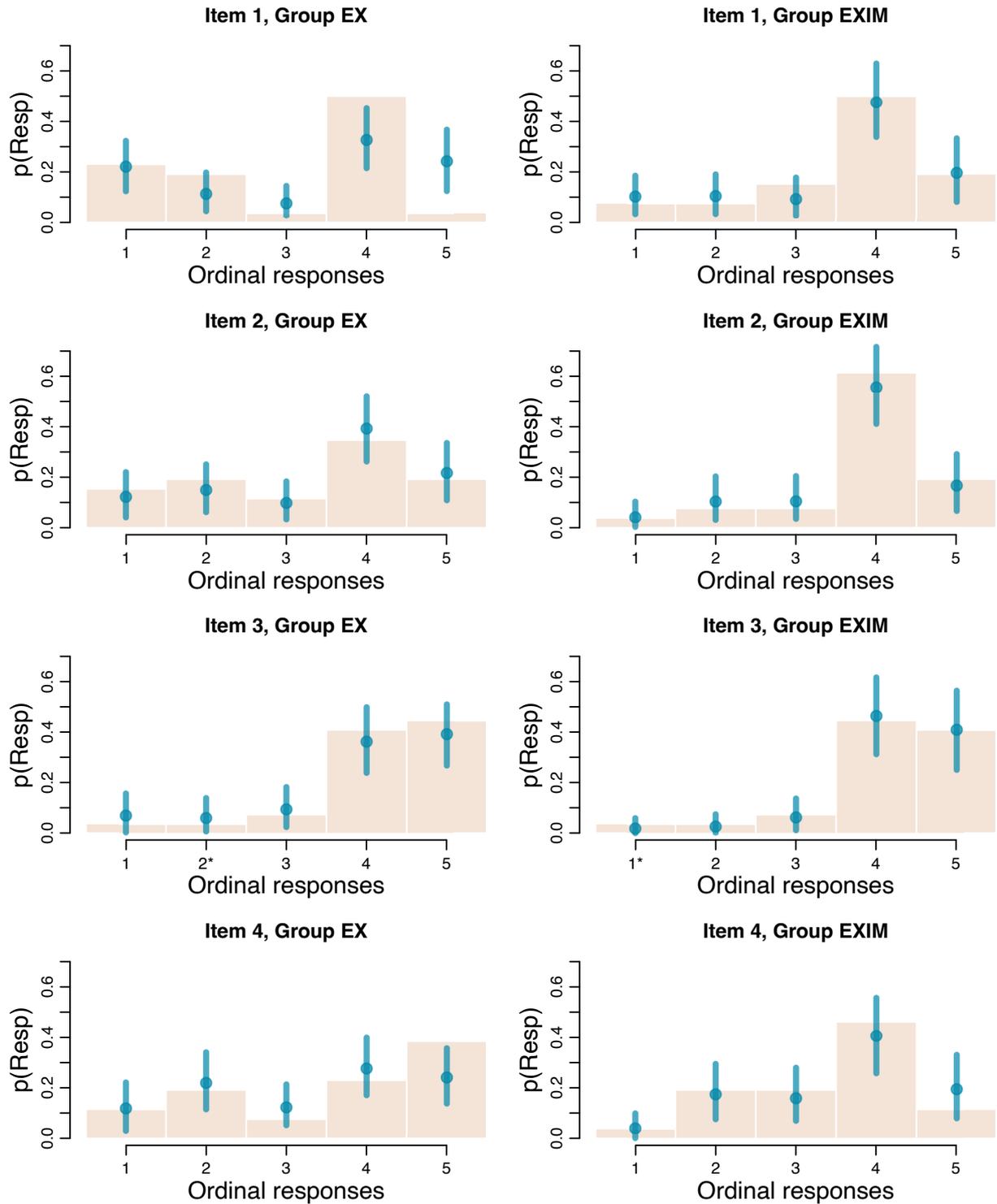

**Fig. 16** Perceived efficiency data histograms superimposed to estimated probabilities to check model adequacy. Each blue dot indicates the estimated median and the vertical line the 95% HDI. Levels that had extra answers added are marked with an asterisk (*).


\end{document}